\documentclass[10pt,twocolumn,letterpaper]{article}

\usepackage{cvpr}              
\usepackage[dvipsnames]{xcolor}
\definecolor{cvprblue}{rgb}{0.21,0.49,0.74}
\definecolor{myYellow}{rgb}{0.80,0.80,0.15}
\definecolor{myGreen}{rgb}{0.20,0.80,0.20}
\definecolor{DarkGreen}{rgb}{0.00,0.50,0.50}
\definecolor{DarkPurple}{rgb}{0.30,0.00,0.70}

\usepackage{graphicx}
\usepackage{booktabs}
\usepackage{adjustbox}
\usepackage{float} 
\usepackage[accsupp]{axessibility}  

\usepackage[pagebackref,breaklinks,colorlinks,citecolor=cvprblue]{hyperref}

\usepackage{orcidlink}

\definecolor{brickred}{rgb}{0.8, 0.25, 0.33}
\definecolor{brightmaroon}{rgb}{0.76, 0.13, 0.28}
\definecolor{brightlavender}{rgb}{0.75, 0.58, 0.89}
\definecolor{byzantine}{rgb}{0.74, 0.2, 0.64}
\definecolor{byzantium}{rgb}{0.44, 0.16, 0.39}

\newcommand{\termCOLOR}[1]{\textcolor{black}{#1}} 

\newcommand{\cgrasp}{\mbox{\termCOLOR{CGrasp}}\xspace}
\newcommand{\creach}{\mbox{\termCOLOR{CReach}}\xspace}
\newcommand{\contactgen}{\mbox{\textcolor{black}{ContactGen}}\xspace}
\newcommand{\dexgraspnet}{\mbox{\textcolor{black}{DexGraspNet}}\xspace}
\newcommand{\hoidiffusion}{\mbox{\textcolor{black}{HOIDiffusion}}\xspace}
\newcommand{\geneohdiffusion}{\mbox{\textcolor{black}{GeneOH Diffusion}}\xspace}
\newcommand{\grip}{\mbox{\textcolor{black}{GRIP}}\xspace}
\newcommand{\gears}{\mbox{\textcolor{black}{GEARS}}\xspace}
\newcommand{\graspxl}{\mbox{\textcolor{black}{GraspXL}}\xspace}
\newcommand{\cWGrasp}{\mbox{\termCOLOR
{CWGrasp}}\xspace}
\newcommand{\cWGraspFULL}{{\termCOLOR
{C}ontrollable \termCOLOR
{W}hole-body \termCOLOR
{Grasp} synthesis}\xspace}

\newcommand{\reachingField}{\mbox{\termCOLOR{ReachingField}}\xspace}
\newcommand{\cnet}{\mbox{\textcolor{black}{CoarseNet}}\xspace}
\newcommand{\rnet}{\mbox{\textcolor{black}{RefineNet}}\xspace}
\newcommand{\grabnet}{\mbox{\textcolor{black}{GrabNet}}\xspace}

\newcommand{\gnet}{\mbox{\textcolor{black}{GNet}}\xspace}
\newcommand{\circleData}{\mbox{CIRCLE}\xspace}

\newcommand{\ControllableGrasp}{{\textcolor{cvprblue}{C}ontrollable \textcolor{cvprblue}{Grasp}}\xspace}
\newcommand{\ControllableReach}{{\textcolor{cvprblue}{C}ontrollable \textcolor{cvprblue}{Reach}}\xspace}

\newcommand{\cVAE}{\mbox{cVAE}\xspace}

\newcommand{\ourTitle}{3D Whole-Body Grasp Synthesis with Directional Controllability}

\usepackage{ulem} 

\usepackage{enumitem} 

\newcommand{\highlightTERM}[1]{{\textcolor{black}{{#1}}}\xspace}
\newcommand{\direction}{\highlightTERM{direction}}
\newcommand{\directions}{\highlightTERM{directions}}

\newcommand{\IAfeaturesShort}{{\mbox{InterField}}\xspace} 

\newcommand{\IAfeatures}{\IAfeaturesShort}

\newcommand{\mano}{\mbox{MANO}\xspace}
\newcommand{\smplx}{\mbox{SMPL-X}\xspace}
\newcommand{\smplX}{\smplx}
\newcommand{\grab}{\mbox{GRAB}\xspace}
\newcommand{\goal}{\mbox{GOAL}\xspace}

\newcommand{\FLEX}{\mbox{FLEX}\xspace}
\newcommand{\flex}{\FLEX}

\newcommand{\manipnet}{\mbox{ManipNet}\xspace}

\newcommand{\ReplicaGrasp}{\mbox{ReplicaGrasp}\xspace}

\newcommand{\HOI}{\mbox{HOI}\xspace}

\newcommand{\threeD}{{3D}\xspace}

\renewcommand{\etal}{\mbox{et al.}\xspace}
\renewcommand{\ie}{\mbox{i.e.}\xspace}
\renewcommand{\eg}{\mbox{e.g.}\xspace}
\renewcommand{\wrt}{\mbox{w.r.t.}\xspace}
\renewcommand{\emph}[1]{\textit{{#1}}}

\newcommand{\mesh}{M}
\newcommand{\bodyWhole}{{wb}}
\newcommand{\bodyMain}{b}

\newcommand{\hand}{h}
\newcommand{\rhand}{right-hand\xspace}
\newcommand{\lhand}{left-hand\xspace}
\newcommand{\obj}{o}
\newcommand{\trans}{{\gamma}}
\newcommand{\betas}{{\beta}}

\newcommand{\betasNumb}{{10}}
\newcommand{\graspDir}{d_\text{grasp}}

\newcommand{\armDir}{d_\text{arm}}

\newcommand{\creachTarget}{L_\text{target}}
\newcommand{\gnetTarget}{L_\text{target}}
\newcommand{\inthand}{h_\text{int}}

\newcommand{\inter}{{\text{inter}}}
\newcommand{\ifeat}{f_\inter}

\newcommand{\vertex}{v}

\newcommand{\supmatCOLOR}{black}
\newcommand{\video}{\textcolor{\supmatCOLOR}{{video}}\xspace}
\newcommand{\supmat}{\textcolor{\supmatCOLOR}{{Sup.~Mat.}}\xspace}

\newcommand{\wrist}{\text{wrist}}

\usepackage[capitalize]{cleveref}
\newcommand{\colorRef}[1]{\textcolor{black}{#1}} 
\crefname{figure}{\colorRef{Fig.}}{\colorRef{Figs.}}
\Crefname{figure}{\colorRef{Figure}}{\colorRef{Figures}}
\crefname{section}{\colorRef{Sec.}}{\colorRef{Secs.}}
\Crefname{section}{\colorRef{Section}}{\colorRef{Sections}}
\Crefname{table}{\colorRef{Table}}{\colorRef{Tables}}
\crefname{table}{\colorRef{Tab.}}{\colorRef{Tabs.}}
\Crefname{equation}{\colorRef{Equation}}{\colorRef{Equations}}
\crefname{equation}{\colorRef{Eq.}}{\colorRef{Eqs.}}

\usepackage{dsfont}

\usepackage{cuted} 
\usepackage{lipsum} 
\usepackage{tabularx}
\usepackage{layouts}

\usepackage{balance}
\usepackage{afterpage}
\usepackage{placeins} 

\newcommand{\qheading}[1]{\noindent\textbf{#1:}}
\newcommand{\zheading}[1]{\textbf{#1:}}
\usepackage{xfrac}

\usepackage{nicefrac, xfrac} 

\usepackage{fontawesome}

\usepackage{pifont}
\newcommand{\cmark}{\color{green}\ding{51}}
\newcommand{\xmark}{\color{red}\ding{55}}

\usepackage{stfloats} 

\newcommand{\graspbody}{\textcolor{black}{{\mathcal{B}}}}
\newcommand{\grasprhand}{\textcolor{black}{{\mathcal{RH}}}}
\newcommand{\receptacle}{\textcolor{black}{{\mathcal{M}}}}
\newcommand{\object}{\textcolor{black}{{\mathcal{O}}}}
\newcommand{\objectcentroid}{\textcolor{black}{{\mathbf{c}}}}

\newcommand{\ray}{\textcolor{black}{{r_i}}}
\newcommand{\rayvertical}{\textcolor{black}{{\ray_j}}}

\newcommand{\gridpoint}{\textcolor{black}{{s_i}}}

\title{\ourTitle}
\author{
Georgios Paschalidis$^{1}$    \quad 
Romana Wilschut$^{1}$       \quad 
Dimitrije Antić$^1$         \quad 
Omid Taheri$^2$             \quad 
Dimitrios Tzionas$^1$       \\
{\small
$^1$University of Amsterdam, the Netherlands \quad
$^2$Max Planck Institute for Intelligent Systems, T{\"u}bingen, Germany
}\\
{\tt\small \{g.paschalidis, d.antic, d.tzionas\}@uva.nl
\quad 
romana.wilschut@gmail.com
\quad
otaheri@tue.mpg.de}
}

\begin{document}

\twocolumn[{%
\maketitle
\begin{center}
    \centering
    \captionsetup{type=figure}
    \includegraphics[trim=000mm 000mm 000mm 000mm, clip=true, width=1\textwidth]{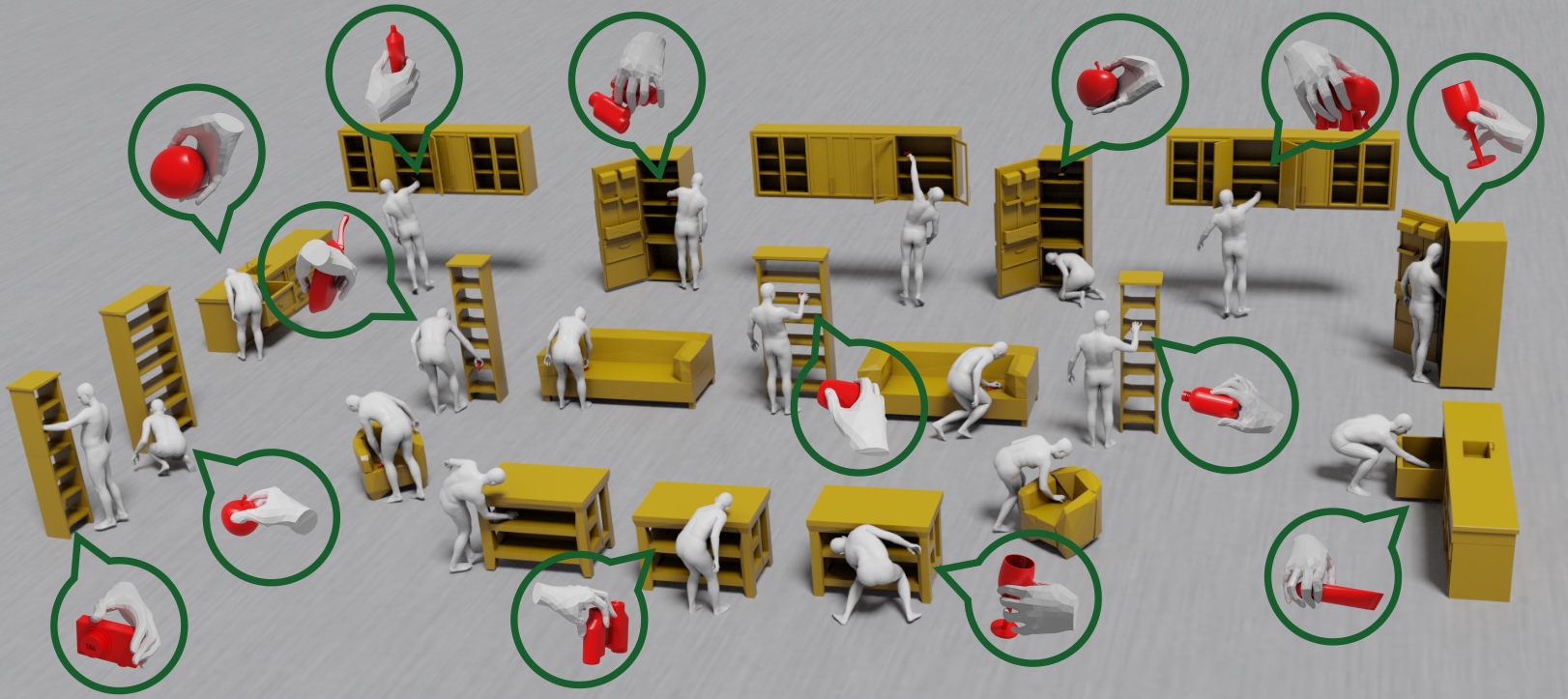}
    \vspace{-1.5 em}
    \captionof{figure}{
        We develop \cWGrasp, a novel framework for synthesizing 3D whole-body grasps for an object placed on a receptacle. 
        Our framework builds on a novel combination of geometric-based reasoning and controllable data-driven synthesis methods. 
        By adding a novel controllability in the synthesis process, we achieve realistic results at a fraction of the computational cost \wrt the state of the art \cite{flex}. 
    }
\end{center}%
\vspace{+1.2 em}
}]

\begin{abstract}
    Synthesizing \threeD whole bodies that realistically grasp objects is useful for animation, mixed reality, and robotics. 
    This is challenging, because the hands and body need to look natural \wrt each other, the grasped object, as well as the local scene (\ie, a receptacle supporting the object). 
    Moreover, training data for this task is really scarce, while capturing new data is expensive. 
    Recent work goes beyond finite datasets via a divide-and-conquer approach; 
    it first generates a ``guiding'' \rhand grasp, and then searches for bodies that match this. 
    However, the guiding-hand synthesis lacks controllability and receptacle awareness, so it likely has an implausible \direction (\ie, a body can't match this without penetrating the receptacle) and needs corrections through major post-processing. 
    Moreover, the body search needs exhaustive sampling and is expensive. 
    These are strong limitations. 
    We tackle these with a novel method called \cWGrasp. 
    Our key idea is that performing geometry-based reasoning ``early on,'' instead of ``too late,'' provides rich ``control'' signals for inference.
    To this end, \cWGrasp first samples a plausible reaching-\direction vector (used later for both the arm and hand) from a probabilistic model built via ray-casting from the object and collision checking. 
    Then, it generates a reaching body with a desired arm \direction, as well as a ``guiding'' grasping hand with a desired palm \direction that complies with the arm's one. 
    Eventually, \cWGrasp refines the body to match the ``guiding'' hand, while plausibly contacting the scene. 
    Notably, generating already-compatible ``parts'' greatly simplifies the ``whole''. 
    Moreover, \cWGrasp uniquely tackles both right- and left-hand grasps. 
    We evaluate on the \grab and \ReplicaGrasp datasets. 
    \cWGrasp outperforms baselines, at lower runtime and budget, while all components help performance. 
    Code and models are available at 
    \url{https://gpaschalidis.github.io/cwgrasp}.
\end{abstract}

\newcommand{\footnoteCoarseNet}{\footnote{\grabnet~\cite{grab} uses wrist translation and rotation only for training. 
For inference the only input is object shape, so grasps have a random direction.}}

\newcommand{\footnoteGNet}{\footnote{\gnet \cite{goal} takes as input only object shape and height.}} 


\section{Introduction}

Synthesizing virtual \threeD humans that grasp objects realistically is important for applications such as virtual assistants, animation, robotics, games, or synthetic image datasets. 
Importantly, this involves the whole body, so that the body approaches an object, arms reach it, and hands grasp it. 
But this is challenging; the body and hands should look natural and fully coordinated, the body should approach an object without penetrating the scene, the hands should dexterously contact the object. 
Due to these challenges, most of the existing work tackles only parts of the problem, namely
disembodied hands, or bodies with non-dexterous hands. 

To make matters worse, 
3D training data for whole-body grasps is very scarce. 
The recent 
\flex \cite{flex} method 
tackles data scarcity in
a divide-and-conquer way.
First, it generates a hand-only grasp through \grabnet~\cite{grab}. 
Then, this grasping hand guides a search for a plausible body. 
That is, many bodies are sampled in random poses and 
locations, and are optimized to match the guiding hand. 
However, there exists  
a key problem; the guiding hand has a random \direction that likely disagrees with the \direction bodies can approach from without penetrating receptacles. 
So, the guiding hand needs major corrections via post-processing. 
This produces promising results but needs exhaustive sampling (500 bodies), and is expensive (separate refinement per sample). 

We identify two main reasons for the above problems: (1)~Performing body- and receptacle-aware reasoning ``too late'', and 
(2)~\grabnet's total lack of controllability\footnoteCoarseNet. 
These are key limitations. 
We tackle these by developing \textbf{\cWGrasp} (``\cWGraspFULL''), a new method 
composed of
the following novel modules. 

\zheading{\reachingField model}
First, we
detect 
the \directions from which a body's arm and hand can reach an object without penetrating the receptacle supporting the object. 
Think of a mug lying on a shelf and emitting ``light''; some rays travel unblocked in free space, while other ones get blocked by shelf panels. Our key insight is that the ``well-lit'' space near the object reveals its reachability. 
So, we cast rays from the object, detect collisions with nearby receptacles,
and consider only the non-colliding rays  
for building a new probabilistic \threeD vector field, called \reachingField.

\newcommand{\severalColorrrs}{\textcolor{blue}{se}\textcolor{DarkGreen}{ve}\textcolor{orange}{ral} \textcolor{blue}{co}\textcolor{DarkGreen}{lo}\textcolor{orange}{rs}}

\begin{figure}[t]
    \vspace{-0.30 em}
    \centering
    \includegraphics[trim=000mm 000mm 000mm 000mm, clip=true, height=0.40 \linewidth]{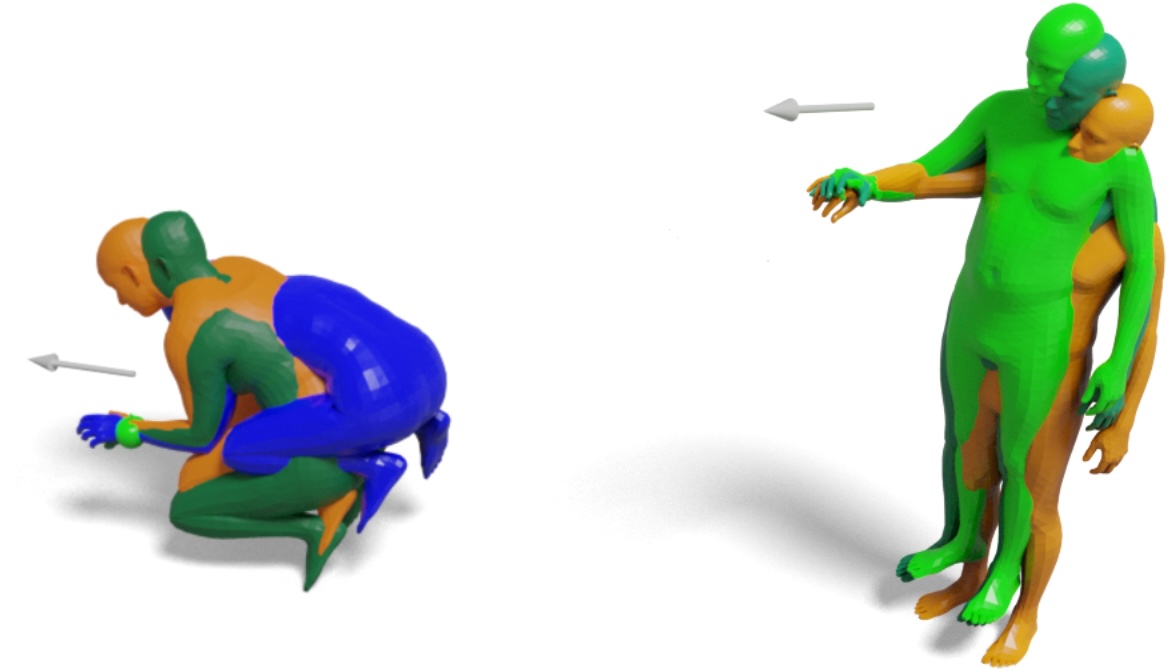}
    \vspace{-0.8 em}
    \caption{
                \textbf{Controllable reaching-body synthesis (\creach).} 
                We show examples 
                where multiple bodies (shown with \severalColorrrs) are generated to reach a target wrist
                location (shown as a \textcolor{DarkGreen}{green} sphere), while having 
                a \emph{desired \threeD~arm direction} 
                (\textcolor{gray}{gray arrow}). 
    }
    \label{fig:controlReachBody}
    \vspace{-0.5 em}
\end{figure}

\newcommand{\severalColors}{\textcolor{orange}{se}\textcolor{blue}{ve}\textcolor{myYellow}{ral} \textcolor{purple}{co}\textcolor{DarkGreen}{lo}\textcolor{orange}{rs}}

\begin{figure}[t]
    \centering
    \includegraphics[trim=000mm 000mm 000mm 000mm, clip=true, width=0.80 \linewidth]{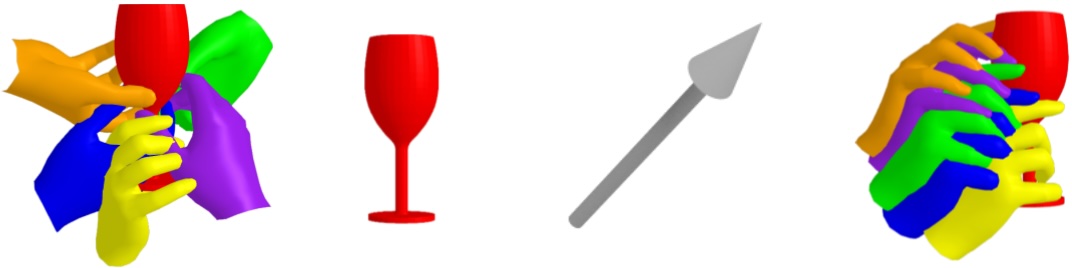}
    \vspace{-0.6 em}
    \caption{
                \textbf{Controllable hand-grasp synthesis.}
                The goal is to grasp the \textcolor{BrickRed}{red wineglass}. 
                \qheading{Left -- \grabnet~\cite{grab}} 
                Due to \grabnet's lack of controllability\textcolor{red}{$^1$}, sampling its latent space produces plausible grasps (shown with \severalColors) but with \emph{random \direction}. 
                \qheading{Right -- Our \cgrasp} 
                We add controllability, so drawing samples produces plausible and varied grasps
                (shown with \severalColors), that have a \emph{desired \threeD palm \direction} (shown with a \textcolor{gray}{gray arrow}).
    }
    \label{fig:controlGraspHand}
    \vspace{-0.5 em}
\end{figure}

Sampling the \reachingField provides a single 3D \direction vector that can be used as a ``control signal'' for 
the synthesis of both a reaching body and grasping hand. 
But 
existing synthesizers for this, 
such as \gnet~\cite{goal} for the body and \grabnet~\cite{grab} for the  hand, lack such controllability\textcolor{red}{$^{1,}$\footnoteGNet}. 
We resolve this with two novel modules, as follows. 

\zheading{\creach model}
We train a conditional variational autoencoder (\cVAE)
for producing a reaching \smplX~\cite{smplx} body. 
This goes beyond \gnet in three ways: 
(1)~It is conditioned not only on target object/wrist location, but, uniquely, also on a desired 
3D arm \direction; see \cref{fig:controlReachBody}. 
(2)~It is trained not only on \grab~\cite{grab} data, which has a limited range of target wrist (and object)
locations, but also on \circleData~\cite{circle} data that is richer for reaching body poses. 
(3)~It can generate both right- and left-arm reaching. 
We call the resulting model \creach for ``\ControllableReach.'' 

\zheading{\cgrasp model}
We train a \cVAE to generate a grasping \mano~\cite{mano} hand. 
This goes beyond \grabnet in two ways: 
(1)~It is conditioned not only on object shape\textcolor{red}{$^1$}, but  
also on a desired 3D palm \direction; see examples in \cref{fig:controlGraspHand}-right. 
(2)~It can generate both right- and left-hand grasps. 
We call the resulting model \cgrasp for ``\ControllableGrasp.''  

\zheading{\cWGrasp framework} 
We condition \emph{both} \creach and \cgrasp on the \emph{same}~\direction, produced by \reachingField. 
Crucially, 
this produces a reaching \smplX 
body (\creach) and a guiding \mano grasping hand (\cgrasp) that are 
\emph{already 
``compatible''} with each other, so they only need a small refinement to be ``put together.'' 
To this end, we conduct optimization \cite{Hassan2019prox, smplx, flex} that searches for the \smplX pose that lets \smplX's hand match the guiding \mano hand, 
while 
the body contacts the floor without penetrating the receptacle. 
Thanks to our controllable inference, we 
can sample
only $1$ body and hand from \creach and \cgrasp, respectively, in strong contrast to  \flex's $500$ different samples. 
This makes our framework roughly $16 \times$ faster. 

We evaluate on the \grab~\cite{grab} and \ReplicaGrasp~\cite{flex} datasets. 
Both \cgrasp and \creach accurately preserve a specified palm and arm \direction, respectively. 
Importantly, adding controllability does not harm; \cgrasp performs 
on par with three 
baselines 
\cite{grab,contactgen,dexgraspnet} 
while 
being able to control 
palm \direction  . 
Last, 
our \cWGrasp method outperforms \flex~\cite{flex} in almost all metrics, 
while 
its 
generated 
whole-body grasps 
are perceived as more realistic. 

In summary, here we make four main contributions:
\begin{enumerate}
    \item 
    The \emph{\reachingField} model that generates \threeD~\directions for 
    reaching a \threeD object, 
    helping as a control signal. 
    \item 
    The \emph{\creach} model that generates a \smplX body reaching objects with a desired (right/left) arm \direction. 
    \item 
    The \emph{\cgrasp} model that generates a (right/left) \mano hand grasping an object with a desired palm \direction. 
    \item 
    The novel \emph{\cWGrasp} method that combines the above for generating dexterous \smplX grasps for an object lying on a receptacle. 
    This is 16x faster than a \mbox{SotA} baseline, and uniquely tackles both right- and left-hand grasps.
\end{enumerate}

\section{Related Work}

\subsection{Hand-only Grasps}

Early research focused on modeling \cite{modeling_grips, shape_primitives} and 
classifying \cite{grasp_choice, grasp_taxonomy} grasps. 
Then, research focused on generating grasps for robot \cite{robotics_review, robotics_survey} and human hands \cite{contactdb, contactgrasp, grab}. 

\zheading{Hand models}
Some work models hand shape explicitly with \threeD meshes \cite{hand_primitives, hand_meshes} with 
statistical models \cite{mano,nimble} being popular.
Other work uses implicit shape, such as 
3D distance fields \cite{grasping_field, lisa} or a sum of 3D Gaussians \cite{hand_gaussians}. 
Here we use \mano~\cite{mano}~due its wide user base, and because it lets us compute accurate contacts and penetrations.

\zheading{Data} 
Many datasets have been captured with 
single- \cite{contactpose, dexYCB, fhpa, honnotate, freihand, hoi4d} or two-hand \cite{h2o, h2o-3d} images. 
Recent work captures whole-body meshes \cite{smplx} grasping rigid objects \cite{grab}, or articulated objects while also containing RGB images \cite{arctic}. 
\hoidiffusion~\cite{hoidiffusion} uses a diffusion model for generating synthetic hand-object images conditioned on \threeD hand-object grasps produced by \grabnet~\cite{grab}. 
\dexgraspnet~\cite{dexgraspnet} builds a large dataset by applying an optimization framework on \threeD objects, leveraging a differentiable force closure estimator and energy functions. 
Here we extend the \grabnet~\cite{grab} model and use its \grab~\cite{grab} dataset to train our model to facilitate fair comparisons.

\zheading{Contact}
\mbox{ContactGrasp}~\cite{contactgrasp} uses real contact maps from the \mbox{ContactDB}~\cite{contactdb} dataset to infer a grasping hand pose, given a posed object mesh. 
\mbox{ContactOpt}~\cite{contactopt} infers likely hand-object contacts and optimizes hand pose to match these. 
\mbox{GraspTTA}~\cite{jiang2021handobjectcc} infers an initial grasp for an object point cloud, and optimizes it to match a target contact map. 
\mbox{Grasp'D}~\cite{turpin2022graspddc} takes a hand, an object as a 
point cloud and as a SDF, and generates grasps via optimization on contact forces.
\contactgen~\cite{contactgen} learns an object-conditioned joint distribution of a contact-, part- and direction-map, 
exploiting 
the direction of contact at a low level 
for synthesis. 
\grabnet~\cite{grab} infers an initial grasp for a BPS-encoded~\cite{bps} object and 
refines it with a neural net that considers a per-vertex contact likelihood. 
\grabnet lacks controllability, so it produces grasps with random directions. 
Here we extend \grabnet by adding the missing controllability; 
only our and 
concurrent work \cite{graspxl} 
condition on 
the palm's direction. 

\zheading{Grasps from images}
\mbox{ObMan}~\cite{hasson2019obman} infers hand and object meshes from a color image, while \mbox{H+O}~\cite{h+o} infers keypoints. 
\mbox{GanHand}~\cite{ganhand} infers object pose and grasp type with a rough hand pose \cite{grasp_taxonomy}, refining it via contact constraints. 
\mbox{TOCH}~\cite{toch} does a refinement using a 3D SDF. 
More recent work tackles grasps with unknown objects from color video \cite{fan2024hold, swamy2023showme}.
For a more detailed overview please see \cite{fan2024challenge}

\zheading{Motion generation}
\mbox{D-Grasp}~\cite{dgrasp} learns hand-object interaction via RL; 
the task is to grasp and move a given object to a goal pose. 
\manipnet~\cite{manipnet} generates hand-object interaction (\HOI) motions for single or both hands, using spatial features. 
\geneohdiffusion~\cite{geneoh} denoises \HOI motion via diffusion, and a hand-keypoint trajectory representation. 
\grip~\cite{taheri_3dv2024_grip} and \gears~\cite{gears} synthesize interacting finger motion from given hand and object trajectories. 
Concurrently to us, \graspxl~\cite{graspxl} 
generates grasping motions via RL (without using pre-captured \HOI data) 
while 
conditioning on the palm direction, as we do for static synthesis. 

\subsection{Whole-body Grasps}
The shape representation used for body models ranges 
from cylinders \cite{body_cylinders} and super-quadrics \cite{superquadratics1996} to 
mesh-based statistical \threeD models \cite{allen_learning, scape, smpl, smplx, xu2020ghum, SUPR:2022}. 
We use the \smplX \cite{smplx}~statistical model that is widely used for interactions.

\zheading{Interacting with scenes} 
Wang \etal~\cite{wang2021synthesizing} first infer intermediate key poses and then generate 
in-between motions. 
\mbox{SAMP}~\cite{samp} and NSM~\cite{nsm} infer several goal locations and orientations on target objects, 
(stochastically and deterministically, respectively), 
and then infer in-between motion. 
Given a 
body pose and chair mesh, COUCH~\cite{couch} infers diverse contacts on the chair, 
and 
body poses that match these. 

\zheading{Static grasps} 
\flex~\cite{flex} generates \smplX grasps, by optimizing the body to match a guiding hand-grasp inferred via \grabnet~\cite{grab}. 
Our \cWGrasp method is inspired by this, but is more efficient thanks to its controllable inference.

\zheading{Dynamic grasps} 
\mbox{CIRCLE}~\cite{circle} and \mbox{WANDR}~\cite{diomataris2024wandr} infer (short- and long-term, respectively) motion for reaching a target wrist location. \goal~\cite{goal} infers a static target body grasp via interaction-aware features, and infers motion to the goal. \mbox{SAGA}~\cite{saga} generates such motions stochastically. \mbox{IMoS}~\cite{imos} infers guiding arm-only motions that ``drive'' the whole body. 
Given object trajectories, \mbox{OMOMO}~\cite{omomo} uses a conditional denoising diffusion model to generate 
wrist
joint positions for each object state, and then conditions on these to generate a full body with non-articulated hands.

\section{Method}
\begin{figure*}
    \centering
    \vspace{-1.5 em}
    \includegraphics[trim=000mm 000mm 000mm 000mm, clip=true, width=0.98 \textwidth]{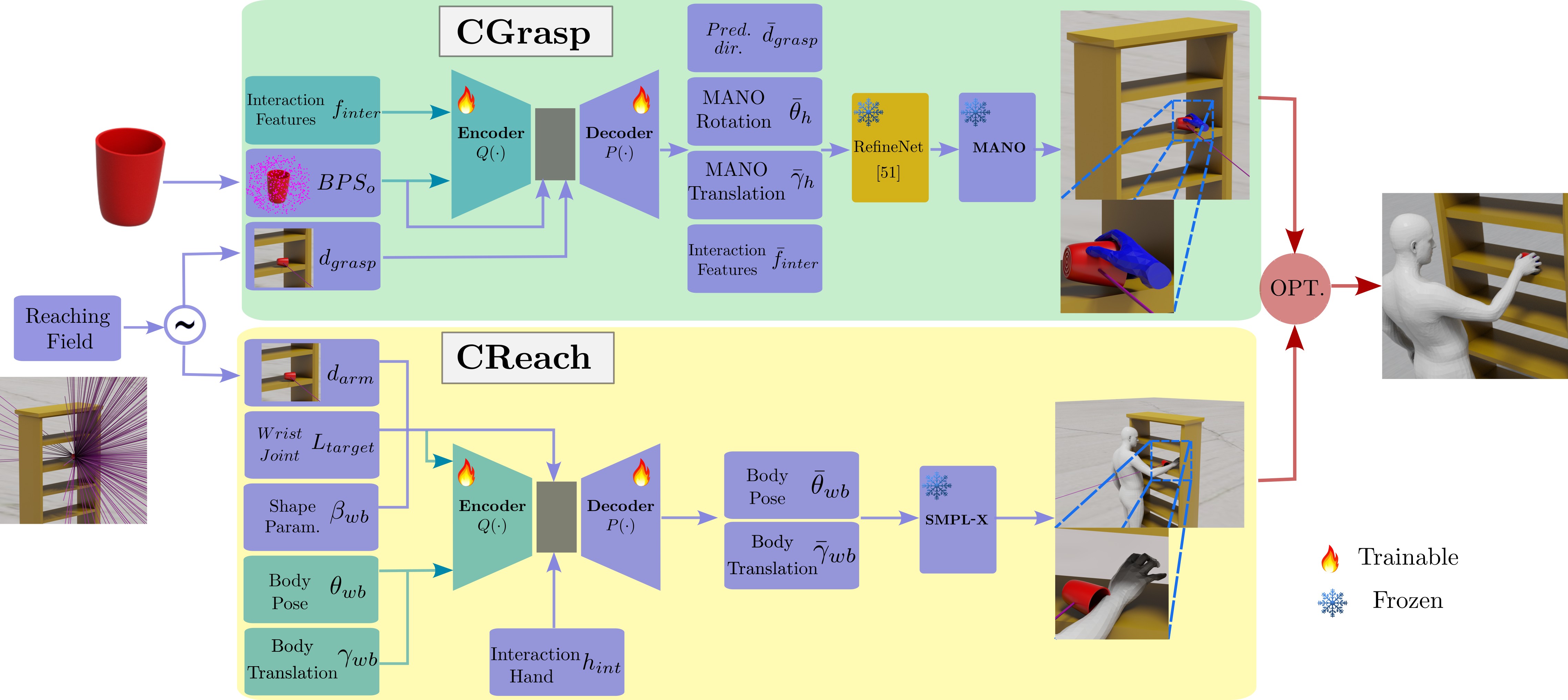}
    \vspace{-0.6 em}
    \caption{
            \textbf{\cWGrasp~framework}.
            We first sample a single
            reaching direction from \reachingField. 
            Next, we condition both \cgrasp and \creach on the same direction and obtain a guiding hand grasp (shown in \textcolor{blue}{blue}) and a reaching body (shown in \textcolor{gray}{gray}), respectively, that satisfy the sampled direction, so they are ``compatible'' with each other. 
            Finally, 
            an optimization stage 
            refines the body to match the guiding hand 
            while 
            resolving 
            penetrations with the 
            object and/or receptacle. 
            Note that our framework 
            can generate both left- and \rhand grasps. 
            Parts 
            in \textcolor{DarkPurple}{purple} are used for both training and inference, 
            in \textcolor{DarkGreen}{green} only for training, 
            in \textcolor{olive}{brown} only for inference, and 
            in \textcolor{BrickRed}{red} for optimization.
    }
    \label{fig:pipeline_architecture}
    \vspace{-0.5 em}
\end{figure*}

We build \cWGrasp, a novel 
framework (\cref{fig:pipeline_architecture}) that 
generates a whole-body grasp, given an object on a receptacle.
To this end, we 
develop \reachingField (\cref{sec:reachingField}), a novel 
model 
that generates a likely 
reaching 
\direction. 
We 
condition on the same \direction two novel 
models for producing a reaching body (\creach, \cref{sec:CReachControl}) 
and hand grasp (\cgrasp, \cref{sec:CGraspControl}). 
We combine all these in \cWGrasp (\cref{sec:CWGrasp}). 

\subsection{Preliminaries}

\hspace{\parindent}
\zheading{Hand model (used in \cgrasp)}
We use \mano \cite{mano}, 
a differentiable function $\mesh_\hand(\beta_\hand, \theta_\hand, \trans_\hand)$ 
parameterized by 
translation,    $\trans_\hand   \in \mathbb{R}^3$, 
shape,          $\beta_\hand    \in \mathbb{R}^\betasNumb$, and 
pose,           $\theta_\hand$. 
The output is a \threeD mesh, $\mesh_\hand$, 
rigged with a skeleton of $16$ joints; 1 for the wrist and 15 for fingers. 
The pose $\theta_\hand \in \mathbb{R}^{16 \times 3}$ is encoded 
as axis-angle rotations; 
the global rotation (first 3 parameters) is $\theta_\hand^\wrist \in \mathbb{R}^3$.
The shape parameters $\beta_\hand$ live in a low-dimensional linear space. 

\zheading{Whole-body model (used in \creach, \cWGrasp)} 
We use the \smplX \cite{smplx}~model, 
a differentiable function $\mesh_\bodyWhole(\beta_\bodyWhole, \theta_\bodyWhole, \trans_\bodyWhole)$ 
parameterized by 
shape,          $\beta_\bodyWhole       \in \mathbb{R}^\betasNumb$, 
pose,           $\theta_\bodyWhole$,    and 
translation,    $\trans_\bodyWhole      \in \mathbb{R}^3$; 
here we ignore facial parameters. 
The output is a \threeD mesh, $\mesh_\bodyWhole$, rigged with a skeleton of $22$ body joints and $15$ joints per hand. 
The pose $\theta=(\theta_\bodyMain, \theta_\hand)$ consists of $\theta_\bodyMain \in \mathbb{R}^{22 \times 3}$ for the body and $\theta_\hand \in \mathbb{R}^{2 \times 15 \times 3}$ for hands as axis-angle rotations. 
The shape parameters $\beta_\bodyWhole$ live in a low-dimensional linear space.

\zheading{\cnet~-- part of \grabnet \cite{grab}}
\grabnet generates \threeD \mano grasps for a given object, and consists of:
\cnet, for producing an initial grasp, 
and \rnet, for refining it. 
Here we focus only on grasp generation, 
so
we build on \textbf{\cnet}. 
This is modeled as a VAE; given an object shape represented with Basis Point Sets~\cite{bps}, $BPS_o$, a wrist 
rotation,       $\theta_\hand^\wrist$, and 
translation,    $\trans_\hand$, 
the encoder     $Q$ generates a latent code $Z \in \mathbb{R}^{16}$, 
namely:         $Q(Z|\theta_\hand^\wrist, \trans_\hand, BPS_o)$. 
The decoder maps this, concatenated with the object shape, $BPS_o$, to an estimated \mano 
translation,    $\bar\trans_\hand \in \mathds{R}^3$, and 
joint angles,   $\bar\theta_\hand \in \mathds{R}^{16\times6}$, 
\ie:         $P(\bar\theta_\hand, \bar\gamma_\hand | Z, BPS_o)$.
To train \cnet we use both its encoder and decoder, 
and use 5 losses:
$\mathcal{L}_{KL}$, 
$\mathcal{L}_{edge}$, 
$\mathcal{L}_{vertex}$,
$\mathcal{L}_{d_{o2h}}$, 
$\mathcal{L}_{d_{h2o}}$;
for details see \cite{grab}. 
In test time, we use only the decoder conditioned on object shape, $BPS_o$; there is no other input. 
Thus,  
sampling 
different latent codes produces grasping hands with a random direction; see \cref{fig:controlGraspHand}.

\zheading{\gnet~-- part of  \goal \cite{goal}}
\gnet generates a \smplX grasping body for a given object shape and location. 
\gnet is modeled with a VAE, like \cnet, 
so it has an
encoder, $Q(Z|\theta_\bodyWhole, \trans_\bodyWhole, \gnetTarget)$, and
decoder, $P(\bar\theta_\bodyWhole, \bar\trans_\bodyWhole | Z, \gnetTarget)$, 
where 
$Z$                     is the latent code, 
$\theta_\bodyWhole$     is body pose, 
$\gamma_\bodyWhole$     is translation. 
$\gnetTarget$           is a target condition comprising the object's shape, $BPS_o$, and its centroid height. 
For details see \cite{goal}.

\subsection{\reachingField~-- Reaching Direction} 
\label{sec:reachingField}
\begin{figure}
    \centering
    \begin{subfigure}[b]{0.23\textwidth}
        \centering  
    \includegraphics[trim=000mm 120mm 000mm 100mm, clip=true, width=0.99\textwidth]{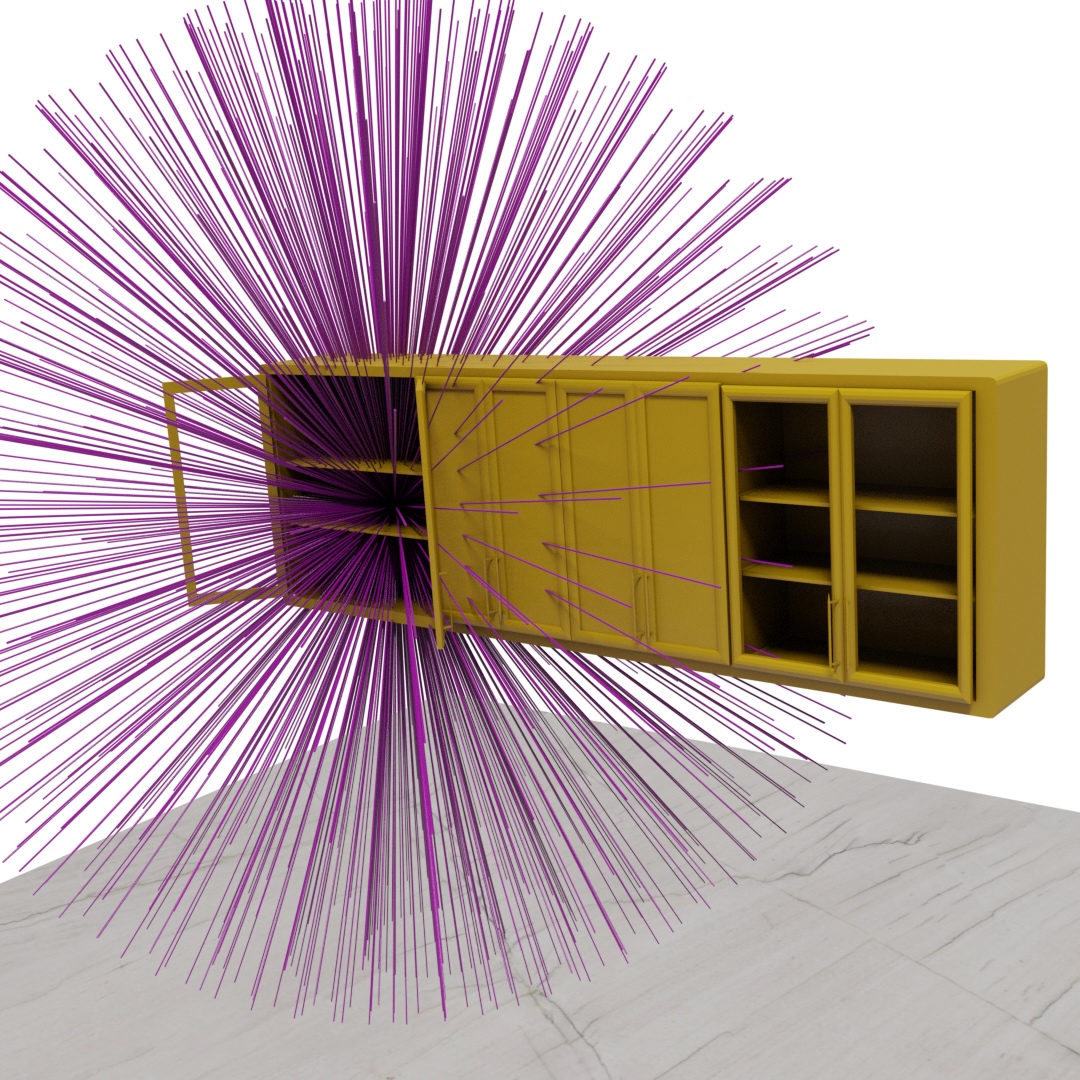}
    \end{subfigure}
    \begin{subfigure}[b]{0.23\textwidth}
        \centering  
    \includegraphics[trim=000mm 120mm 000mm 100mm, clip=true, width=0.99\textwidth]{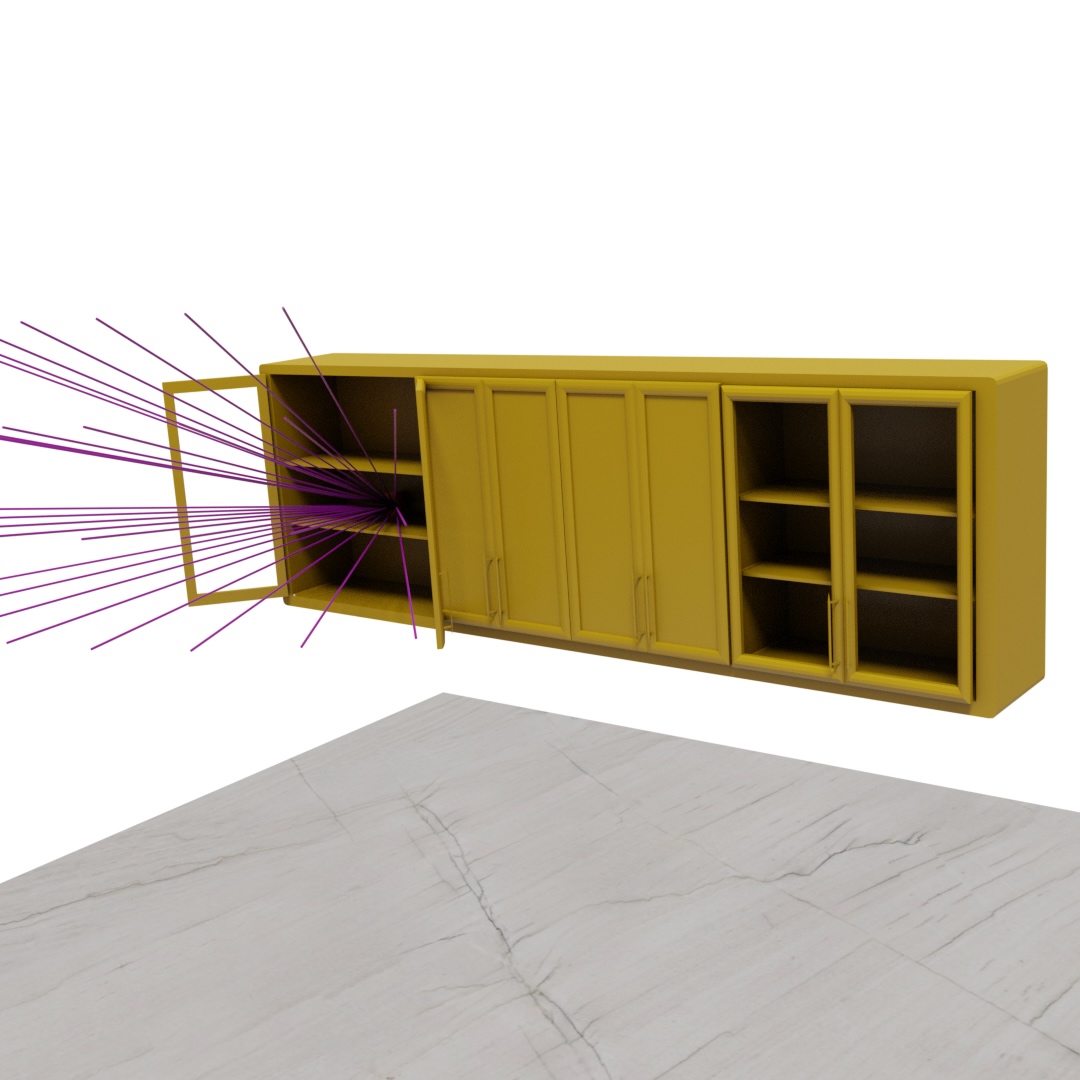}
    \end{subfigure}
     \vspace{-0.5 em}
    \caption{
        \textbf{Arm/hand direction (\cref{sec:reachingField}, Filter \#1)}. 
        \textbf{Left:} 
        We cast rays from the object to surrounding space. 
        \textbf{Right:} 
        We 
        prune rays intersecting with a receptacle and  
        keep non-intersecting ones; 
        the latter 
        represent \directions an arm/hand can reach the object from.
    }
    \label{fig:reaching_field}
\end{figure}

Given an object on a receptacle, 
we build \reachingField, a novel
probabilistic 3D vector field
of \directions the object can be reached by a body
(see~\cref{fig:reaching_field}). 
To this end, we cast 3D rays from the object to surrounding space, check for collisions with the receptacle, filter out colliding ones (considering also the arm's volume and standing on the ground), and assign probabilities to remaining rays, as follows.

\zheading{Ray casting}
Let $\object$ be a 3D mesh for the object, and 
$\objectcentroid \in \mathbb{R}^3$ 
be its centroid.  
We sample uniformly a (unit) sphere centered at $\objectcentroid$, 
constructing 
a spherical point grid 
$\mathbf{S} = \{ \gridpoint\}$. 
Then, we cast rays $\ray$ 
going from $\objectcentroid$ 
through each point $s_i$. 

\zheading{Ray filtering}
Let $\receptacle$ be a 3D mesh for the receptacle. 
We evaluate and filter the casted rays $\ray$ with the following. 
 
\qheading{{Filter \#1. Arm/hand direction} (\cref{fig:reaching_field})}
We traverse each ray $\ray$ and 
evaluate whether it intersects 
with $\receptacle$. 
Intersecting rays are pruned, as they represent a \direction from which an arm or hand would ``directly'' penetrate the receptacle.

\begin{figure}
    \vspace{-0.5 em}
    \centering      
        \includegraphics[trim=000mm 020mm 000mm 030mm, clip=true, width=0.90 \columnwidth]{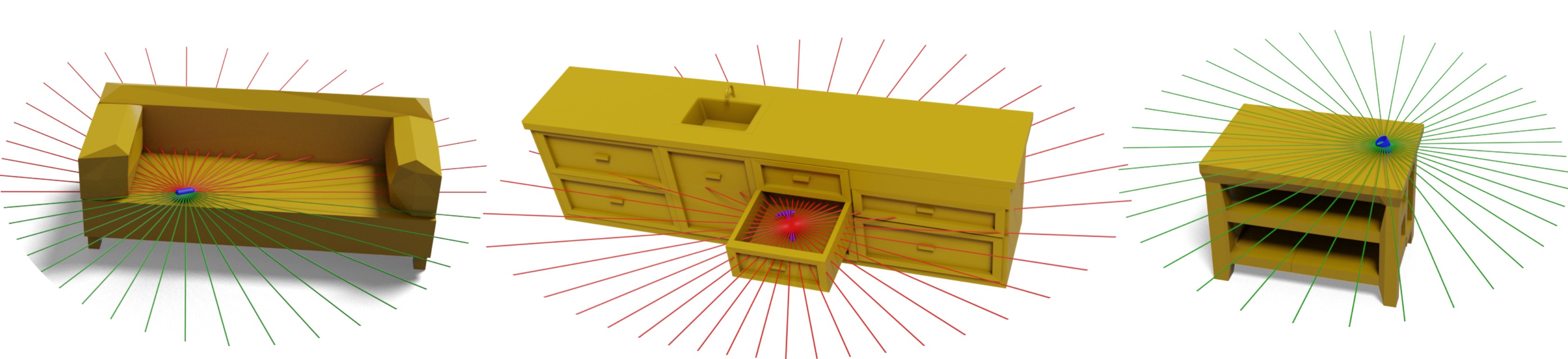}
     \vspace{-0.5 em}
    \caption{
    \textbf{
    Body orientation (\cref{sec:reachingField}), Filter \#2.} 
    We project the curated rays parallel to the ground 
    and detect whether any 
    receptacle 
    parts  
    hinder a body from approaching the object from certain \directions; 
    the \textcolor{red}{red} rays  
    are discarded,
    while \textcolor{green}{green}
    ones 
    are kept.   
    }
    \label{fig:secondary_projection}
    \vspace{-0.5 em}
\end{figure}

\qheading{{Filter \#2 - Body orientation} (\cref{fig:secondary_projection})}
To (optionally) save computational resources (on the expense of pruning some plausible directions), we
project the curated rays onto a horizontal plane parallel to the ground, and detect further intersections with $\receptacle$. 
Intersecting rays denote directions that hinder a body from ``easily'' approaching the object. 
However, in case all rays intersect, \eg when the object is inside a box or drawer, then this step is disregarded altogether.

\qheading{{Filter \#3 - Standing places}} 
To grasp an object, a body needs to stand at a nearby place on the ground without penetrating 
any ``occluders.'' 
To find such places, we traverse the curated rays $\ray$, and at regular intervals (every 30 cm) we
cast vertical rays $\rayvertical$ 
and check whether these collide with 
$\receptacle$ or other ``occluders'' hindering a body from standing. 
In case of collision we prune the ``parent'' ray $\ray$ altogether. 

\qheading{{Filter \#4 - Wiggle room for arm volume}}
The above steps ``detect'' plausible 
rough body positions
and 
arm \directions. 
However, they ignore that a body has a certain \emph{volume}, so 
its vertices can still
penetrate the receptacle. 
To resolve this, we ``swipe'' all projected filtered rays within a small range around the vertical axis, 
and discard those intersecting $\receptacle$.

\begin{figure}
    \vspace{-0.5 em}
    \centering      
        \includegraphics[trim=000mm 010mm 000mm 010mm, clip=true, width=0.65 \columnwidth]{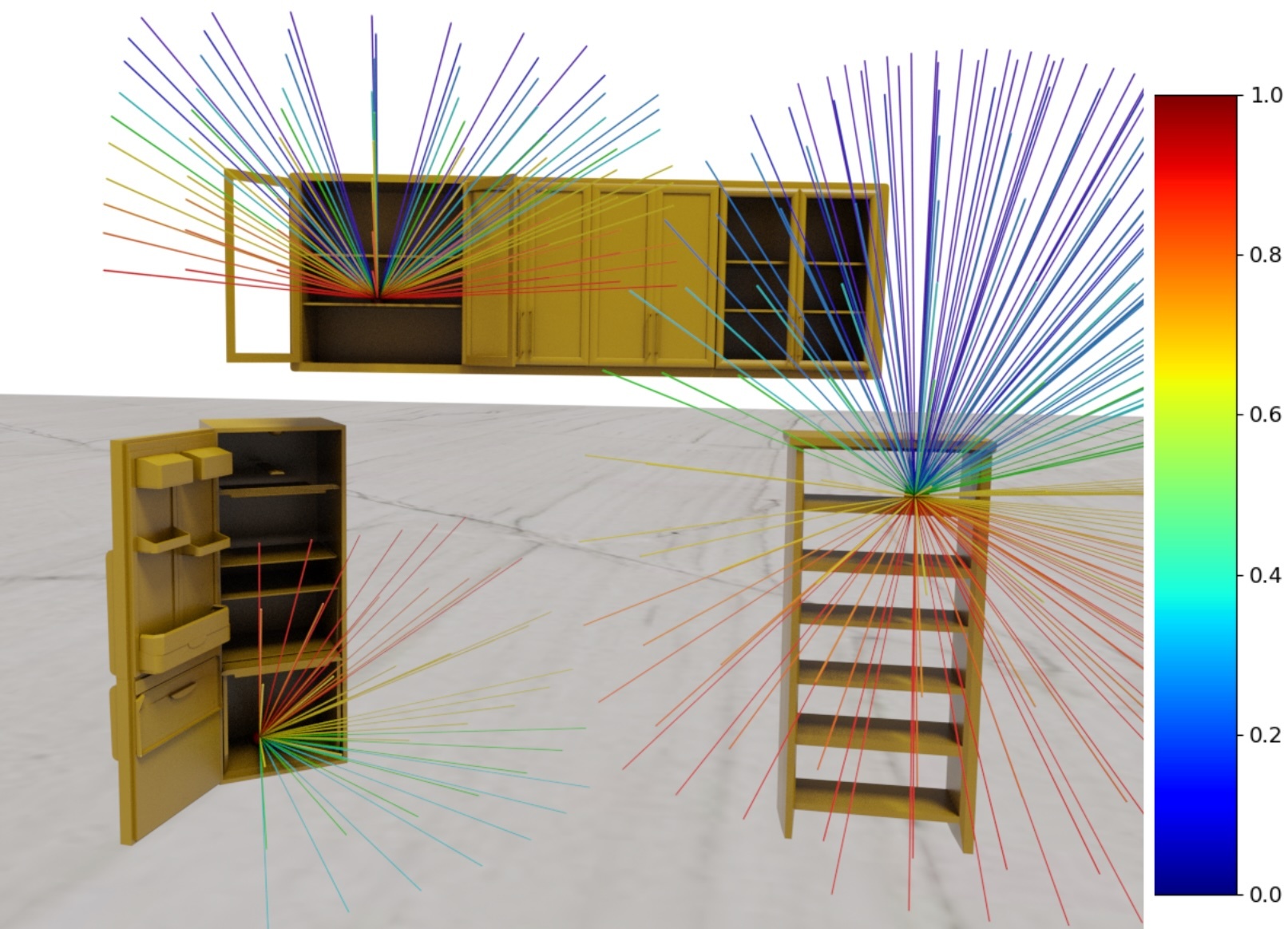}
     \vspace{-0.5 em}
    \caption{
    \textbf{\reachingField~-- Ray likelihood (\cref{sec:reachingField}, \cref{eq:reaching_field})}, 
        shown with color-coding; \textcolor{BrickRed}{red} shows high and \textcolor{blue}{blue} low likelihood. 
        Objects near the ground are likely grasped from above (left). 
        Objects high above the ground are likely grasped from below (right).
    }
    \label{fig:ray_probability}
\end{figure}

\zheading{\reachingField}
The curated rays are plausible reaching \directions.
But not all \directions are equally likely.  
When changing a light bulb on the ceiling, our hand likely approaches it from below, while when tying shoelaces, it approaches from above. 
Thus, likelihood depends 
on how high above the ground an object lies
and is defined as: 
\vspace{-0.3 em}
\begin{align}\label{eq:reaching_field}
    p_i &= \frac{\exp \left( -\nicefrac{1}{  \left(  s_ia_i  \right)  } \right) } 
       {\sum_{i} \exp \left( -\nicefrac{1}{  \left(  s_ia_i  \right)  } \right) }      \text{,}  
\end{align}
where  
$a_i$ is the 
smallest  
angle of ray $\ray$ \wrt the vertical axis $z$, 
while $s_i = -1$ when 
the object height is $\ge$0.7m above ground 
and $r_i$ is directed downward, or 
the height is $<$0.7m and $r_i$ is directed upward. Else, $s_i = 1$.
See likelihood examples in \cref{fig:ray_probability}. 
For details see \supmat~
(\cref{sec:supmat_reaching_field}).

\zheading{Inference}
\reachingField is probabilistic, so sampling it
produces a plausible \threeD reaching \direction. 
Note that objects can be reached from multiple directions; drawing different samples accounts for this.


\subsection{\creach~-- Controllable Reaching Bodies} 
\label{sec:CReachControl}

Our goal is \emph{controllable} synthesis of a \smplX body ``reaching'' an object. 
We do this by extending \gnet~\cite{goal} with a \emph{condition} on \emph{arm \direction};  
see \cref{fig:pipeline_architecture} bottom. 

\zheading{Formulation}
The \emph{\direction} from which a body-arm approaches objects is key for grasping. 
We provide this to \creach as a normalized vector, $\armDir \in \mathds{R}^3$. 
Generated bodies should have an arm \direction that aligns with this, so 
we compute \smplX's normalized elbow-to-wrist vector. 

\zheading{Training}
We use \circleData~\cite{circle} and \grab~\cite{grab} data for training;  
crucially, 
the former 
has 
a rich range of target wrist locations.
We use the \direction, $\armDir$, 
as condition for both encoder $Q(Z|\theta_\bodyWhole, \trans_\bodyWhole, \betas_\bodyWhole, \creachTarget, \armDir)$ and decoder $P(\bar\theta_\bodyWhole, \bar\trans_\bodyWhole | Z, \betas_\bodyWhole, \creachTarget, \armDir,\inthand)$, where 
$Z$                     is the latent code,
$\theta_\bodyWhole$     is body pose, 
$\gamma_\bodyWhole$     is translation,
$\betas_\bodyWhole$     is shape, 
$\armDir$               is the desired arm \direction (new over \gnet), 
$\creachTarget$         is the target GT wrist joint 
(as a proxy for object centroid, as \circleData has no objects), 
and 
$\inthand$ 
denotes using the  
right ($\inthand=0$) or left arm ($\inthand=1$). 
We add (on top of \gnet losses) a loss 
on arm \direction 
as follows, where $w_{\armDir} = 5$: 
\begin{equation}
    \mathcal{L}_{\armDir} = w_{\armDir} \cdot \mathbb{E} \Big[ | {\armDir} - {\bar{d}_\text{arm}} | \Big]
    \text{.}
    \label{eq:lossDirectionCReach}
\end{equation} 

\zheading{Inference} 
The decoder takes the arm \direction, $\armDir$ (from \reachingField), the ``target'' object centroid, $\creachTarget$ (in training we approximate this with the wrist), and parameters $\betas_\bodyWhole$ and $\inthand$, and outputs a \smplX body; see~\cref{fig:controlReachBody}.

\subsection{\cgrasp~-- Controllable Grasping Hands} 
\label{sec:CGraspControl}

Our goal is \emph{controllable} synthesis;
we build \emph{\cgrasp} by extending \grabnet~\cite{grab} with a \emph{condition} on \emph{palm \direction}. 

\zheading{Formulation}
The \direction a hand grasps from is key. 
We provide this to \cgrasp as a unit vector, $\graspDir \in \mathds{R}^3$. 
All generated hands need to have a palm \direction that agrees with 
$\graspDir$. 
To this end, we annotate (offline) two vertices on the outer palm of \mano, as it is quasi-rigid so vertices stay consistent during motion. 
These vertices define $\graspDir$.

Moreover, we 
enhance the \emph{spatial awareness} of \cgrasp. 
Inspired by \gnet \cite{goal} and others \cite{arctic, manipnet}, 
we compute \threeD hand-to-object \IAfeatures vectors, $\ifeat \in \mathbb{R}^{99\times3}$.
In detail, we sample (offline) 99 ``interaction'' vertices, 
$\vertex_{\hand, i}^\inter \text{,~} i \in \{1, \dots, 99\}$,  
evenly distributed across \mano's inner-palm/fingersurface. 
Then, we compute \threeD vectors, $\ifeat$, encoding the distance and direction from the sampled hand vertices, $\vertex_{\hand}^\inter$, to their closest object ones, $\vertex_\obj'$. 

\zheading{Training}
We train on the \grab~\cite{grab} dataset. 
During training, we add the 
GT \IAfeatures, $\ifeat$, 
to the encoder $Q(Z| BPS_o, \ifeat)$ 
In test time, the decoder $P(\bar\theta_\hand, \bar\gamma_\hand, \bar{f}_\text{inter} | Z, BPS_o,\graspDir)$ 
predicts \mano parameters, ($\bar\theta_\hand, \bar\gamma_\hand$), and the \IAfeatures, $\bar{f}_\text{inter}$. 
$Z$     is the latent code, and
$BPS_o$ is the object shape.
We also add (on top of \grabnet losses) a loss on \direction and on \IAfeatures:
\begin{align}
    \mathcal{L}_\text{grasp} 
    &= (1 - c_{KL}) \cdot \mathbb{E} \Big[ | {\graspDir} - {\bar{d}_\text{grasp}} | \Big]
    \text{,}
    \label{eq:lossDirectionCGrasp}
    \\
    \mathcal{L}_\text{inter} 
    &= (1 - c_{KL}) \cdot \mathbb{E} \Big[ | \ifeat - \bar{f}_\text{inter} | \Big]  \text{,}
\end{align}%
where $c_{KL} = 0.005$ is a KL-divergence constant.  

\zheading{Inference}
The decoder takes the desired grasp \direction, $\graspDir$ (sampled from \reachingField), concatenated with the object shape,
$BPS_o$, and outputs a \mano grasp. 
For inference we append a frozen pretrained \rnet~\cite{grab}.

\subsection{\cWGrasp~-- Whole-Body Synthesis}
\label{sec:CWGrasp}

Given a \threeD object lying on a receptacle, 
we aim to generate a 
dexterous and physically-plausible \smplX body grasp. 

\zheading{Objective function}
We build the objective function: 
\begin{equation}\label{opt_fun}
\begin{aligned}
    \mathcal{L}_\text{opt} = 
    & 
    \lambda_{hm}\mathcal{L}_{hm} + 
    \lambda_{\theta} \mathcal{L}_{\theta} + 
    \lambda_{g}\mathcal{L}_{g} + 
    \\
    & 
    \lambda_{grd} \mathcal{L}_{grd} + 
    \lambda_p \mathcal{L}_{p} + 
    \lambda_{reg}\mathcal{L}_{reg}    
    \text{,}
\end{aligned}
\end{equation}
consisting of 
a hand-matching term, $\mathcal{L}_{hm}$, 
a body pose term,                    $\mathcal{L}_{\theta}$, 
a head-direction term,               $\mathcal{L}_{g}$ (often called ``gaze''), 
a ground-body penetration term,      $\mathcal{L}_{grd}$, 
a receptacle-body penetration term,  $\mathcal{L}_{p}$, and 
a regularizer term,                  $\mathcal{L}_{reg}$.
These terms are similar to FLEX~\cite{flex}, except for $\mathcal{L}_{grd}$ and $\mathcal{L}_{reg}$. 
For details on our loss terms, see \supmat~
(\cref{sec:supmat_optimization}). 

\zheading{Search space}
We operate in the original search space \cite{Hassan2019prox, smplx} for flexibility.  
This contrasts to \flex~\cite{flex} that uses a compact ``black-box'' latent space but loses some control. 
Even if \creach generates a body from a desired approaching direction, 
sometimes
the body penetrates the receptacle (see \cref{fig:creach_penetrations}-left). 
Starting the optimization from 
such a local minimum might trap the optimizer. 
To prevent this, we first translate the body by 1m along the 
the floor-projected \direction used to condition \creach, so we free it from big penetrations (see \cref{fig:creach_penetrations}-middle). 
Then, optimization (\cref{opt_fun}) pulls the body back to the object while refining body and finger pose (see \cref{fig:creach_penetrations}-right). 
This makes \cWGrasp robust.

\zheading{Optimizer}
We use Adam;  
for 1 body and for 1500 iterations it takes 
$\sim20$ sec on an Nvidia RTX 4500-Ada GPU.

\zheading{Sample efficiency}
We sample from \reachingField just one \direction and 
condition on it both \creach and \cgrasp. 
Thus, our reaching body and guiding hand are already compatible, 
and refine only the body to match the (fixed) hand. 
Instead, \flex~\cite{flex} samples 500 bodies, and refines both bodies and guiding hands, due to using the non-controllable \grabnet. 
Therefore, our method is very sample efficient.

\zheading{Left-hand interaction for whole bodies}
\cWGrasp uniquely generates both right- and left-hand whole-body grasps. 
For the latter, conditioning \creach with $\inthand=1$ (see \cref{sec:CReachControl}) produces a body that reaches the object with its left arm. 
Then, we mirror both the object and \reachingField's \direction (\wrt the object's sagittal plane), generate a right hand grasp with \creach, and mirror back the hand and object. 
Last, we run \cWGrasp optimization.

\section{Experiments}

\begin{figure*}
    \centering
    \includegraphics[width=0.99 \linewidth]{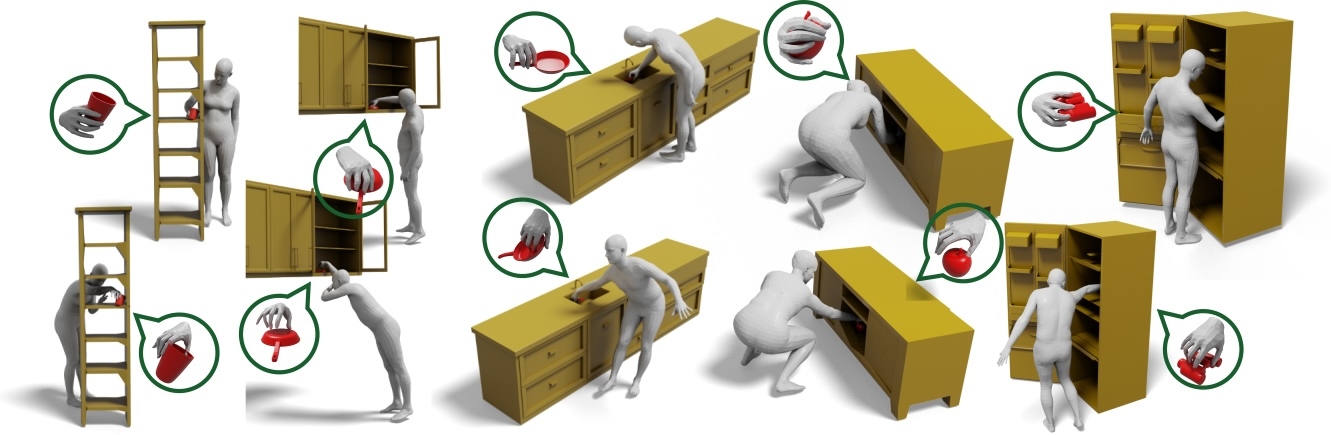}
    \vspace{-0.5 em}
    \caption{
            \textbf{Whole-body grasps} 
            produced by \cWGrasp (top row) and \flex~\cite{flex} (bottom). 
            \flex~samples 500 initial bodies and produces $10$ ones; 
            we show the smallest-loss one. 
            Our \cWGrasp~samples only 1 body 
            and also generates one, 
            yet it produces more realistic grasps.
    }
    \label{fig:cwgrasp_qualitative}
\end{figure*}

\subsection{Conditioning for \creach \& \cgrasp}
\label{sec:CReachCGraspacc}

We evaluate how accurately \cgrasp and \creach preserve their conditioning, \ie, the desired arm and palm \direction. 
\Cref{tab:creach_cgrasp_accuracy} reports results  (incl.~runtime) computed as follows.

\zheading{\cgrasp}
This is conditioned on a palm \direction vector. 
We extract all hand \directions from \grab's \cite{grab} test set and cluster them into $200$ centers using \mbox{K-Means}. 
We then use \grab's $6$ test objects and generate for each of these $2000$ grasps; 
to this end, we run \cgrasp $10$ times per cluster center while conditioning on its \direction. 
We then compute the palm direction of each generated grasp and its angular error \wrt the conditioning \direction. 
A mean angular error of 4.57\textdegree
denotes accurate generation; this is also reflected in qualitative results in \supmat~(\cref{fig:supmat_cgrasp_controllability}).

\zheading{\creach}
This is conditioned on an arm \direction and 
wrist location.
We extract all arm directions and wrist locations of \ReplicaGrasp's \cite{flex} and \grab's \cite{grab} test sets, and cluster each of these 2 modalities into $200$ centers via \mbox{K-means}. 
With these, 
we obtain 40000 combinations of arm directions and wrist locations for conditioning \creach and generating 40000 reaching bodies for each (left/right) arm. 
Then, we compute over all generated bodies the mean angular error for arm \direction (as above for palm \direction for \cgrasp), and the Mean Squared Error (MSE) for wrist locations. 
The values in \cref{tab:creach_cgrasp_accuracy} denote accurate synthesis.

\subsection{Hand-Only Grasps (\cgrasp)}
\label{sec:CGraspquant}
We evaluate \cgrasp on the \grab \cite{grab} dataset 
against 
\dexgraspnet~\cite{dexgraspnet}, 
\contactgen~\cite{contactgen}, and 
\grabnet~\cite{grab}. 
For each method, we generate $200$ grasps for each of the 6 test objects, and 
compute the following metrics 
as in \cite{contactgen, hasson2019obman, saga, zhang2019, zhang2020place, jiang2021handobjectcc, turpin2022graspddc, karunratanakul2021asn}. 
We report results in \cref{tab:cgrasp_quantitative}

\qheading{Contact ratio \cite{contactgen}}
We detect the contacting \mano vertices by thresholding its distances (1mm) from the object, and compute the ratio of these over all \mano vertices. 

\qheading{Penetration percentage (\%)~\cite{flex}}
We compute the percentage of hand vertices penetrating the object via the signed distances of the two meshes 
(distance $\le-1$mm). 

\qheading{Penetration volume \cite{hasson2019obman}}
We voxelize the hand and object meshes using 
voxels of volume $v=1\text{mm}^3$, and detect intersecting voxels $N$. 
Then, the penetration volume is $N \cdot v$. 

\qheading{Penetration depth \cite{hasson2019obman}}
We compute the minimum translation along the opposite palmar \direction ($\graspDir$ in \cref{sec:CGraspControl}) necessary for resolving any hand-object penetrations. 

\qheading{Hand pose diversity \cite{flex}}
We align all hands at the same wrist location and palm orientation, and compute the mean Euclidean vertex distance  over all possible mesh pairs. 

\Cref{tab:cgrasp_quantitative} shows that \cgrasp performs on par with baselines. 
That is, \cgrasp's benefit of controllability does not harm performance.
We show qualitative results in \supmat (\cref{fig:supmat_cgrasp_qualitative_results}); 
these reflect quantitative ones. 
We also compare contact 
heatmaps in \cref{fig:contact_maps_cgrasp}. 
Baselines 
involve mainly the fingertips, while \cgrasp involves also parts of the palm.
\begin{table}[t]
\centering
\scriptsize
\vspace{-0.5 em}
\begin{minipage}{0.45\textwidth}
\centering
\begin{tabular}{l|c|c|c}
    \toprule
      & \textbf{Angle (degrees) $\downarrow$} 
      & \textbf{MSE (cm) $\downarrow$} 
      & \textbf{Inf.~time (s) $\downarrow$}\\
    \midrule
    \textbf{\creach-RA}& 7.67 & 4 & 0.46 \\
    \midrule
    \textbf{\creach-LA}& 7.23 & 3.6 & 0.46 \\
    \midrule
    \textbf{\cgrasp}& 4.57 & N/A & 0.47 \\  
    \bottomrule
\end{tabular}
\vspace{-0.5 em}
\caption{\textbf{Condition accuracy.}
    \creach and \cgrasp generate bodies and hands  
    conditioned on a (arm/hand) direction. 
    We report the angular error of the arm/palm direction,  
    the Mean Squared Error (MSE) of wrist joints, and inference time.
    For \creach we evaluate right- (RA) and left-arm (LA) reaching. 
}
\label{tab:creach_cgrasp_accuracy}
\end{minipage}
\end{table}

\begin{figure}
    \vspace{-0.5 em}
    \centering      
        \includegraphics[trim=000mm 012mm 000mm 025mm, 
        clip=true, width=0.99\columnwidth]{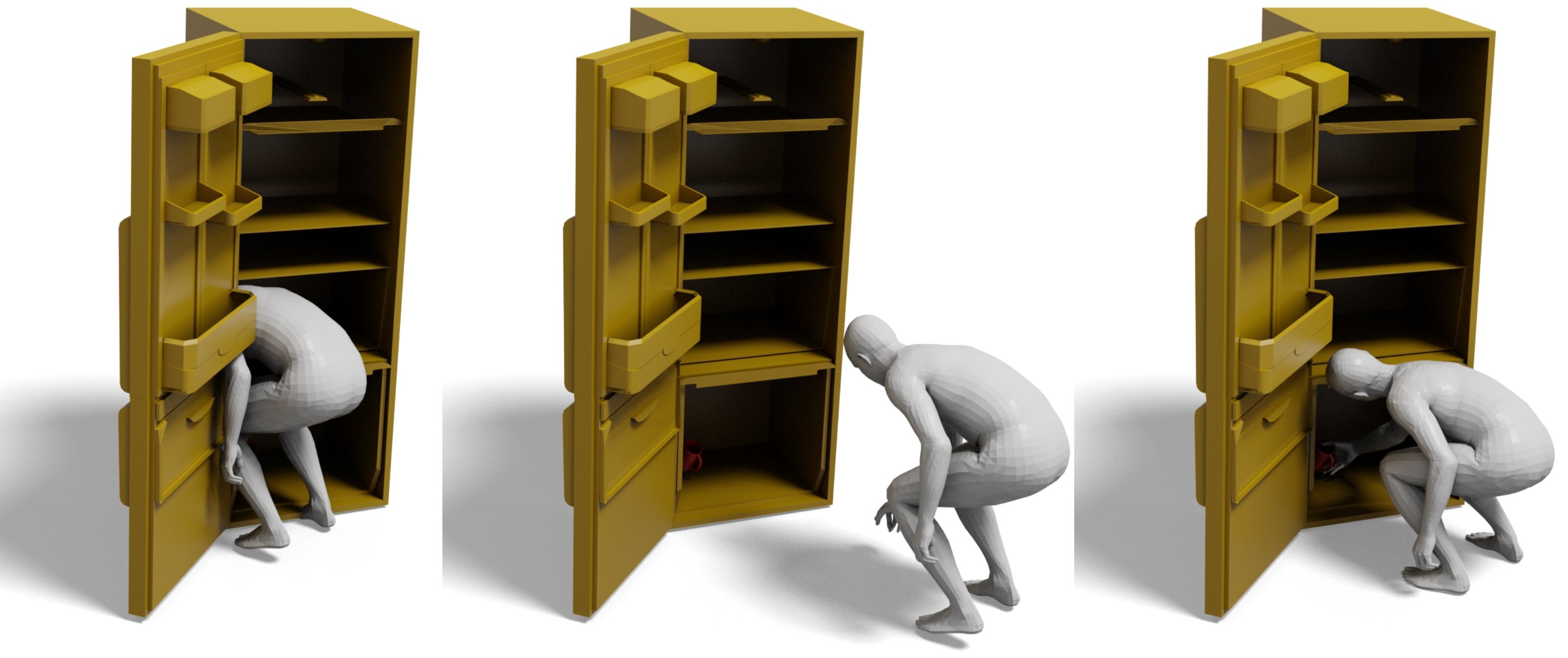}
    \vspace{-0.5 em}
    \caption{
        \textbf{\creach failure}. 
        \creach~might produce a reaching body that penetrates the receptacle 
        (left).  
        To correct for this, 
        we translate the body by $1$m 
        (middle) 
        along the opposite floor-projected 
        arm \direction. 
        Then, \cWGrasp's optimization (\cref{sec:CWGrasp}) pulls the body back to the object, while refining body and finger pose 
        (right).  
    }
    \label{fig:creach_penetrations}
\end{figure}

\subsection{Whole-Body Grasps (\cWGrasp)} 
\label{sec:CWGraspquant}
We evaluate \cWGrasp on the \ReplicaGrasp dataset~\cite{flex} and compare it against the state-of-the-art \flex~\cite{flex} method. 

\zheading{Experimental setup}
\ReplicaGrasp 
places \grab~\cite{grab} 
objects on various 
receptacles, \eg, sofas, tables. 
Each of the $50$ \grab objects appears in $192$ configurations, varying the receptacle and the object's location and orientation on it. 
For our experiments, we use the $6$ test objects and $6$ randomly-sampled training objects of \grab, and randomly select $20$ configurations per object. 
For each configuration, we generate grasping bodies with both \cWGrasp and \flex and compare the two methods. 
Note that \flex optimizes $500$ samples, and eventually keeps $10$ samples with smaller losses; we consider the ``best'' (smallest-loss) one. 
 Instead, our 
\cWGrasp uses only a single sample. 

\zheading{Quantitative evaluation}
We report the five metrics defined in \cref{sec:CGraspquant} also here in \cref{tab:cwgrasp_quantitative}, but with the following adaptations due to switching to whole-body context. 
We compute the ``penetration percentage'' 
separately for body--receptacle ($\graspbody-\receptacle$) and for \rhand--object ($\grasprhand-\object$) interaction. 
We compute the ``contact ratio'' for \rhand--object ($\grasprhand-\object$) interaction. 
We compute the ``body pose diversity'' by extending ``hand pose diversity'' to whole-body meshes. 
Last, we report the mean optimization time for each method. 
We observe that our \cWGrasp framework is highly competitive against \flex, while using 500$\times$ less samples and being one order of magnitude faster. 

\zheading{Qualitative evaluation}
We visualize several whole-body grasps produced by 
\cWGrasp and \flex in \cref{fig:cwgrasp_qualitative}. 
We observe that \cWGrasp produces more natural-looking body poses.  
For many more qualitative results, including close-up views into hands, as well as left-hand whole-body interactions, 
please see \supmat~(\cref{sec:supmat_cwgrasp_qual}).
We also compare aggregated contact heatmaps in \cref{fig:contact_maps_cwgrasp}. 
We see that \flex grasps tend to use mainly the fingertips, while \cWGrasp grasps activate also parts of the palm, so they look richer. 
 
\zheading{Perceptual Study} 
To evaluate the perceived realism of generated grasps, we conduct a perceptual study. 
To this end, we sample object-and-receptacle configurations from the \ReplicaGrasp \cite{flex} dataset, 
and for each one, we generate two whole-body grasps (referred to as ``samples'') with \cWGrasp and \flex, respectively. 
For each sample, we conduct two comparisons by rendering a whole-body view and a zoomed-in view onto the hand and object (see examples in \supmat~\cref{fig:supmat_perceptual_supp_mat_results}). We randomize the order  
that we present samples, as well as their placement. 
Each sample is shown to $35$ participants, who choose which method generates the most realistic grasp (see the protocol shown to participants in \supmat~\cref{fig:supmat_perceptual_description}). 
In total, we show $28$ samples, of which $4$ are catch trials (letting us filter out $2$ of the participants). 
Considering the full-body view, 
\cWGrasp is preferred $70.8\%$ of the times. 
Considering the zoomed-in view, 
it is preferred $71.6\%$ of the times. 
Considering both views, 
it is preferred $71.23\%$ of the times.
That is, our \cWGrasp produces grasps that are perceived as significantly more realistic than the state of the art.

\section{Conclusion}
We develop \cWGrasp, a method that generates whole-body grasps for objects  through novel \emph{controllable synthesis}. 
To this end, we first learn \reachingField, a novel model for estimating \directions a body can approach the object from. 
However, current body and grasp generators lack controllability. 
To fill this gap, we learn the novel \creach and \cgrasp models that generate a reaching body and a grasping hand  
with a
desired arm and palm \threeD~\direction, respectively.
We condition both \creach and \cgrasp on 
the same \direction sampled from \reachingField to produce a 
grasping hand 
and reaching body that are compatible with each other. 
Last, our \cWGrasp method combines these with only a small refinement, efficiently producing grasps that are perceived as significantly more realistic than the state of the art.

\newcommand{\resizeHeatmapsHO}{0.106}

\begin{figure}
    \centering
    \scriptsize
    \captionsetup[subfigure]{labelformat=empty}
    \begin{subfigure}[b]{\resizeHeatmapsHO \textwidth}
        \centering  
        \includegraphics[trim=220mm 065mm 200mm 040mm, clip=true, height=\textwidth]{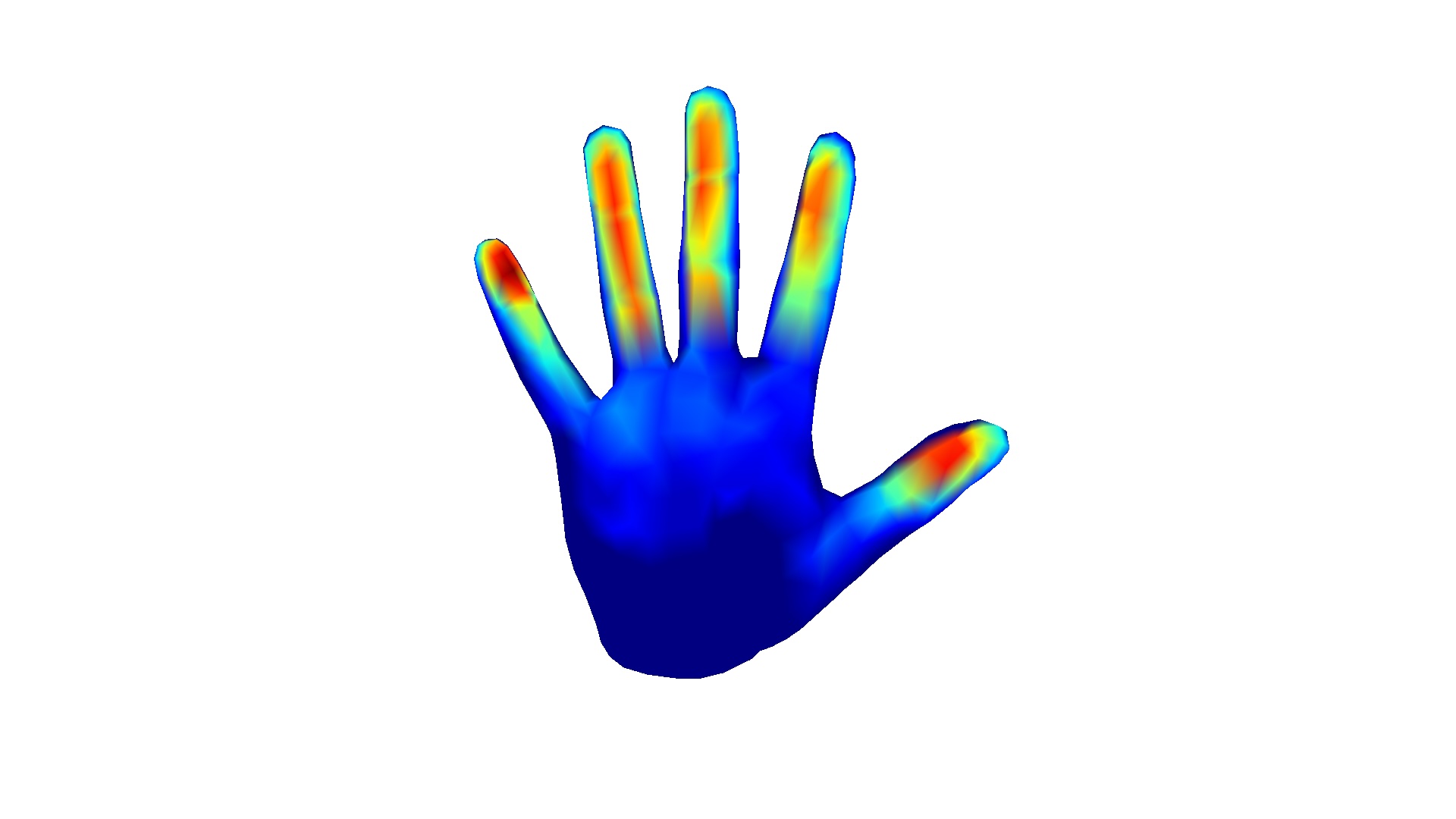}
        \caption{\dexgraspnet}
    \end{subfigure}
    \begin{subfigure}[b]{\resizeHeatmapsHO \textwidth}
        \centering
        \includegraphics[trim=210mm 065mm 200mm 040mm, clip=true, height=\textwidth]{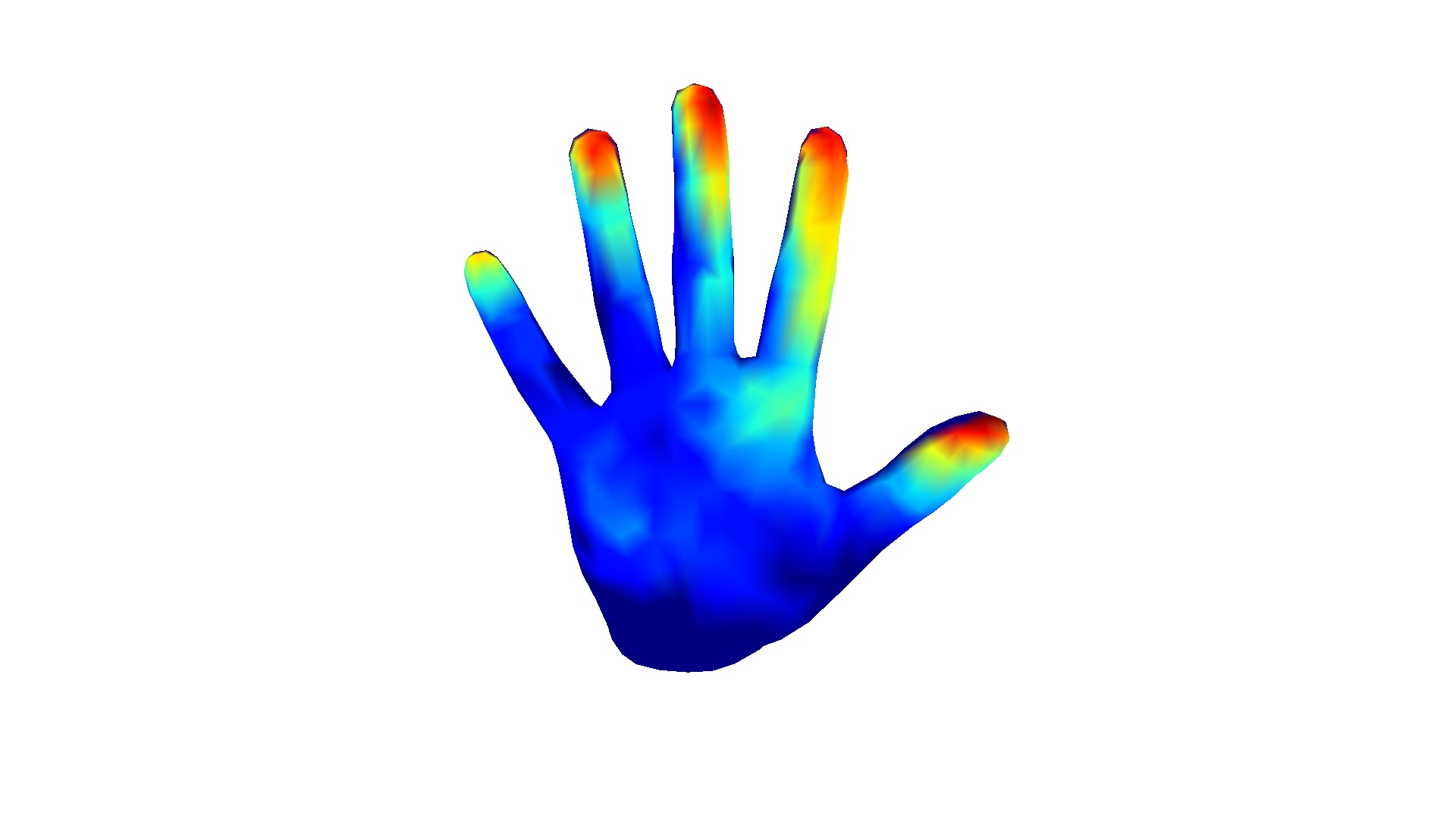}
        \caption{\contactgen}
    \end{subfigure}
    \begin{subfigure}[b]{\resizeHeatmapsHO \textwidth}
        \centering
        \includegraphics[trim=210mm 065mm 200mm 040mm, clip=true, height=\textwidth]{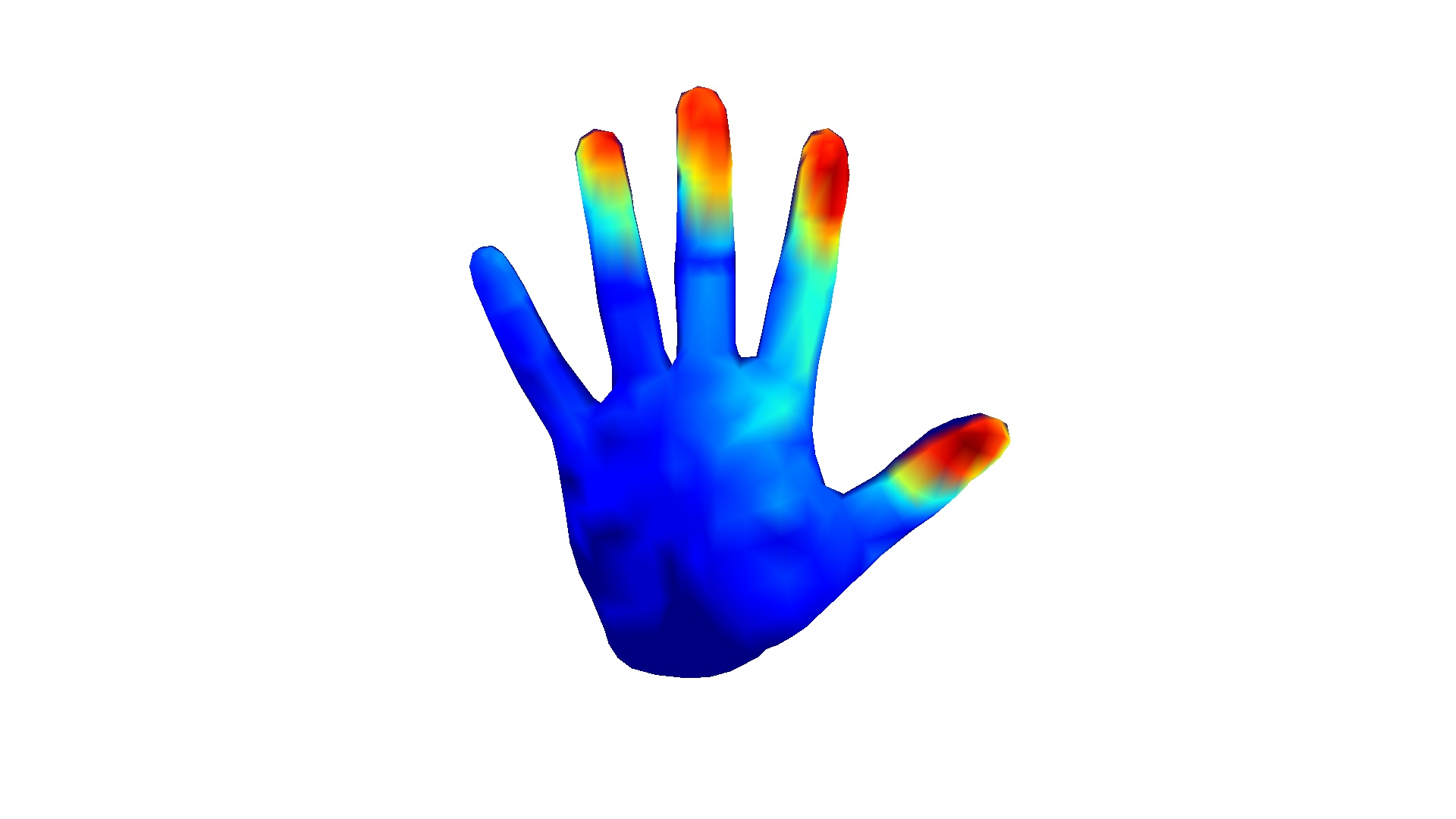}
        \caption{\grabnet}
    \end{subfigure}
    \begin{subfigure}[b]{\resizeHeatmapsHO \textwidth}
        \centering
        \includegraphics[trim=210mm 060mm 200mm 040mm, clip=true, height=\textwidth]{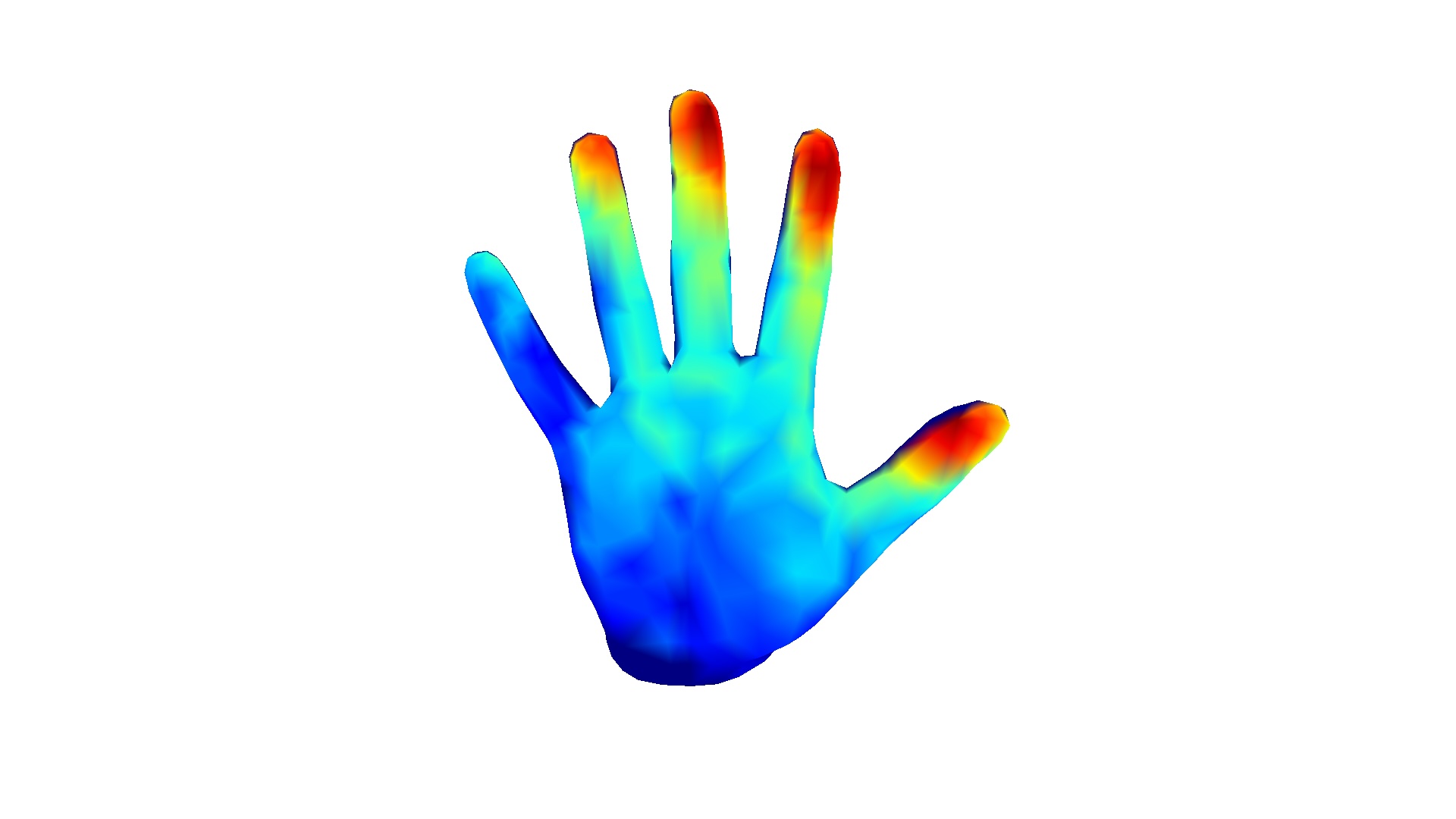}
        \caption{\cgrasp (ours)} 
    \end{subfigure}
    \vspace{-0.5 em}
    \caption{
        \textbf{Contact maps: \cgrasp \& \mbox{SotA} (\cref{sec:CGraspquant})}.  
        Contact likelihood is color-coded via heatmaps; red denotes a high likelihood and blue a low one. 
        We compare against 
        \dexgraspnet~\cite{dexgraspnet}, 
        \contactgen~\cite{contactgen}, and 
        \grabnet~\cite{grab}.
        Existing 
        methods involve mostly finger tips, while \cgrasp also involves parts of the palm. 
    }
    \label{fig:contact_maps_cgrasp}
    \vspace{-0.5 em}
\end{figure}

\begin{table}
\resizebox{0.96\columnwidth}{!}{
\begin{tabular}{l|c|c|c|c|c|c|c}
    \toprule
    & \rotatebox{90}{\textbf{\shortstack{Type}}}
    & \rotatebox{90}{\textbf{\shortstack{Control}}}
    & \textbf{\shortstack{Cont.   \\ ratio              \\$\uparrow$} } 
    & \textbf{\shortstack{Penetr. \\ perc.              \\$\%$~$\downarrow$}}
    & \textbf{\shortstack{Penetr. \\ vol.~$\downarrow$  \\$mm^{3}$}}
    & \textbf{\shortstack{Penetr. \\ depth              \\$mm\downarrow$}}
    & \textbf{\shortstack{Hand    \\ pose~div.          \\ $cm\uparrow$}}\\
    \midrule
    \textbf{\dexgraspnet}
    &O& \xmark &$0.11$ &$\textbf{0.13}$ & $1.25$ &$\textbf{1.2}$ &$\textbf{7.08}$ \\
    \midrule
    \textbf{\contactgen}
    &R& \xmark &$0.09$ &$1.15$ & $\textbf{1.04}$ & $2$ &$6.75$ \\
    \midrule
    \textbf{\grabnet}
    &R& \xmark & $\textbf{0.13}$& $2.4$& $1.27$ &$2.6$ &$6.72$ \\
    \midrule
    \textbf{\cgrasp}~(ours)
    &R& \cmark &$0.12$ & $2.9$& $1.16$ & $2.8$&$6.72$  \\
    \bottomrule
\end{tabular}
}
\vspace{-0.5 em}
\caption{
            \textbf{Evaluation: 
            \cgrasp \& \mbox{SotA} (\cref{sec:CGraspquant})}. 
            The ``type'' column denotes regression (R) or optimization (O) methods. 
            The ``control'' column indicates whether a method is controllable via directional conditioning. 
            Our \cgrasp performs \emph{on par} with existing methods, while 
            being controllable via a direction condition. That is, the benefit of \emph{controllability does not harm} performance.
}
\label{tab:cgrasp_quantitative}
\end{table}
\newcommand{\resizeHeatmapsWB}{0.84}

\begin{figure}
    \vspace{-1.0 em}
    \centering
    \captionsetup[subfigure]{labelformat=empty}
    \begin{subfigure}{0.24 \columnwidth}
        \centering 
        \includegraphics[width=\resizeHeatmapsWB \columnwidth]{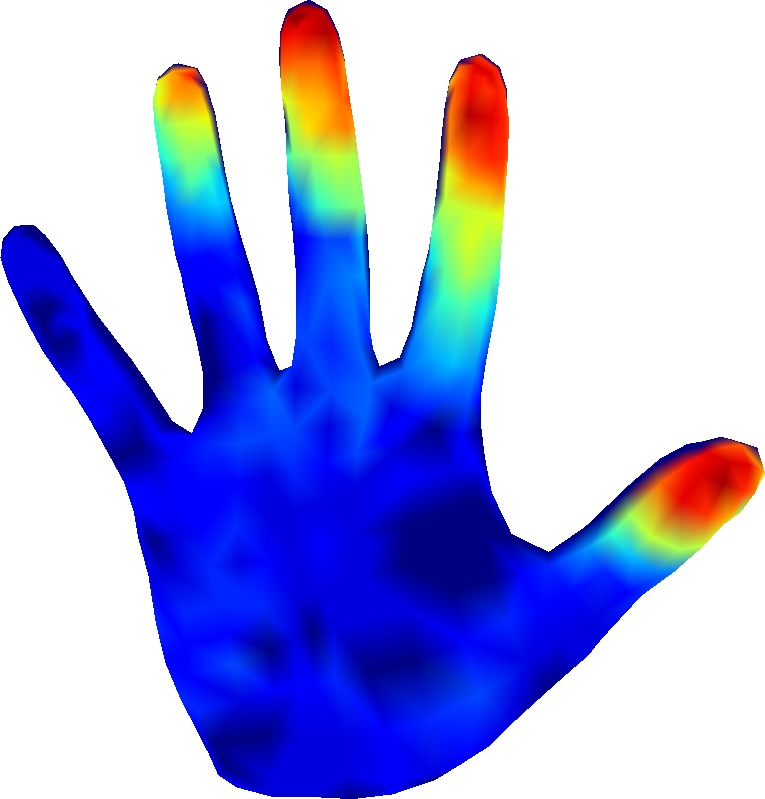}
        \caption{\flex \cite{flex}}
    \end{subfigure}
    \hspace{4.0 em}
    \begin{subfigure}{0.24 \columnwidth}
        \centering
        \includegraphics[width=\resizeHeatmapsWB \columnwidth]{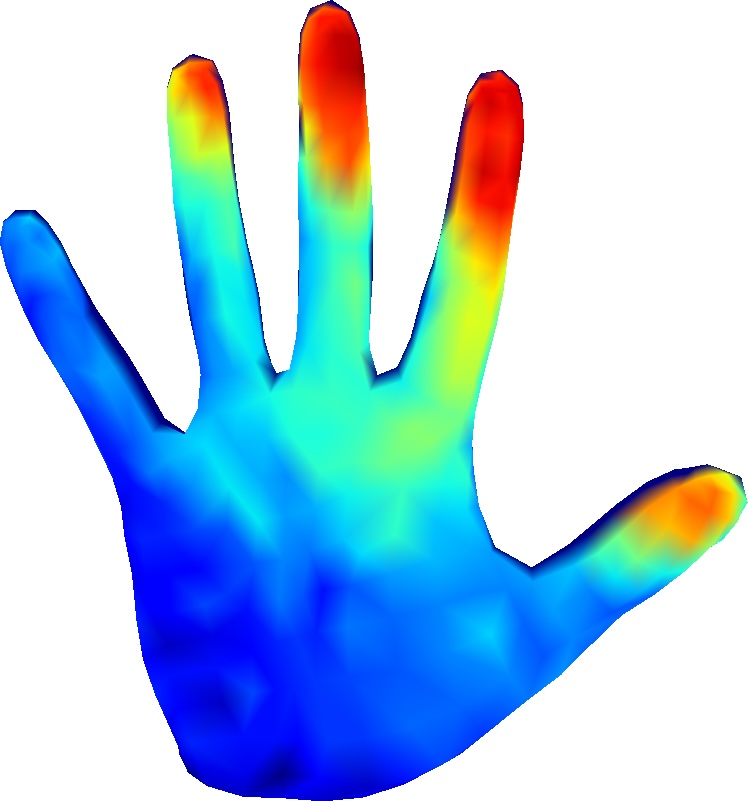}
        \caption{\cWGrasp (ours)}
    \end{subfigure}
    \vspace{-0.5 em}
    \caption{
        \textbf{Contact maps: \cWGrasp \& \flex~\cite{flex} (\cref{sec:CWGraspquant})}. 
        Contact likelihood is color-coded via heatmaps; red denotes a high likelihood and blue a low one.  
        \flex involves mainly the fingertips, while our \cWGrasp also involves 
        parts of the pam. 
        }
    \label{fig:contact_maps_cwgrasp}
    \vspace{-0.5 em}
\end{figure}

\begin{table}
\centering
\resizebox{0.94 \columnwidth}{!}{
\begin{tabular}{l|c|c|c|c|c|c}
    \toprule
    {} & 
    \rotatebox{0}{Samples} & 
    \rotatebox{0}{$\underset{\mathcal{B} -\receptacle}{Pen.~\%}$} & 
    \rotatebox{0}{$\underset{\mathcal{RH} -\object}{Pen.~\%}$} & 
    \rotatebox{0}{$\underset{\mathcal{RH} -\object}{Contact}$} & 
    \shortstack{Body div.} & 
    \rotatebox{0}{Time} 
    \\          
    {} & 
    \# & 

    $\downarrow$ & 
    $\downarrow$ & 
    $\uparrow$   & 
    $cm\uparrow$   & 
    (s)~$\downarrow$ 
    \\       
    \midrule
    \flex~\cite{flex} & 500 & \textbf{0.3} & 1.16 & 0.15 & \textbf{63.86} & 357 \\
    \midrule
    \cWGrasp & 1 & 0.7 & \textbf{0.7} & \textbf{0.3} & 61.77 & \textbf{23} \\
    \bottomrule
\end{tabular}
}
\vspace{-0.5 em}
\caption{
    \qheading{Evaluation: \cWGrasp \& \flex (\cref{sec:CWGraspquant})} 
    We report
    the number of body samples, 
    the number of optimization iterations, 
    the penetration percentage for 
        the whole body ($\graspbody$) and receptacle ($\receptacle$),
    and for the 
        \rhand ($\mathcal{RH}$) and object ($\object$), 
    the contact ratio,  
    body pose diversity, and 
    average runtime.
}
\vspace{-1.0 em}
\label{tab:cwgrasp_quantitative}
\end{table}

\pagebreak

\noindent\rule{\columnwidth}{1pt}

\qheading{Future Work}
We tackle right- and left-hand grasps; future work will look into bi-manual grasping \cite{manipnet, arctic, grab, imos}. 
Sometimes bodies look ``unstable'' when kneeling down or stretching up; 
intuitive-physics  reasoning \cite{tripathi2023ipman} might help.
Last, we will use generated grasps as targets for motion synthesis \cite{goal, saga, wang2021synthesizing, samp} to navigate scenes and grasp objects. 

{\small
\qheading{Acknowledgments} 
This work is partially 
supported by the ERC Starting Grant (project \mbox{STRIPES}, \mbox{101165317}, PI: D. Tzionas).

\qheading{Disclosure}  D. Tzionas has received a research gift from Google.}

\vfill

    \pagebreak
{
    \small
    \bibliographystyle{config/ieeenat_fullname}
    \bibliography{config/BIB}
}

\clearpage
\maketitlesupplementary

\renewcommand{\thesection}{S.\arabic{section}}
\renewcommand{\thefigure}{S.\arabic{figure}}
\renewcommand{\thetable}{S.\arabic{table}}
\renewcommand{\theequation}{S.\arabic{equation}}
\setcounter{section}{0}
\setcounter{figure}{0}
\setcounter{table}{0}
\setcounter{equation}{0}

\section{Implementation Details}
Here we describe details of our methodology. 
\Cref{sec:supmat_reaching_field} discusses details on the steps for building the \reachingField 
probabilistic model. 
\Cref{sec:supmat_creach} discusses details for the \creach model. 
\Cref{sec:supmat_cgrasp} discusses details for the \cgrasp model. 
\Cref{sec:supmat_optimization} discusses details of our \cWGrasp framework that employs all above components.

\subsection{\reachingField}
\label{sec:supmat_reaching_field}

We discuss details 
for building a \reachingField, and 
visualize steps 
in \cref{fig:supmat_ray_casting_examples}, 
where we show  
examples for 
varying 
object ``heights,'' namely distances from the ground. 

Given an object (see \mbox{\cref{fig:supmat_ray_casting_examples} \textcolor{red}{A}}), we first define a spherical grid around the object (see \mbox{\cref{fig:supmat_ray_casting_examples}} \textcolor{red}{B}), and cast rays towards all directions formed between the 
object centroid 
and the spherical-grid points (see \mbox{\cref{fig:supmat_ray_casting_examples} \textcolor{red}{C}}). 
We then follow a filtering process that includes the following steps:

\qheading{{Filter \#1. Arm/hand direction}} 
We only keep the casted rays that do not intersect with the receptacle; see \mbox{\cref{fig:supmat_ray_casting_examples}~\textcolor{red}{D}}.

\qheading{{Filter \#2 - Body orientation}} 
We project the remaining rays to the horizontal plane and check again for intersections with the receptacle; see \cref{fig:supmat_ray_casting_examples} \textcolor{red}{E}.

\qheading{{Filter \#3 - Standing places}}
We sample points across the remaining rays and cast vertical rays towards the ground (see \cref{fig:supmat_ray_casting_examples} \textcolor{red}{F}). 
Then, we detect potential ``occluders'' that lie beneath a ray and prevent a human from standing there and approaching the object along that ray direction. 

\qheading{{Filter \#4 - Wiggle room for arm volume}} 
Some remaining rays 
might still result 
in body-receptacle penetrations, 
as ray casting does not 
consider the  
volume of body limbs. 
To account for this we ``perturb'' ray directions while checking for intersections, to allow for a ``wiggle room'' that can be occupied by a body's arm.
To speed up this process, we empirically observe that 
we tend to approach an object from the right to interact with the right hand, and from the left to interact with the left hand. 
Thus, we rotate rays clockwise around the vertical axis (rotating arbitrarily could generalize better but would be slower) within a range of $[0^{o},30^{o}]$ for \rhand interactions, and counterclockwise for \lhand ones, 
while pruning intersecting rays. 
For a visualization see the \video on our website.
The above is a simple speed-up heuristic that empirically works well for our scenarios. 

\newcommand{\resizeSTEPS}{0.99}

\begin{figure}[htbp]
    \centering
    \begin{adjustbox}{max width=\resizeSTEPS\textwidth}
        \raisebox{3em}{\makebox[0.3\linewidth][l]{\textbf{A.}}} 
        \hspace{-5em} 
        \includegraphics[width=0.3\linewidth]{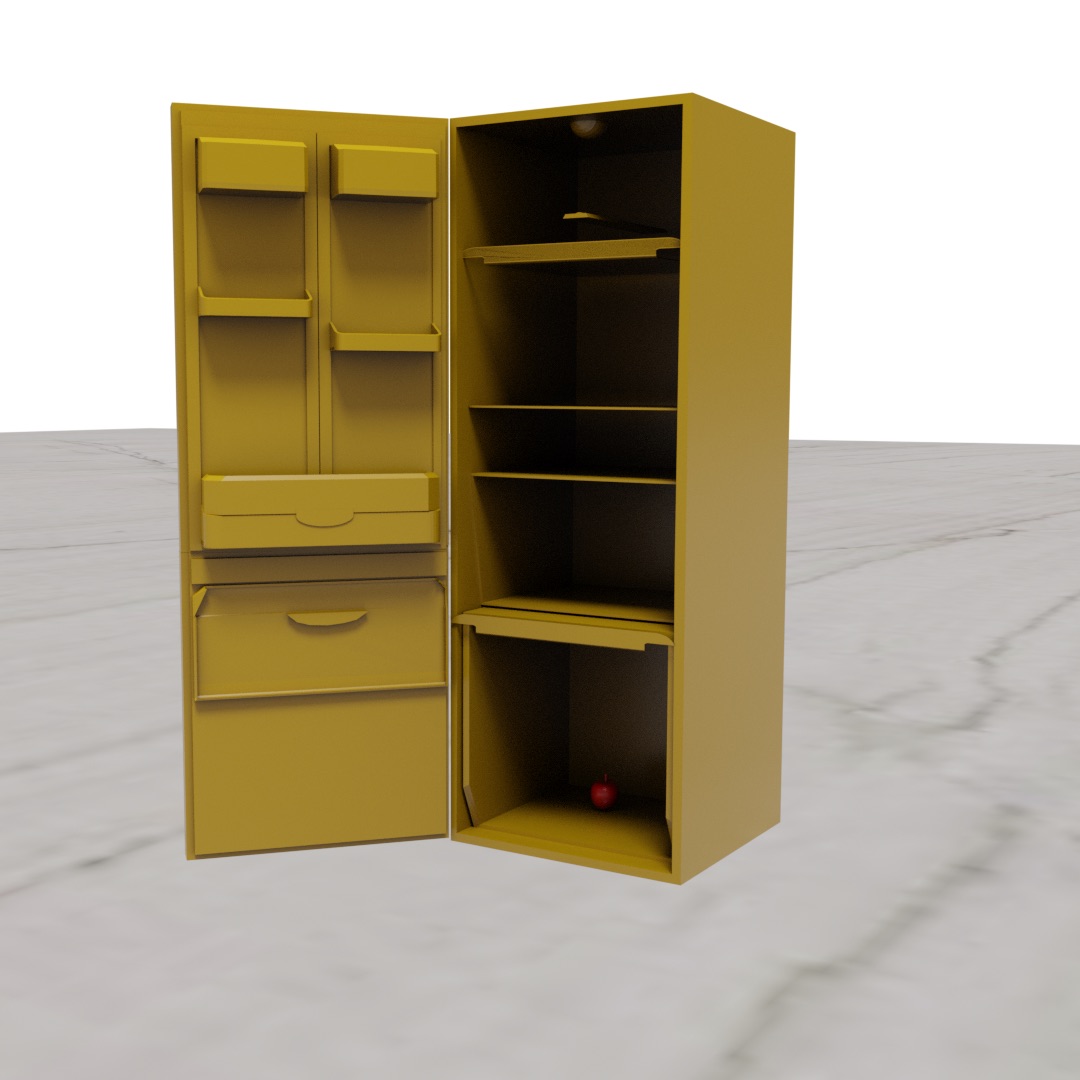}
        \includegraphics[width=0.3\linewidth]{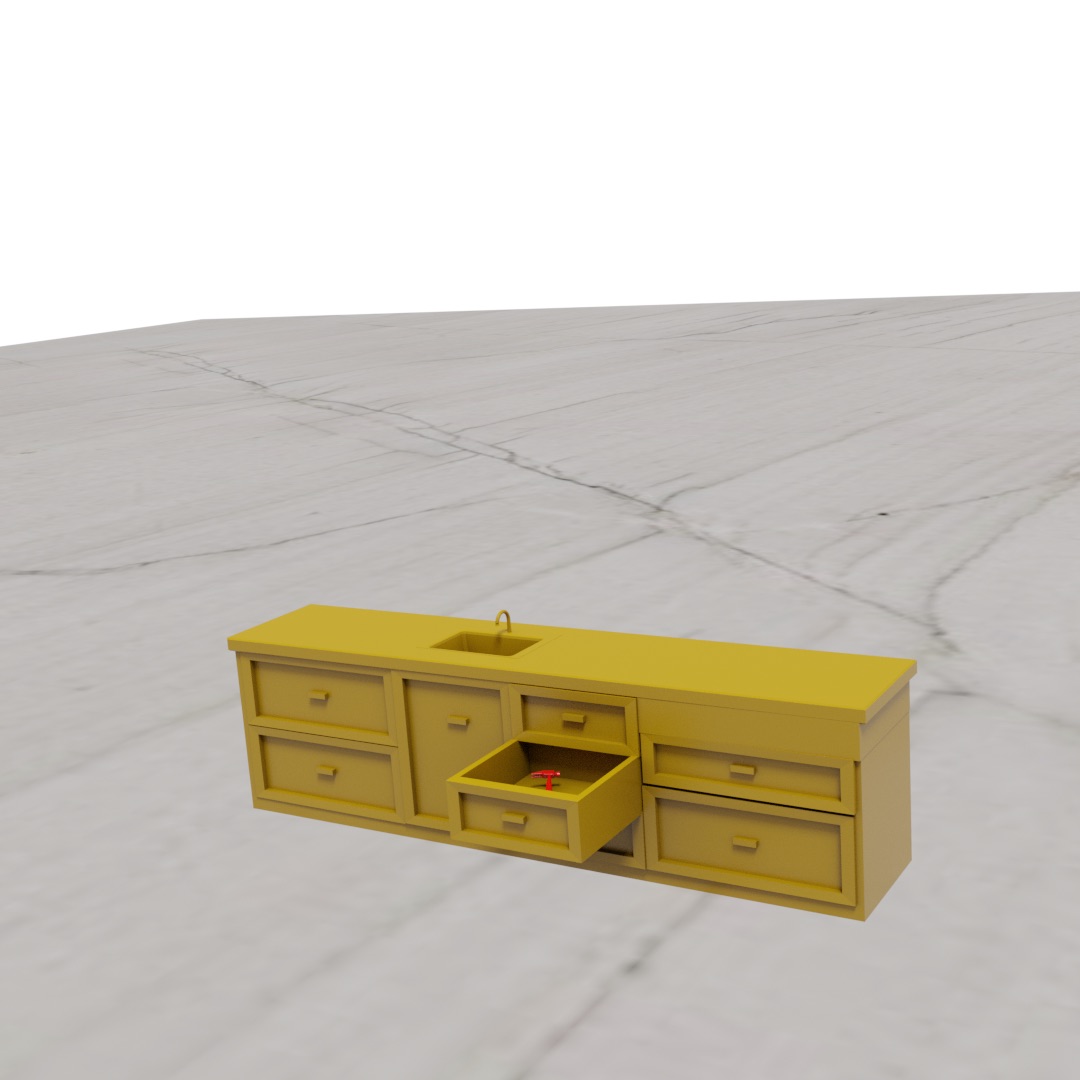}
        \includegraphics[width=0.3\linewidth]{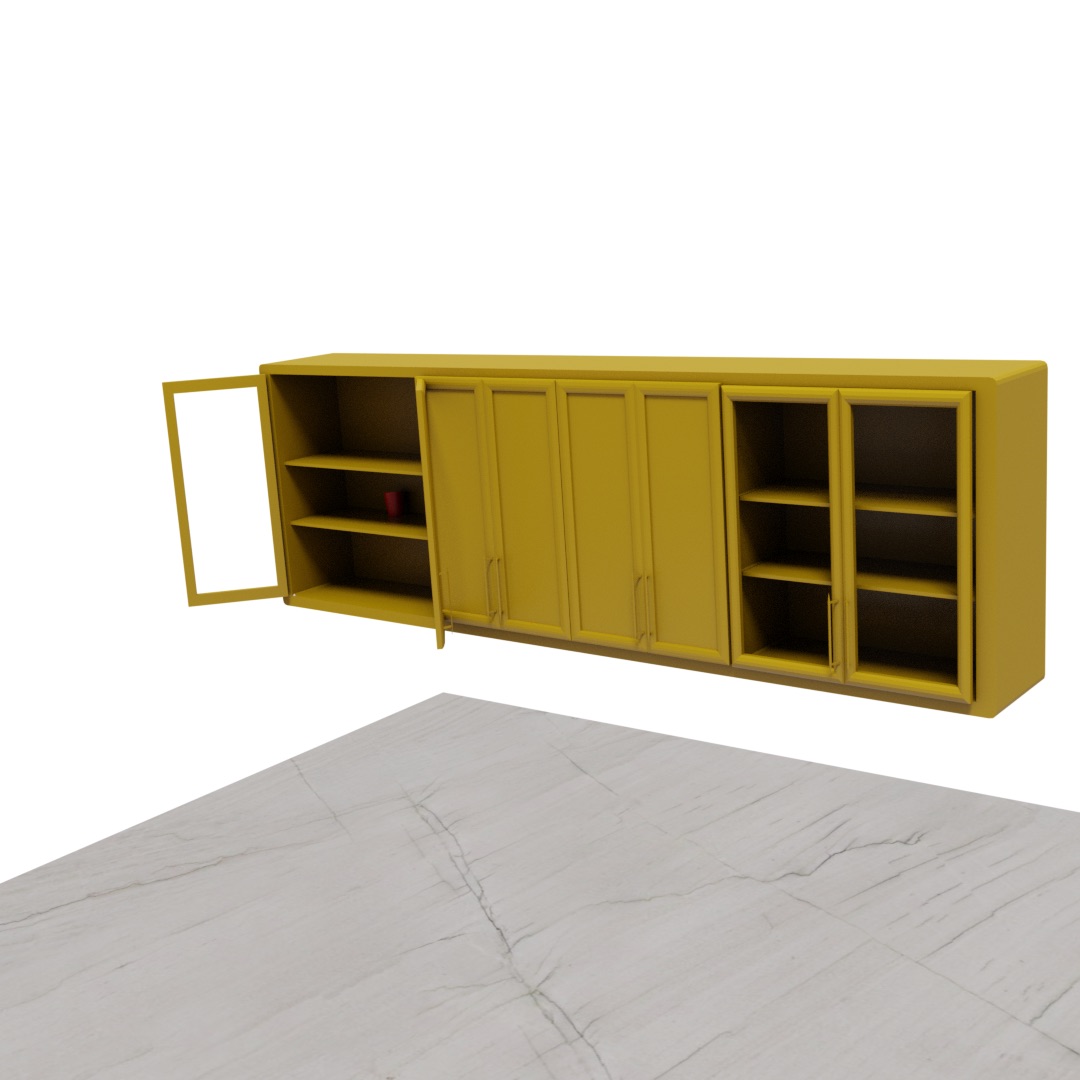}
    \end{adjustbox}

    \begin{adjustbox}{max width=\resizeSTEPS\textwidth}
        \raisebox{3em}{\makebox[0.3\linewidth][l]{\textbf{B.}}} 
        \hspace{-5em} 
        \includegraphics[width=0.3\linewidth]{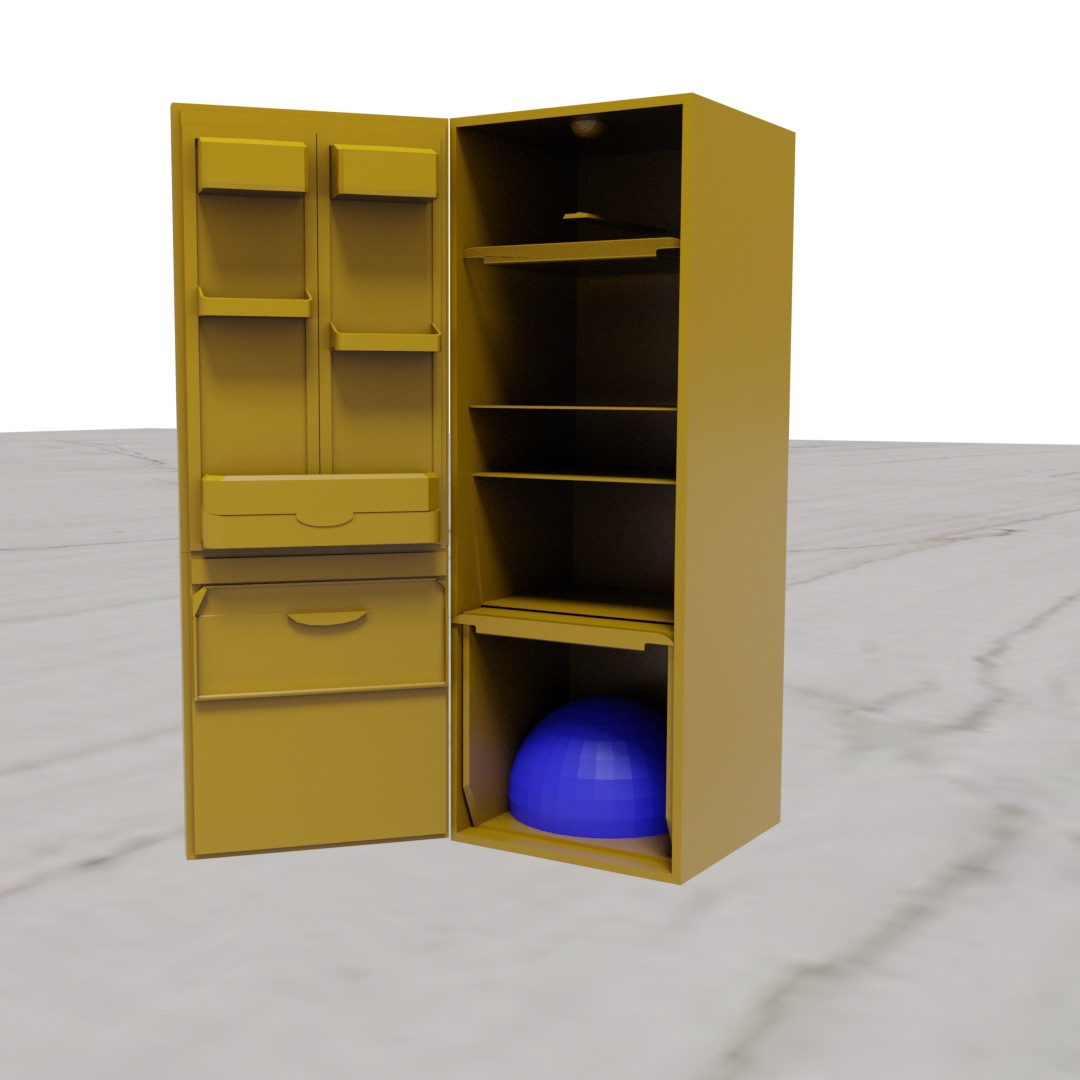}
        \includegraphics[width=0.3\linewidth]{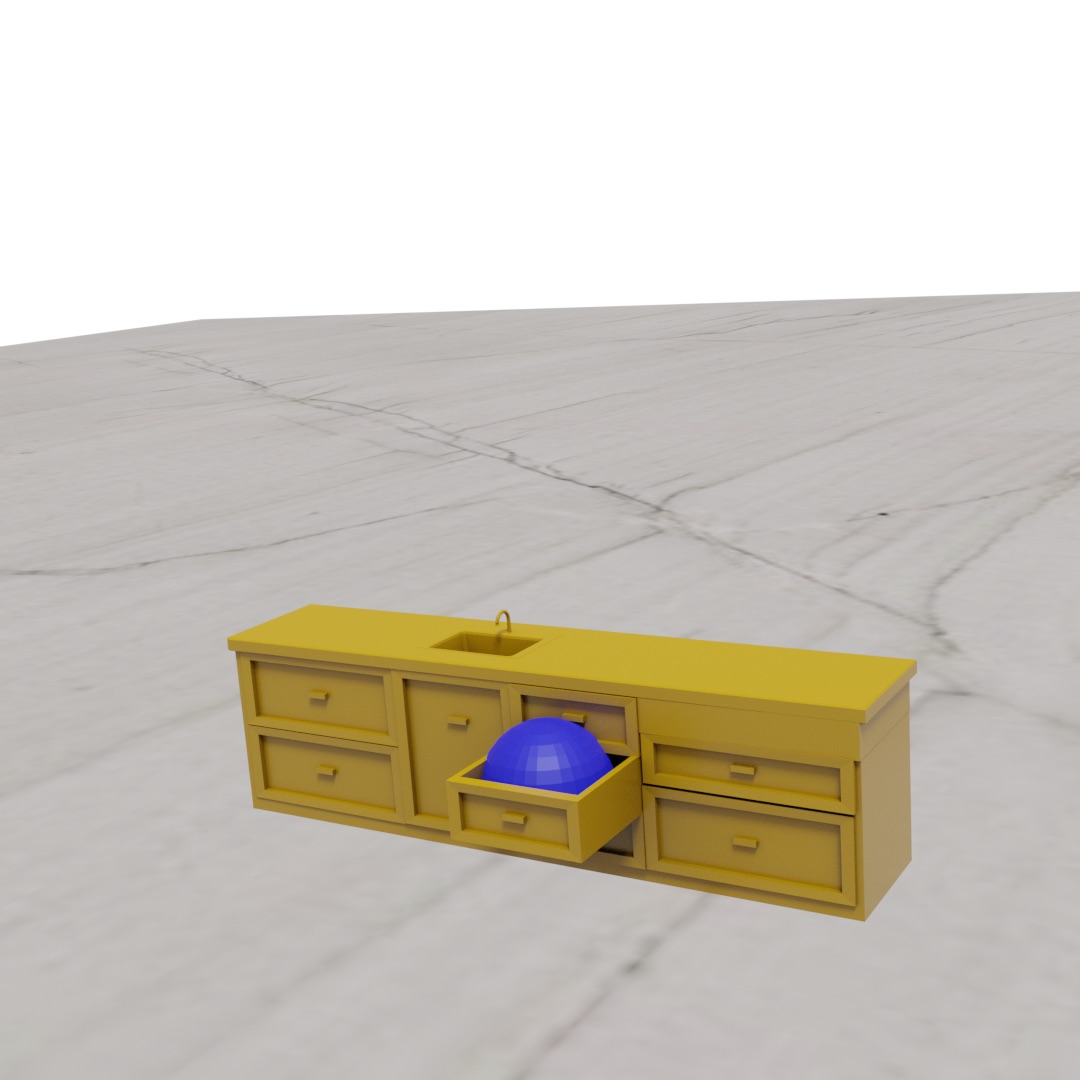}
        \includegraphics[width=0.3\linewidth]{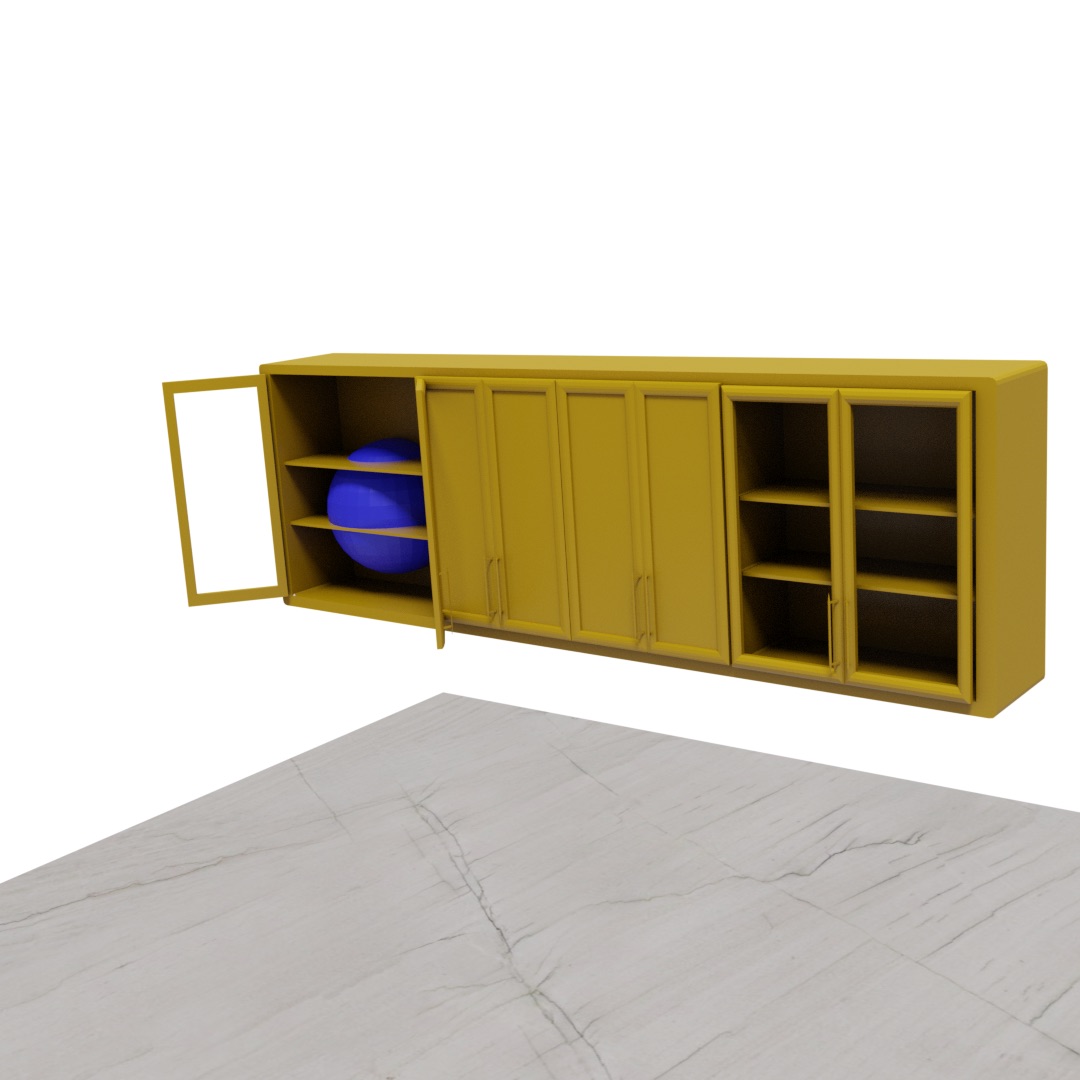}
    \end{adjustbox}

    \begin{adjustbox}{max width=\resizeSTEPS\textwidth}
        \raisebox{3em}{\makebox[0.3\linewidth][l]{\textbf{C.}}} 
        \hspace{-5em} 
        \includegraphics[width=0.3\linewidth]{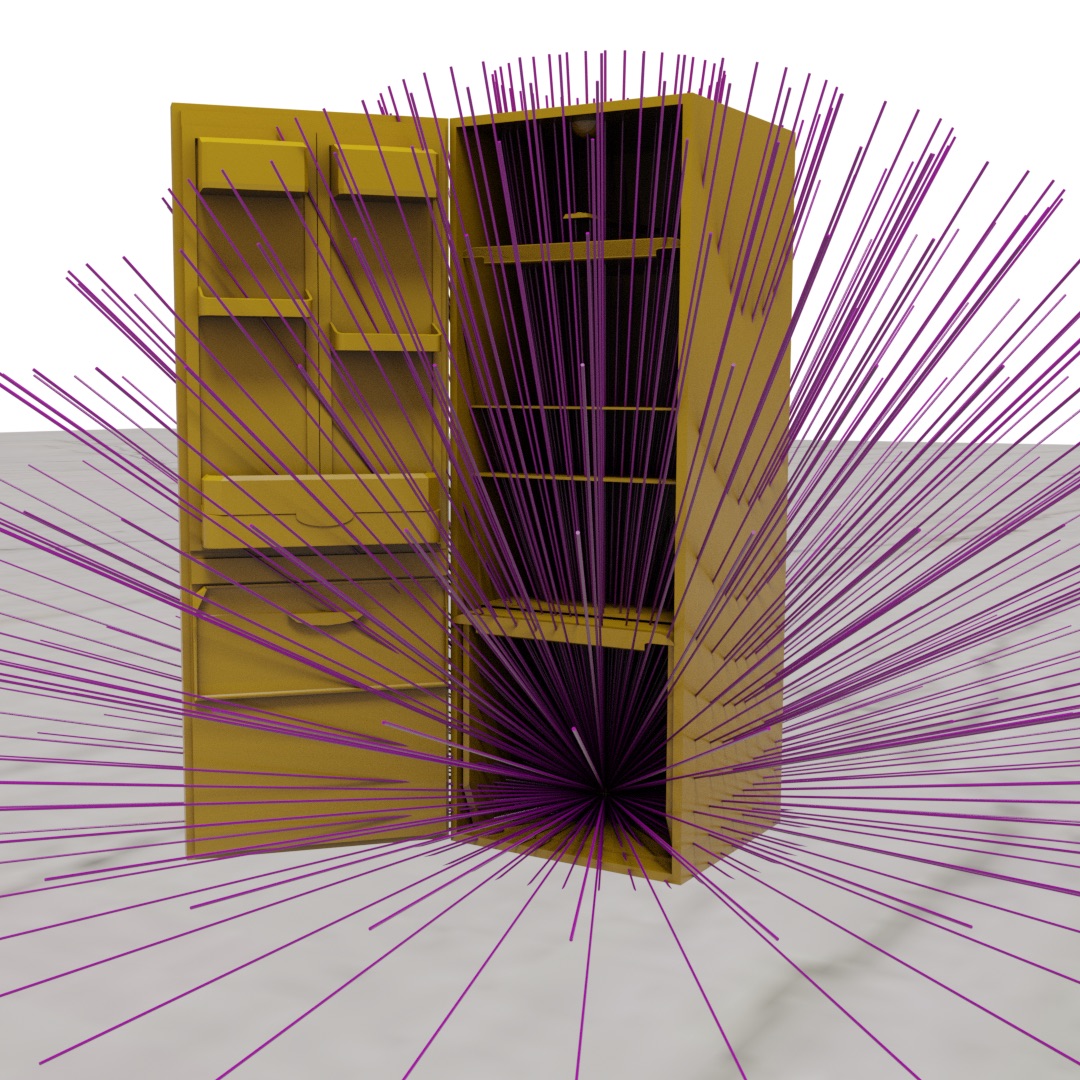}
        \includegraphics[width=0.3\linewidth]{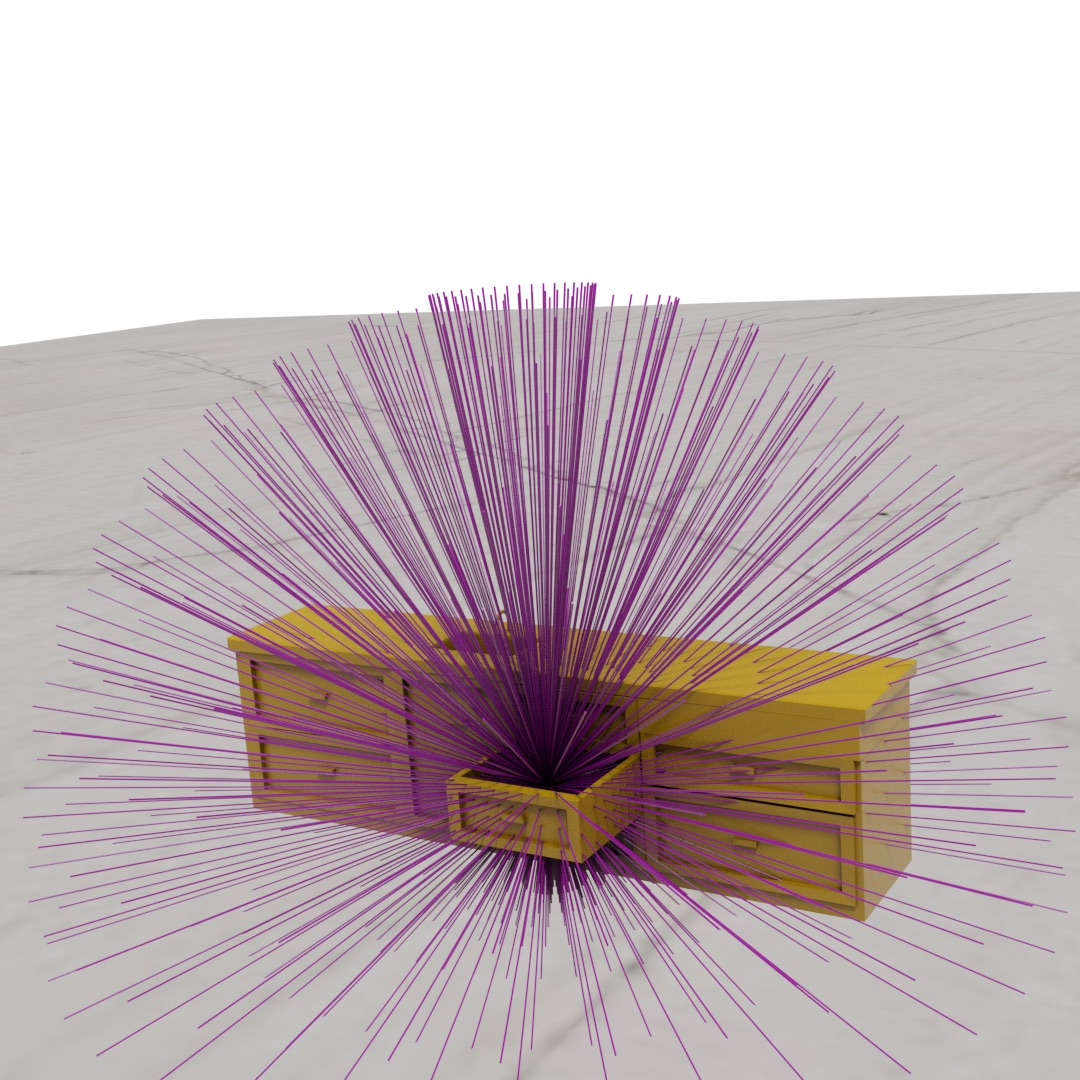}
        \includegraphics[width=0.3\linewidth]{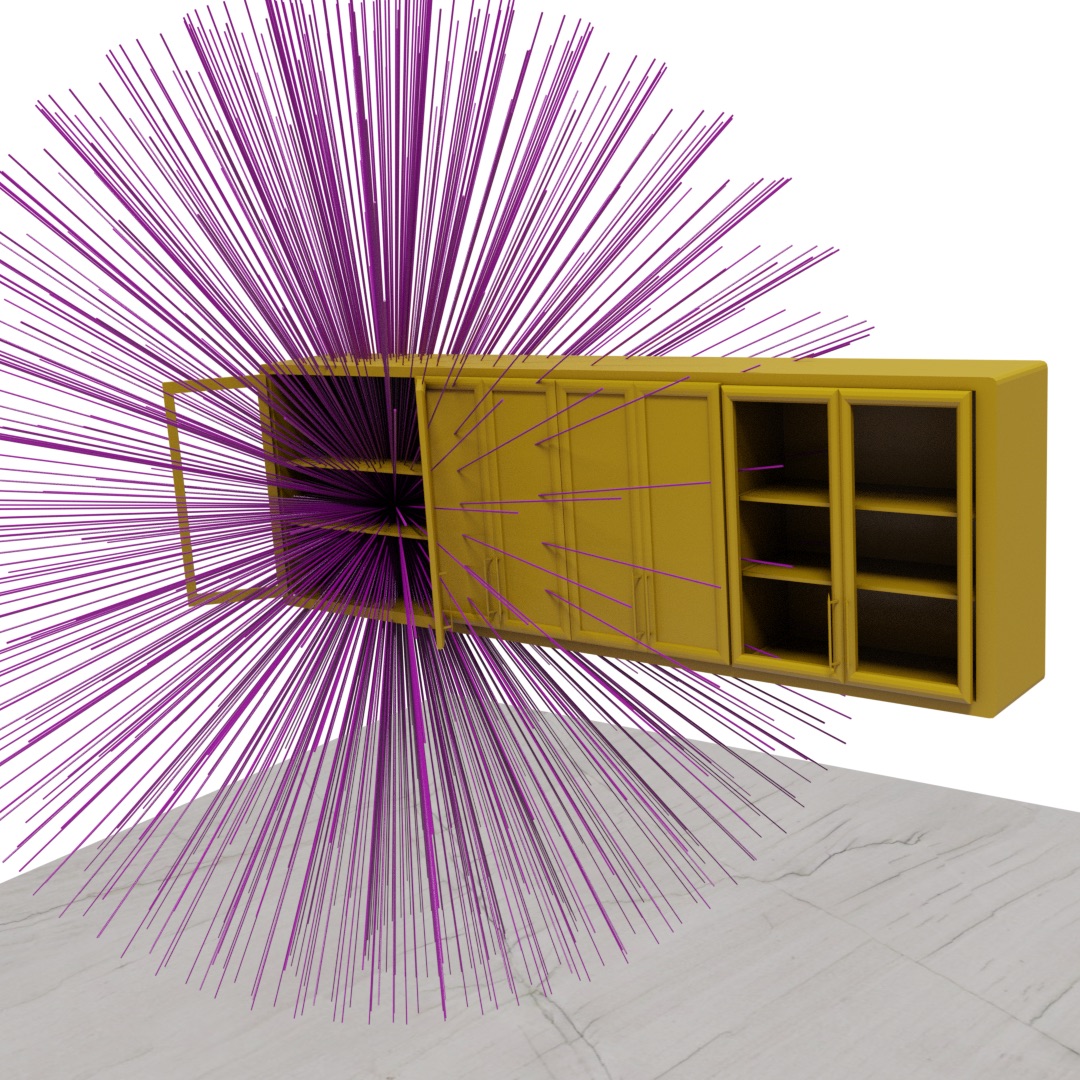}
    \end{adjustbox}

    \begin{adjustbox}{max width=\resizeSTEPS\textwidth}
        \raisebox{3em}{\makebox[0.3\linewidth][l]{\textbf{D.}}} 
        \hspace{-5em} 
        \includegraphics[width=0.3\linewidth]{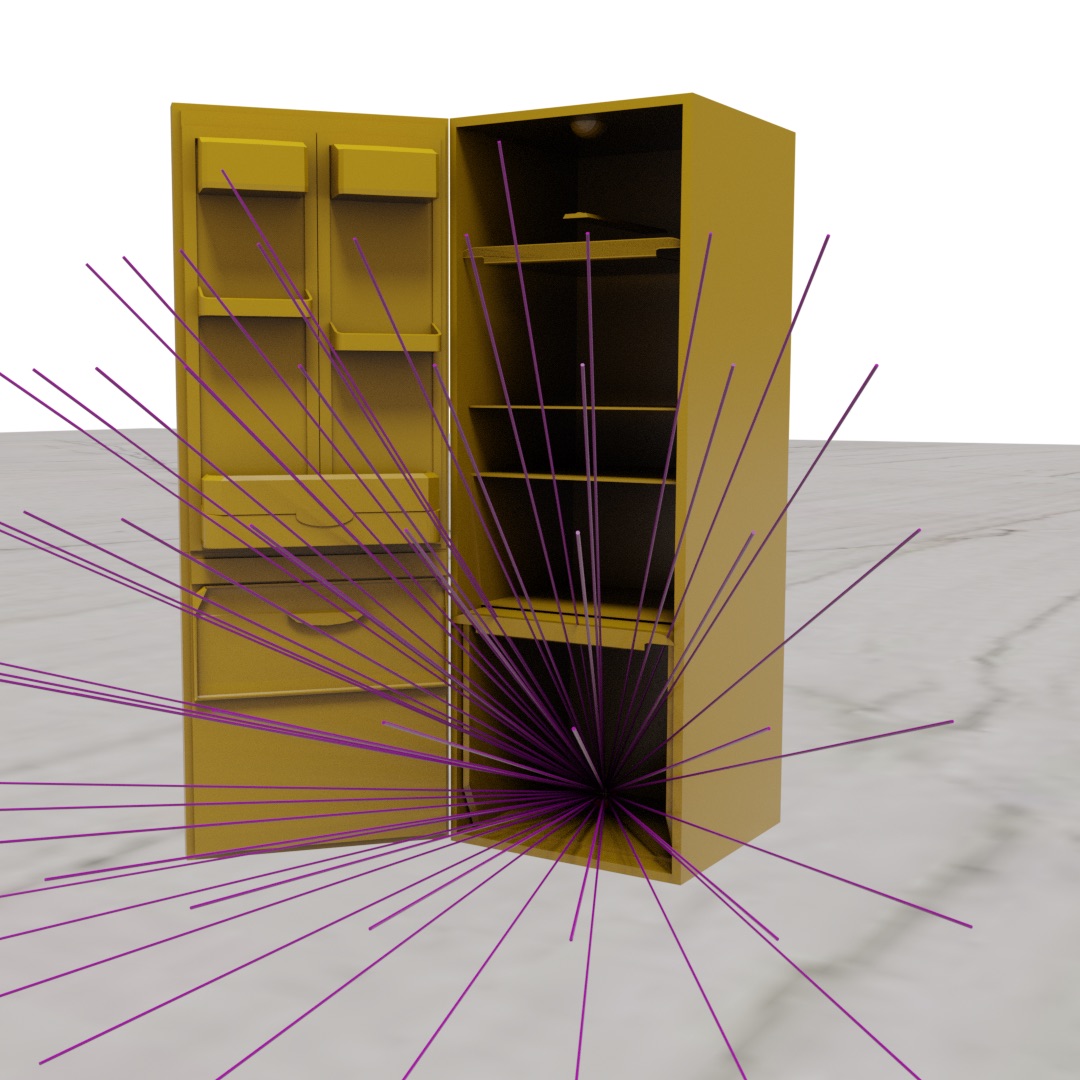}
        \includegraphics[width=0.3\linewidth]{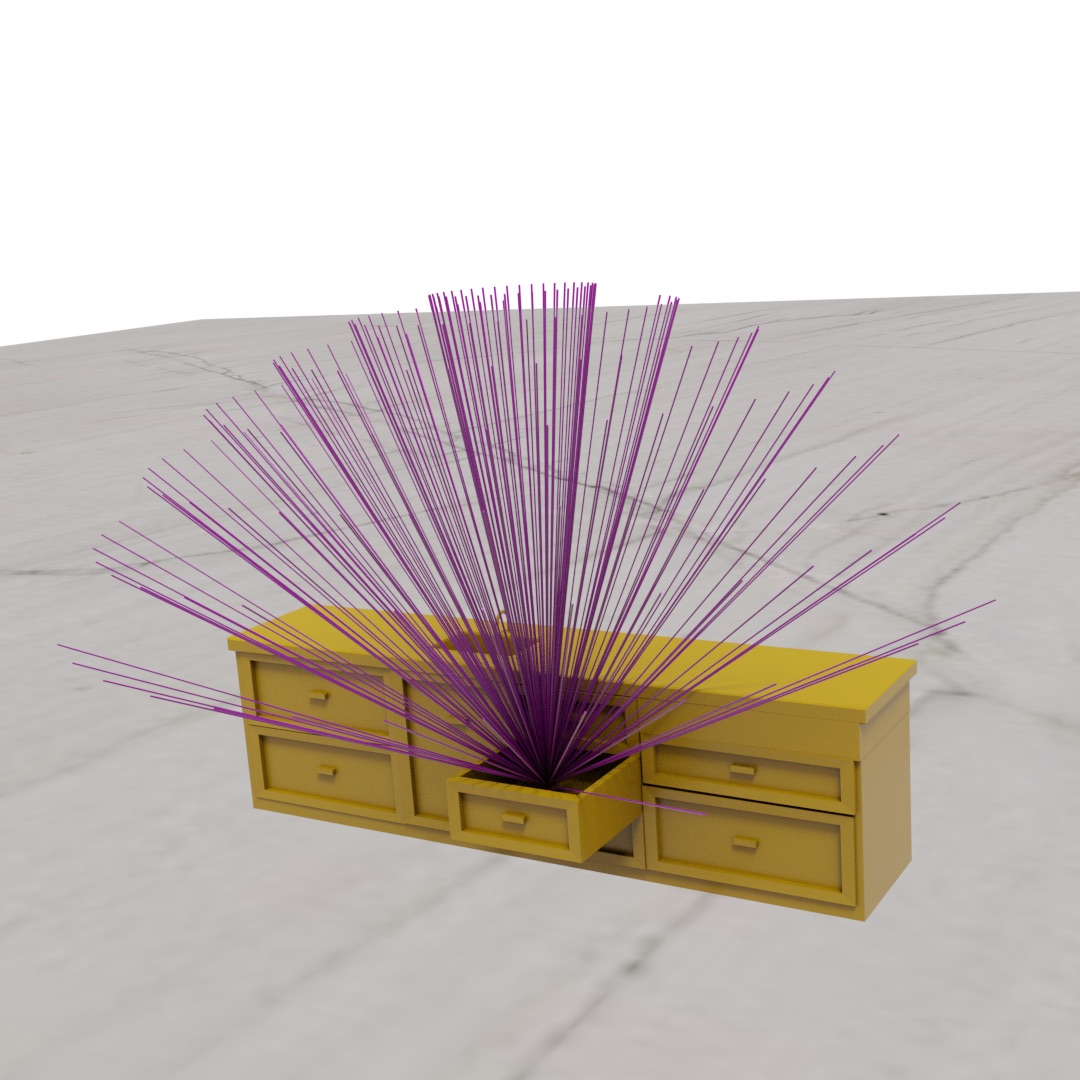}
        \includegraphics[width=0.3\linewidth]{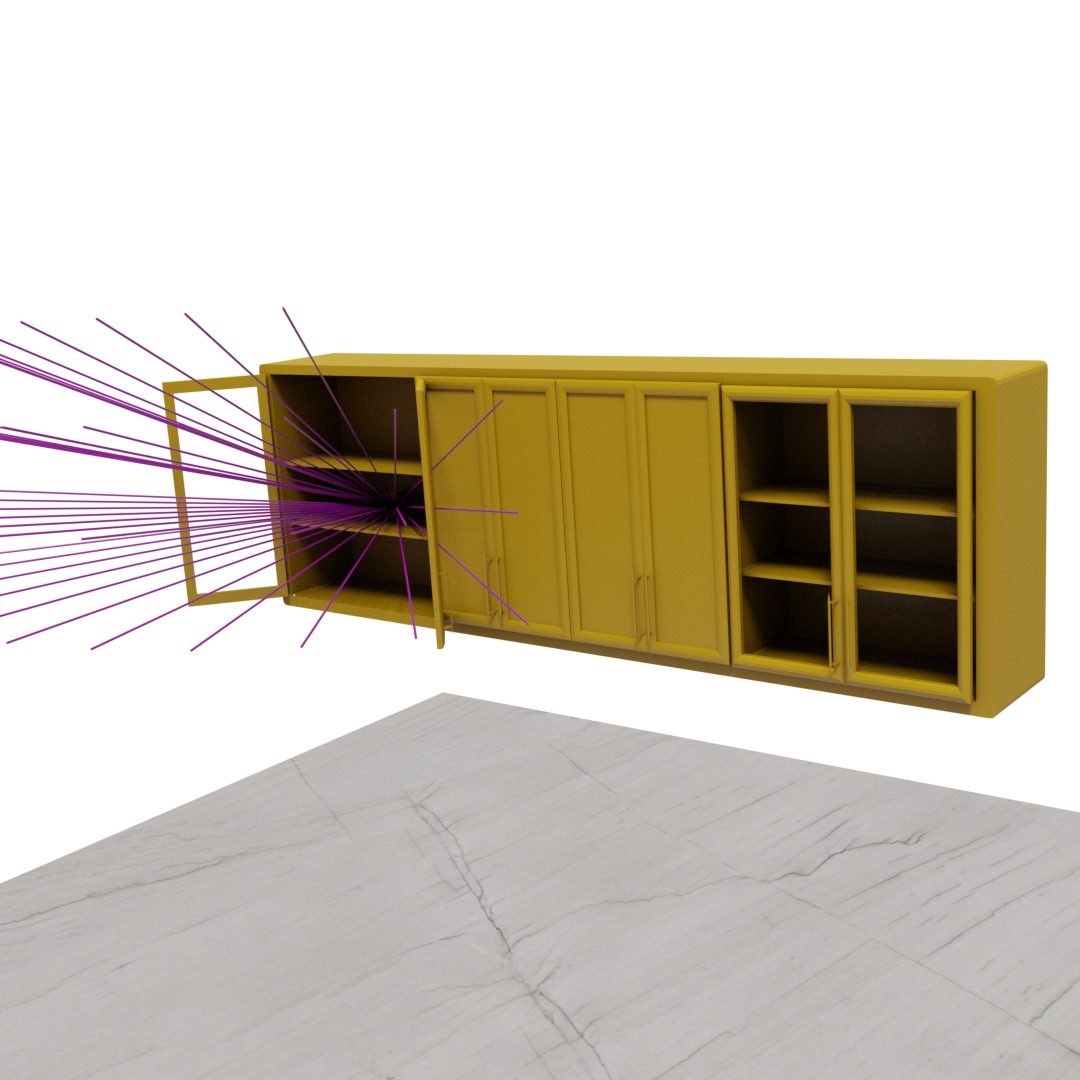}
    \end{adjustbox}

    \begin{adjustbox}{max width=\resizeSTEPS\textwidth}
        \raisebox{3em}{\makebox[0.3\linewidth][l]{\textbf{E.}}} 
        \hspace{-5em} 
        \includegraphics[width=0.3\linewidth]{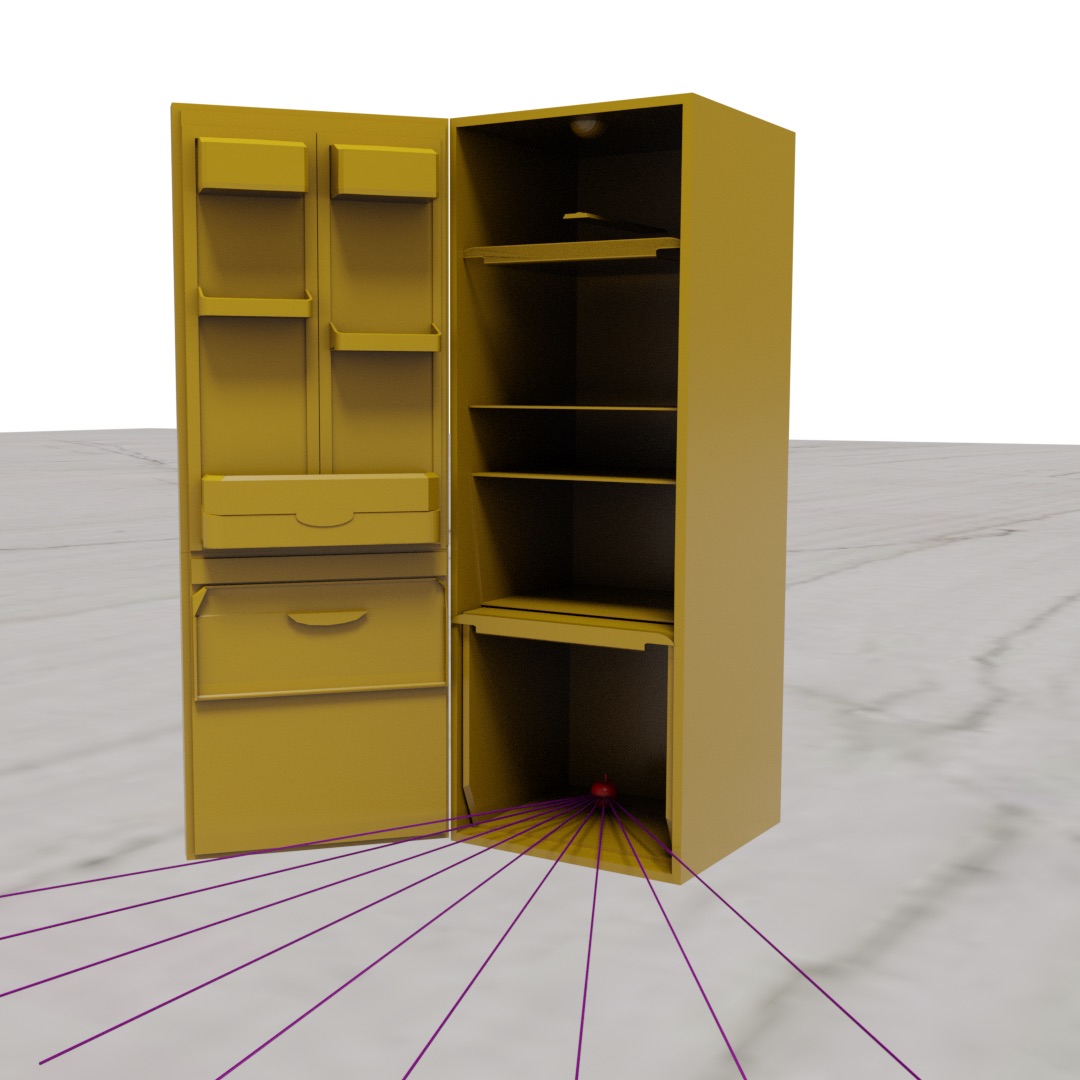}
        \includegraphics[width=0.3\linewidth]{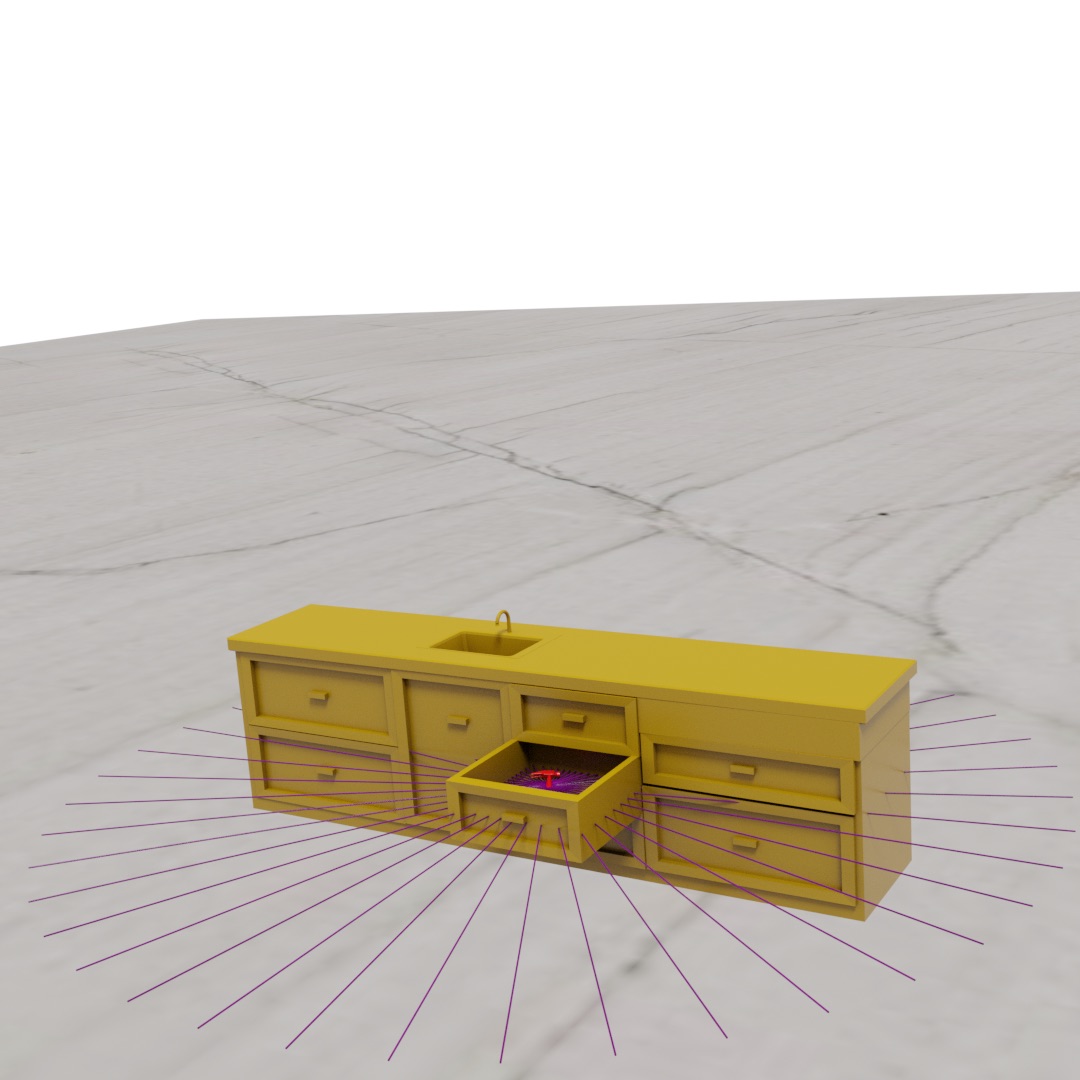}
        \includegraphics[width=0.3\linewidth]{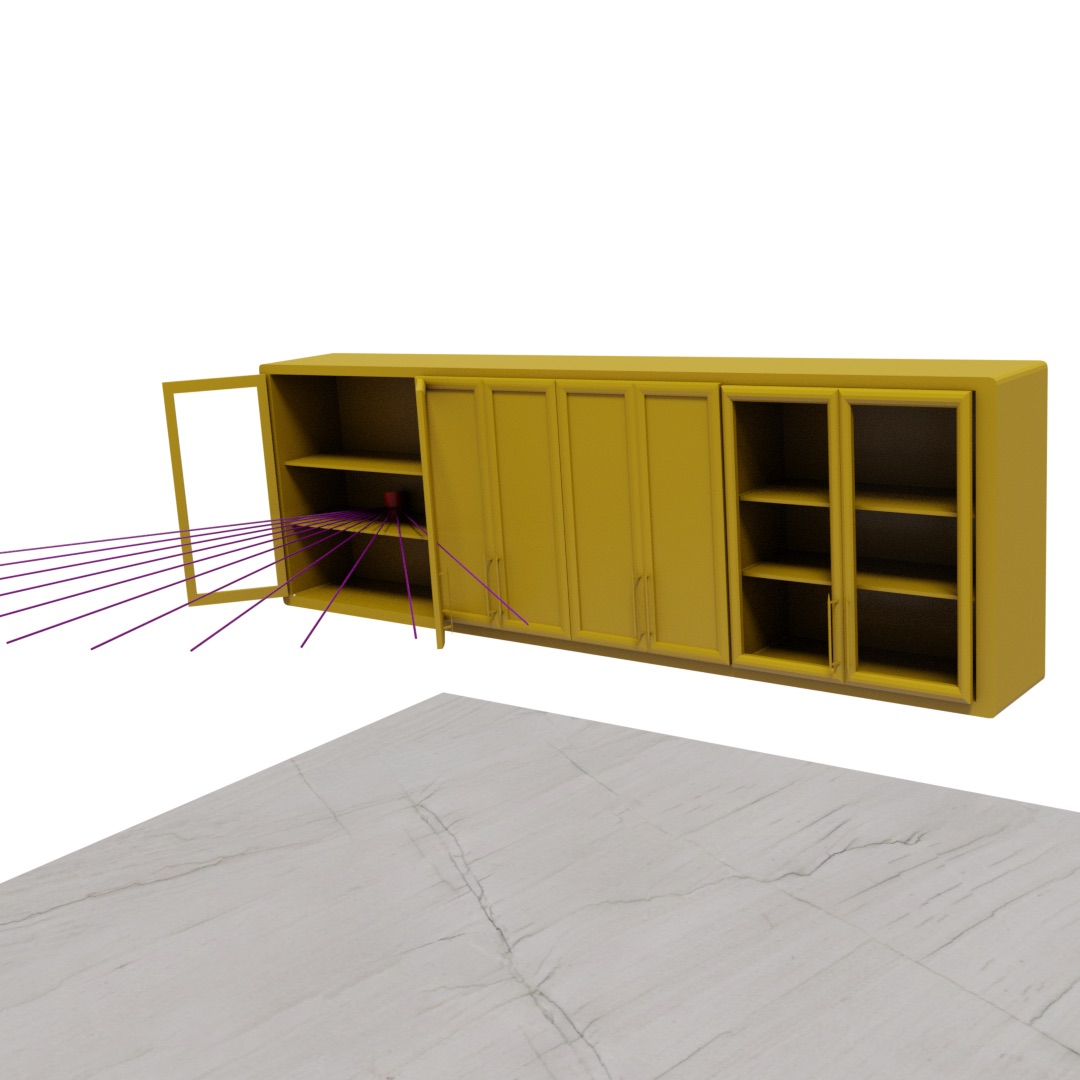}
    \end{adjustbox}

    \begin{adjustbox}{max width=\resizeSTEPS\textwidth}
        \raisebox{3em}{\makebox[0.3\linewidth][l]{\textbf{F.}}} 
        \hspace{-5em} 
        \includegraphics[width=0.3\linewidth]{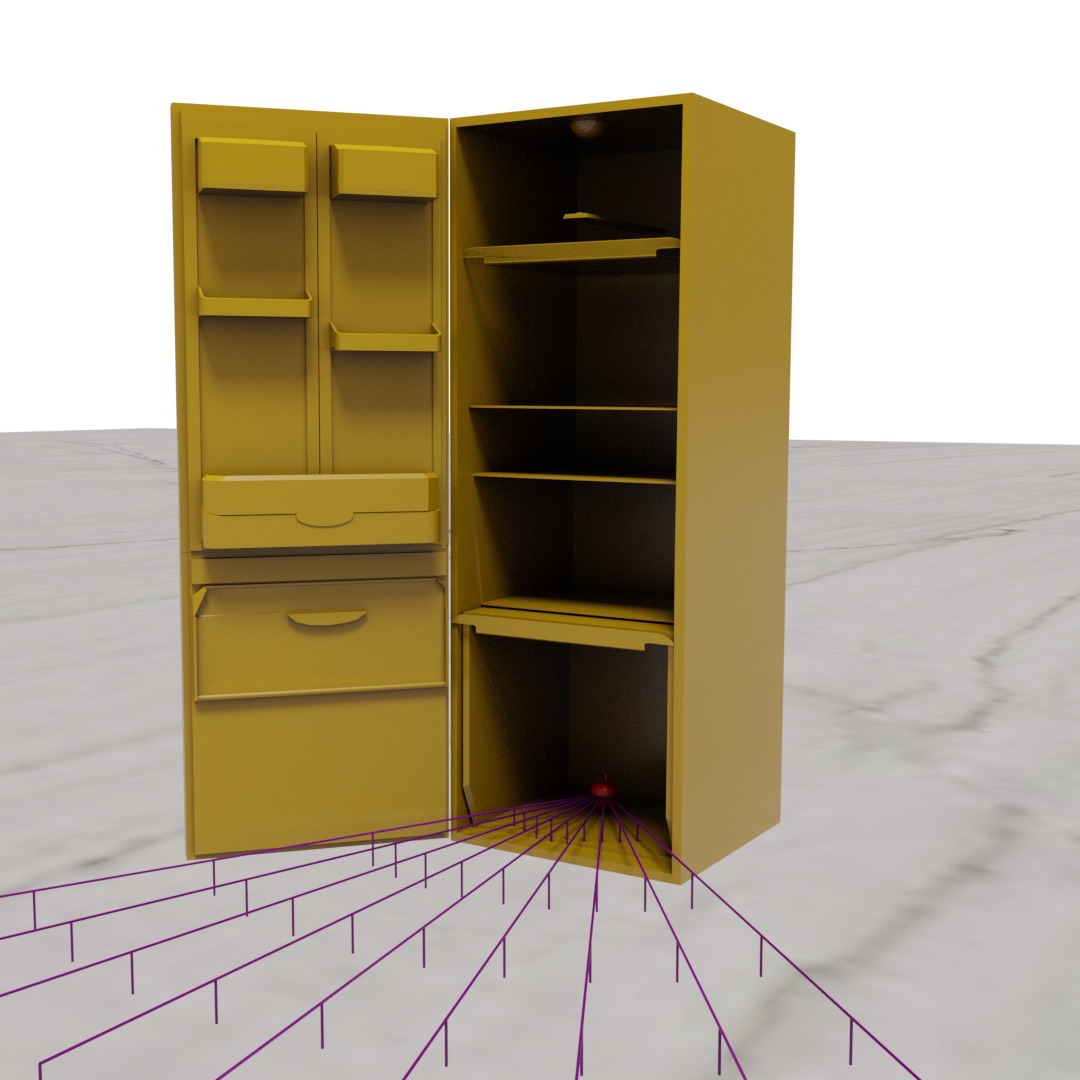}
        \includegraphics[width=0.3\linewidth]{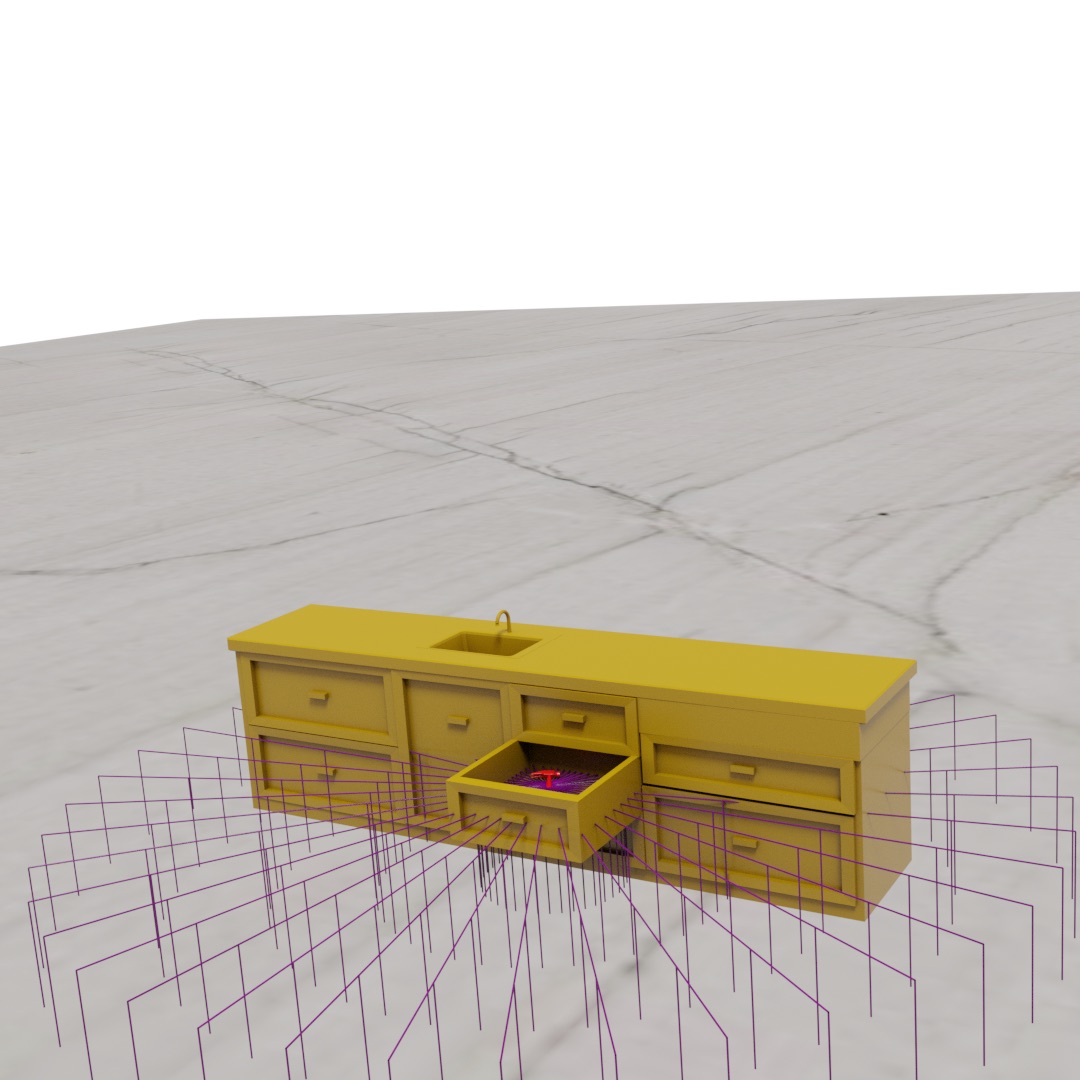}
        \includegraphics[width=0.3\linewidth]{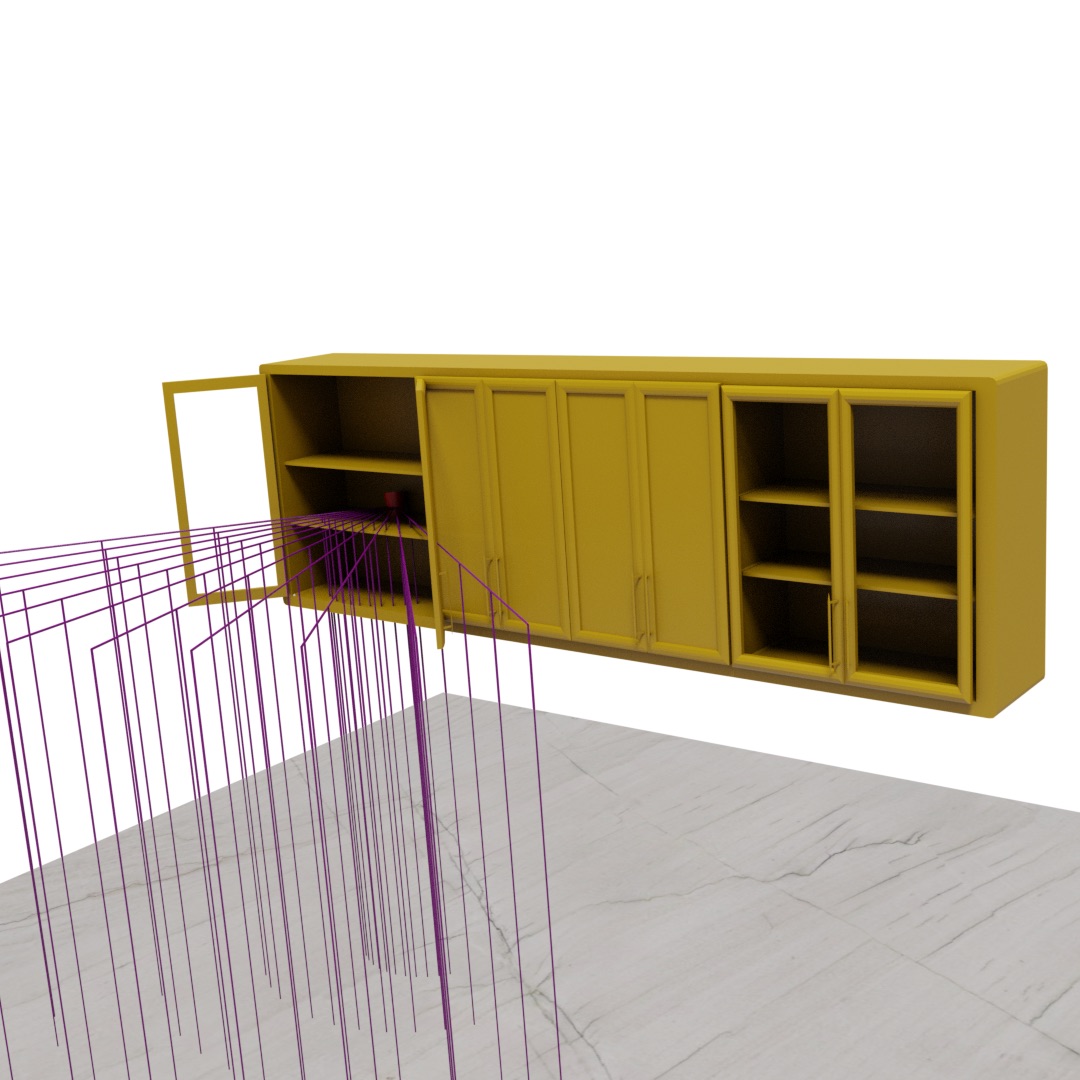}
    \end{adjustbox}

    \begin{adjustbox}{max width=\resizeSTEPS\textwidth}
        \raisebox{3em}{\makebox[0.3\linewidth][l]{\textbf{G.}}}
        \hspace{-5em} 
        \includegraphics[width=0.3\linewidth]{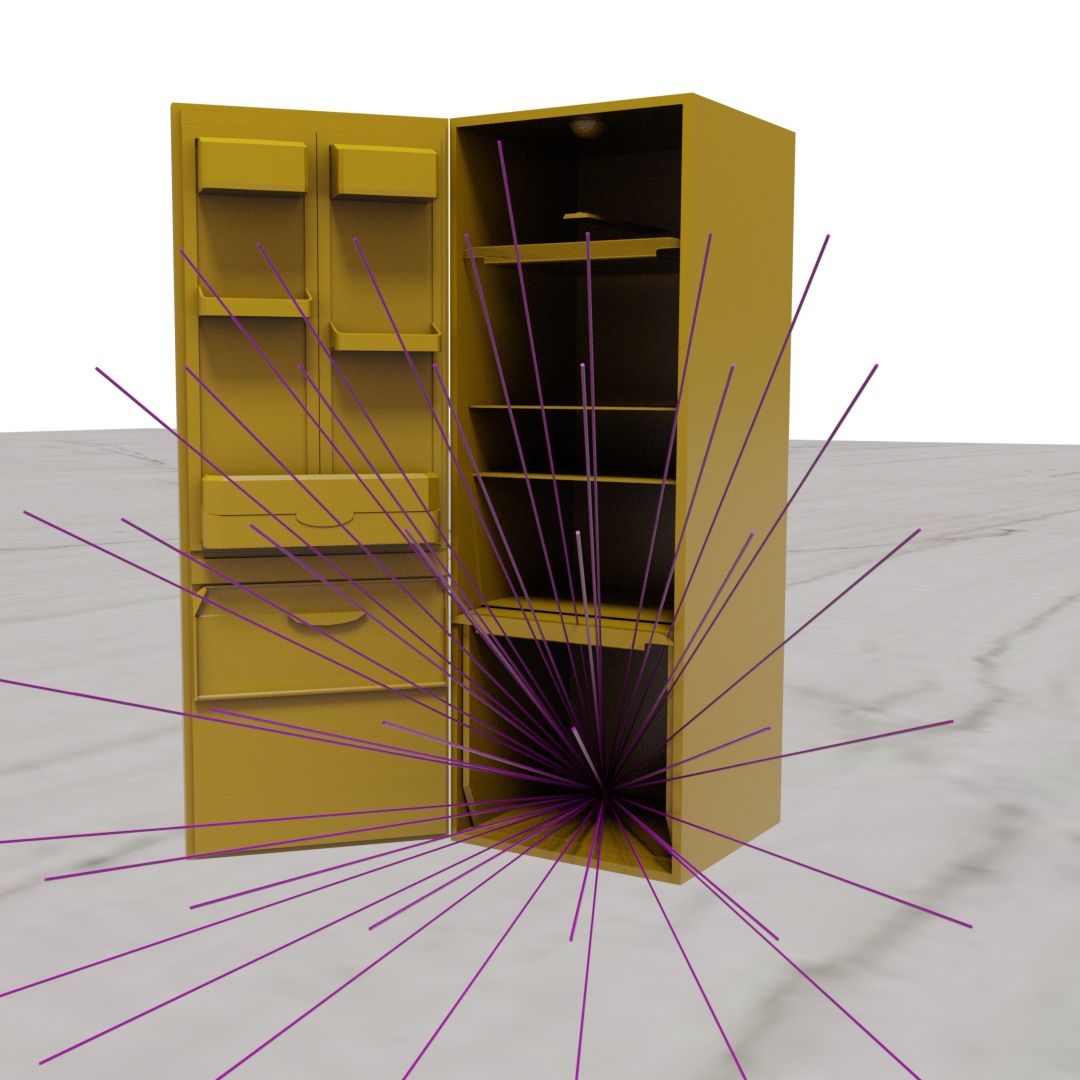}
        \includegraphics[width=0.3\linewidth]{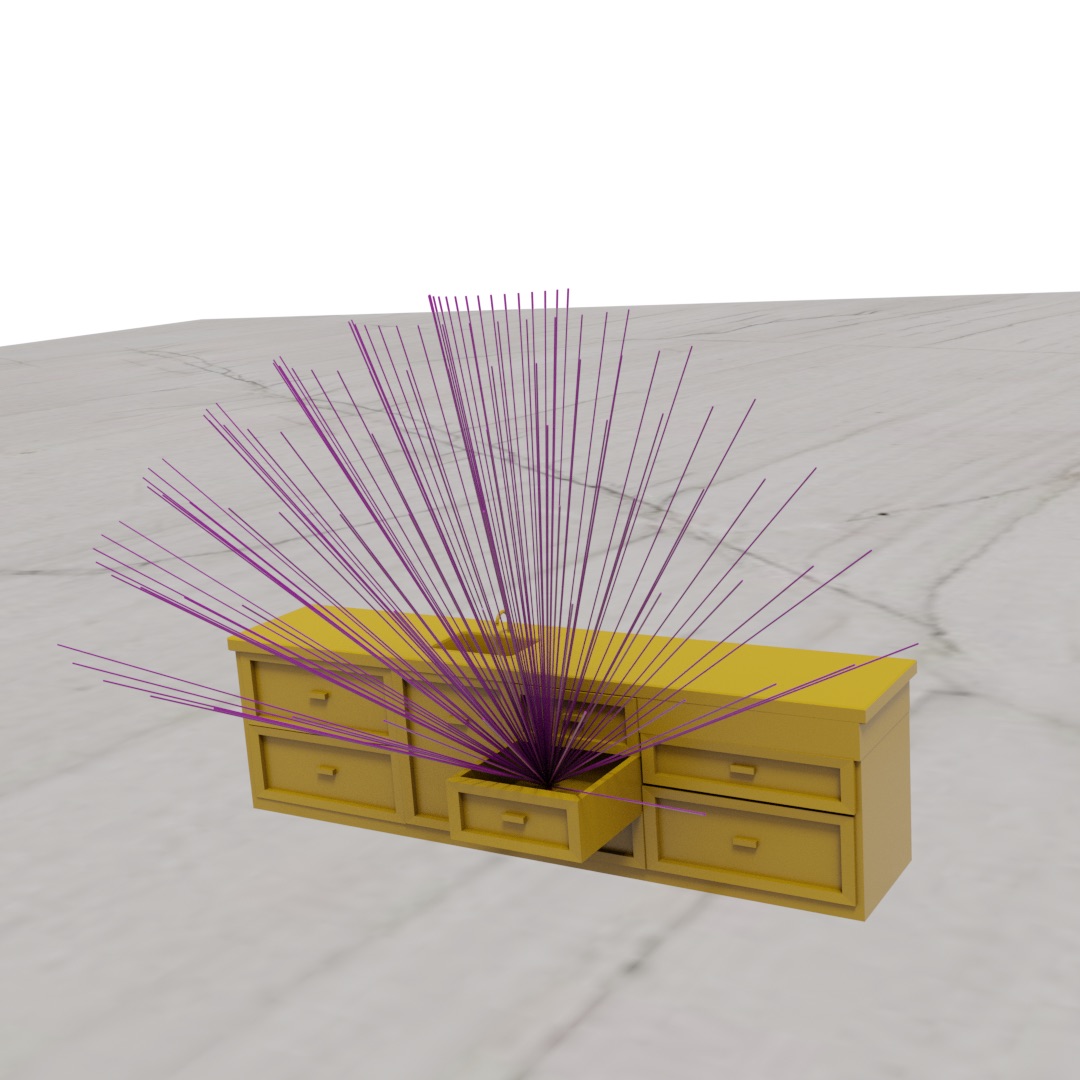}
        \includegraphics[width=0.3\linewidth]{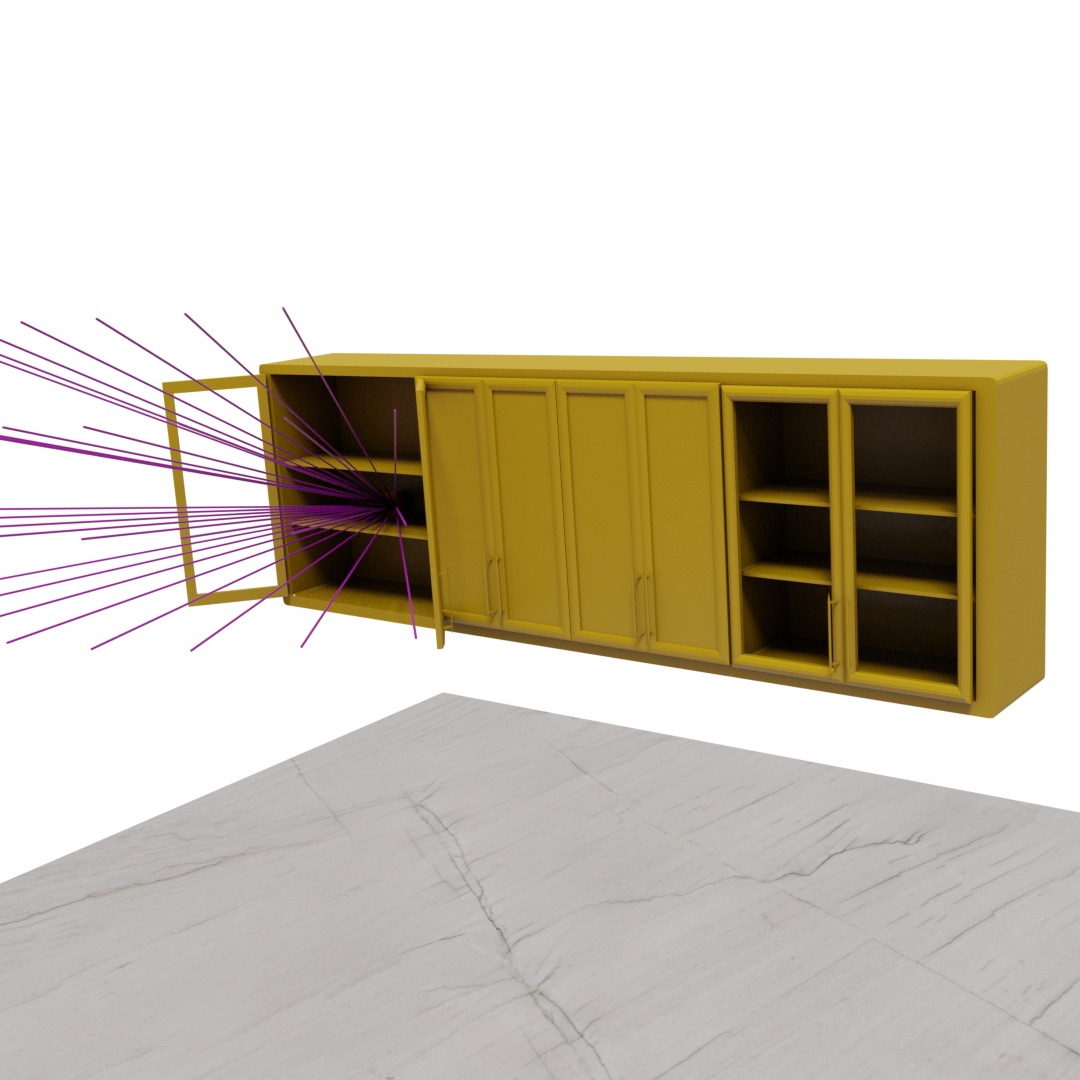}
    \end{adjustbox}
    \vspace{-1.5 em}
    \caption{
            \reachingField steps for three object ``heights.'' 
    }
    \label{fig:supmat_ray_casting_examples}
\end{figure}

\subsection{\creach}
\label{sec:supmat_creach}

\hspace{\parindent}
\textbf{Training datasets:} 
To train our \creach model we combine the \grab~\cite{grab} and \circleData~\cite{circle} datasets. 
\grab captures dexterous interactions, but is limited mainly to standing bodies.  
Instead, \circleData captures bodies without fingers, but has a wide range of ``reaching'' body poses, including kneeling down and stretching up. 
To get the best of both worlds, namely rich body poses with dexterous fingers, we combine \grab and \circleData. 

Note that these datasets capture both left- and right-arm interactions. 
To train a single network for 
these, we need to identify the interaction ``handedness'' 
in the data, since relevant annotations are missing.
We achieve this by 
exploiting the hand-eye coordination taking place for grasping. 
To this end, for each body, we compute three vectors: 
the gaze vector, $g_{dir}$, via two pre-annotated vertices at the nose tip and the back of the head, 
and two vectors, $d_{erw}$ and $d_{elw}$, from the glabella (point between the two eyes) to the right and left wrist joints, respectively. 
We will release the annotated vertex IDs.
Then, interaction ``handedness'' is defined as:
\begin{equation}\label{eq:supmat_hand_interaction}
\begin{aligned}
    \widehat{g_{dir}d_{erw}} &\leq \widehat{g_{dir}d_{elw}}: \textbf{\rhand interaction} 
    \\
    \widehat{g_{dir}d_{erw}} &> \widehat{g_{dir}d_{elw}}: \textbf{\lhand interaction}
\end{aligned}
\end{equation}
Since \rhand interactions are much more frequent than \lhand ones in \grab and \circleData, 
we mirror all data for balancing the ``handedness.'' 

\subsection{\cgrasp}
\label{sec:supmat_cgrasp}

\cgrasp is a CVAE that exploits an \IAfeatures and a vector denoting the desired palm \direction. 
Specifically, the \IAfeatures is input to the encoder, while the \direction vector conditions the decoder. 
\Cref{fig:supmat_cgrasp_controllability} shows grasps generated by our \cgrasp for the $6$ test objects of the \grab dataset, along with the corresponding \direction vectors used as condition. 
In all cases, the \direction of the generated grasp ``agrees'' with the conditioning direction vector. 
\Cref{fig:supmat_interfield} shows a visual example of the \IAfeatures. 
\Cref{fig:supmat_cgrasp_qualitative_results} shows qualitative results of \cgrasp compared to existing methods. 

\begin{figure*}[ht]
    \centering
    \includegraphics[width=\linewidth]{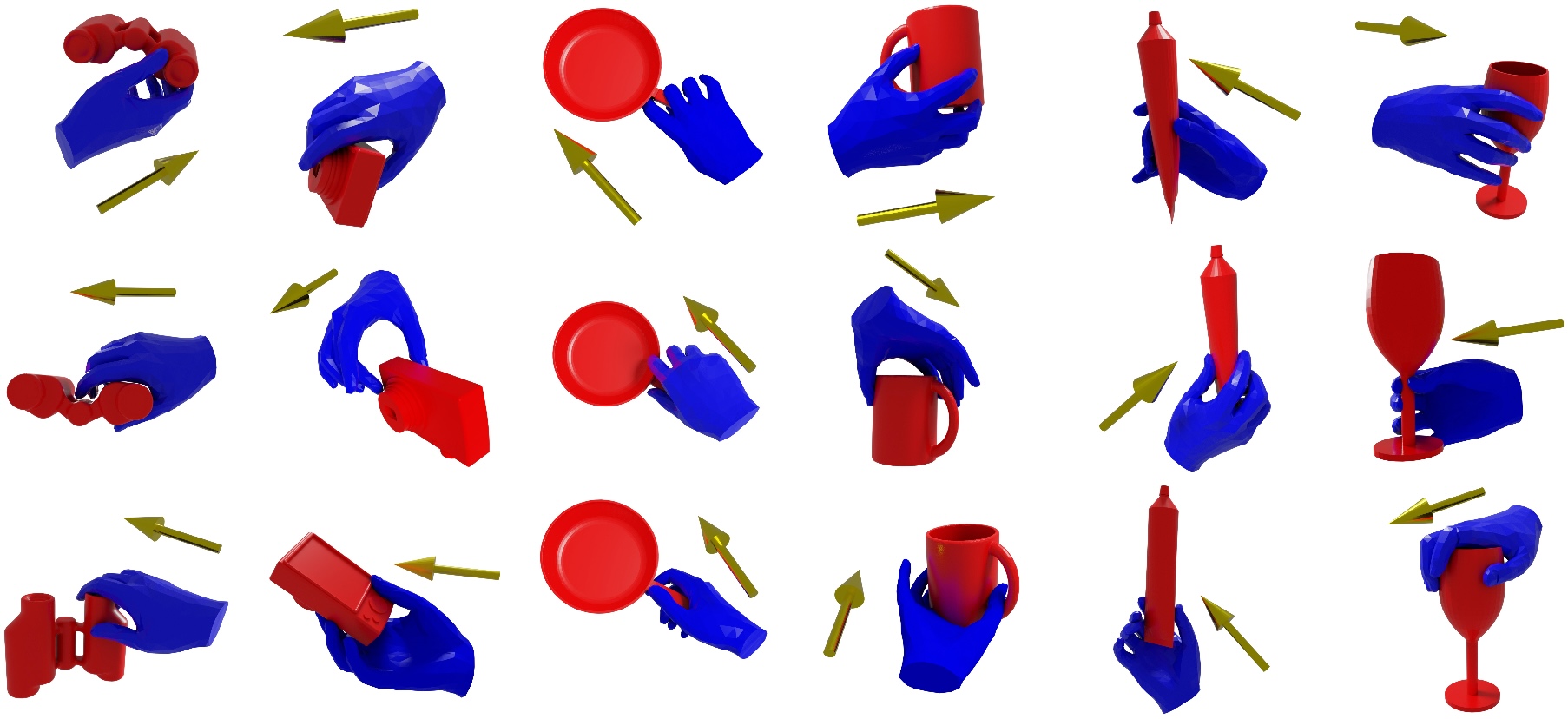}
    \caption{
        Qualitative results of \cgrasp, along with the given direction vectors. 
        For the six test objects in the \grab dataset, we generate three grasps per object using different direction vectors as conditions. 
        The conditioning direction vectors are shown by gold arrows.  
    }
     \label{fig:supmat_cgrasp_controllability}
\end{figure*}
\begin{figure}
    \centering
    \includegraphics[width=0.7 \linewidth]{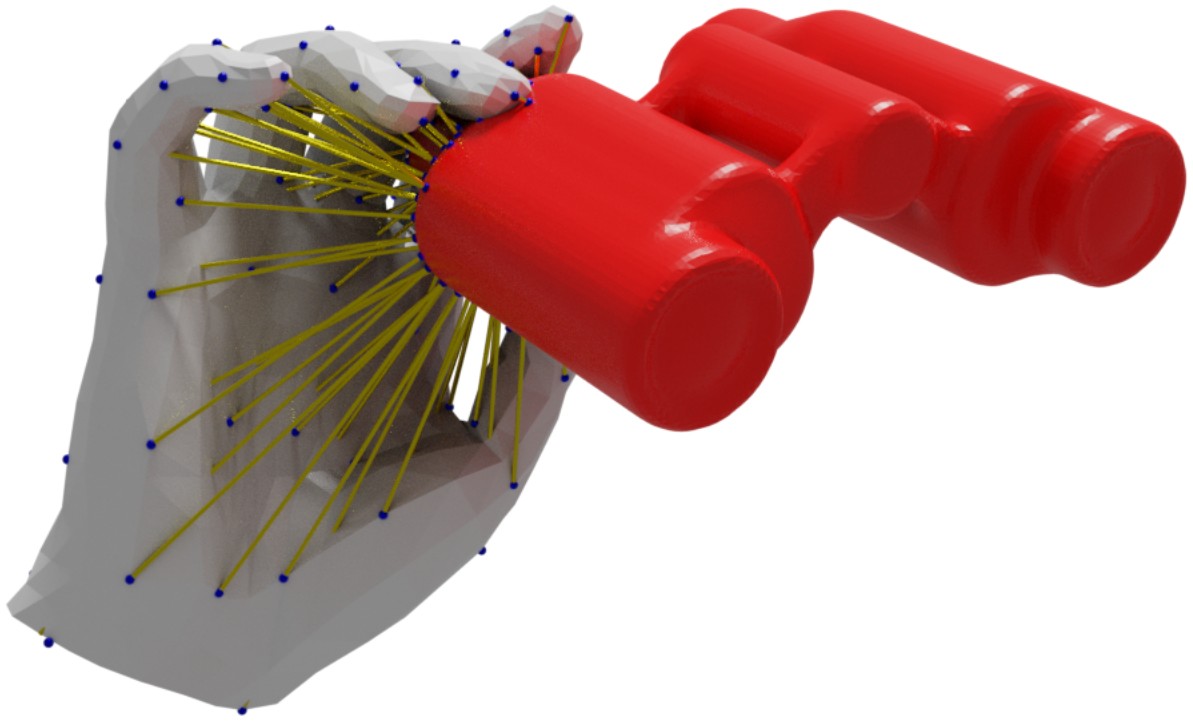}
    \vspace{-0.5 em}
    \caption{
            Visual representation of the \IAfeatures. 
            We depict with \textcolor{blue}{blue} spheres the $99$ sampled ``interaction'' hand vertices, $\vertex_{\hand, i}^\inter 
            \text{,~} i \in \{1, \dots, 99\}$, evenly-distributed across \mano's inner palm/finger surface. 
            With 
            \textcolor{olive}{olive-color} 
            lines we depict the \threeD vectors, $\ifeat$, that encode the distance and direction from the sampled hand vertices, $\vertex_{\hand}^\inter$, to their closest object vertices, $\vertex_\obj'$. 
    }
    \label{fig:supmat_interfield}
    \vspace{+2.0 em}
\end{figure}

\begin{figure*}
\centering
\includegraphics[width=0.99 \textwidth]{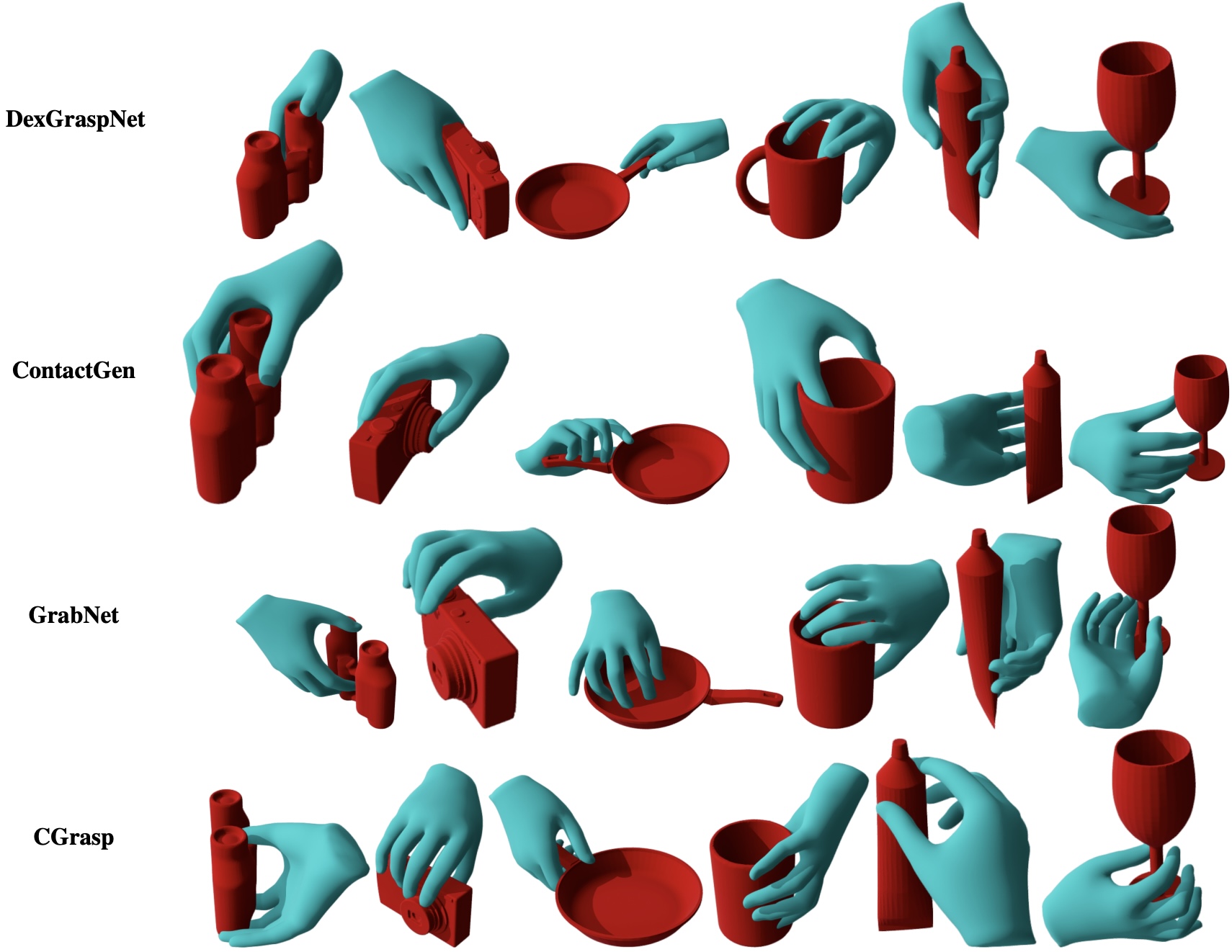}
\caption{
    Comparison of hand-only grasp synthesis models: \dexgraspnet~\cite{dexgraspnet}, \contactgen~\cite{contactgen}, \grabnet~\cite{grab}, and \cgrasp (ours). 
    Each row shows grasps generated by a different model for the same object set (binoculars, camera, frying pan, mug, toothpaste, wineglass).
}
\label{fig:supmat_cgrasp_qualitative_results}
\vspace{+1.0em}
\end{figure*}

\subsection{\cWGrasp~-- Optimization details}
\label{sec:supmat_optimization}

Our goal is to optimize over the \smplx pose, translation, and global-orientation parameters, so that the 
hand of the body aligns with the ``guiding'' 
hand generated by \cgrasp, while 
contacting the ground without penetrating the receptacle. 
To this end, we use the objective function of \cref{opt_fun}:
\begin{equation*}
\begin{aligned}
    \mathcal{L}_{opt} = 
    &\lambda_p \mathcal{L}_{p} +  \lambda_{grd} \mathcal{L}_{grd} + \lambda_{\theta} \mathcal{L}_{\theta} + \\
    &\lambda_{g}\mathcal{L}_{g} + \lambda_{hm}\mathcal{L}_{hm} + \lambda_{reg}\mathcal{L}_{reg}
    \text{.}
\end{aligned}
\end{equation*}
The term $\mathcal{L}_{p}$ is a penetration loss between the body and the receptacle, consisting of two terms:
\begin{equation}\label{eq:supmat_penetration_loss}
\begin{aligned}
    \mathcal{L}_{p} 
    &= \mathcal{L}_{p_{inter}} +  \mathcal{L}_{p_{con}}
    \text{,}
    \\
    \mathcal{L}_{p_{inter}} 
    &=\frac{1}{N_{\mathcal{V}_{b}}}\sum_{i=1}^{N_{\mathcal{V}_{b}}}\mid min(0, d(\mathcal{V}_{b_i},\receptacle))\mid 
    \text{,}
    \\
    \mathcal{L}_{p_{con}} 
    &=\frac{1}{N_{\mathcal{V}_{b}}}\sum_{i=1}^{N_{\mathcal{V}_{b}}^{disc}}d(\mathcal{V}_{b_i}^{disc},\receptacle))
    \text{,}
\end{aligned}
\end{equation}
where 
$d(\cdot)$ denotes the signed distance function between the vertices of the body, $\mathcal{V}_{b}$, and the receptacle mesh, $\receptacle$. 
The first term penalizes vertices that penetrate the receptacle $\receptacle$.  
The second term penalizes the body vertices that get ``disconnected'' from the rest of the body when the latter gets ``intercepted'' by the penetrated receptacle \cite{flex}. 

For the ground loss, $\mathcal{L}_{grd}$, we first find the height, $h(\mathcal{V}_{b_i})$, of body vertices \wrt the ground plane. 
Then we have two cases. 
For vertices that penetrate the ground 
we use the under-ground term of \cite{tripathi2023ipman}:
\begin{equation}
    \mathcal{L}_{grd} = 
    \beta_{1} tanh\left(\frac{h\left(\mathcal{V}_{b_i}\right)}{\beta_{2}}\right)^2
    \text{,}
    \quad 
    \text{for~} h(\mathcal{V}_{b_i})<0
    \text{,}
    \label{eq:supmat_floor_loss}
\end{equation}
where $\beta_1=1$ and $\beta_2=0.15$, as in \cite{tripathi2023ipman}. 
For vertices above the ground (\ie, for missing contacts), we use:
\begin{equation}
    \mathcal{L}_{grd} =
    \mid min({\mathcal{V}_{b_i}}_{z}) \mid 
    \text{,}
    \quad 
    \text{for~} 
    h(\mathcal{V}_{b_i})>0
    \text{,}
    \label{eq:supmat_floor_loss_alter}
\end{equation}
where $i=1\cdots N_{\mathcal{V}_{b}}$. 
Both \cref{eq:supmat_floor_loss} and \cref{eq:supmat_floor_loss_alter} are responsible for keeping bodies on the ground. 

To ensure that the generated body has ``eye contact'' with the object of interaction we use the gaze loss of \mbox{GOAL} \cite{goal}. 
To this end, we annotate two vertices $\mathcal{A}$ and $\mathcal{B}$ on the head, namely vertex $\mathcal{A}$ at the back of the head, and vertex $\mathcal{B}$ lying between the eyes. 
The \threeD position of the object $\mathcal{O}$ is known. 
Thus, we 
define the vectors $\overrightarrow{\mathcal{A}\mathcal{O}}$ and $\overrightarrow{\mathcal{B}\mathcal{O}}$ and specify the gaze loss as their in-between angle: 
\begin{equation}
    \mathcal{L}_{g} =
    cos^{-1}\left(\frac{\overrightarrow{\mathcal{A}\mathcal{O}}\quad\overrightarrow{\mathcal{B}\mathcal{O}}}{\mid\overrightarrow{\mathcal{A}\mathcal{O}}\mid\mid\overrightarrow{\mathcal{B}\mathcal{O}}\mid}\right).
    \label{eq:supmat_gaze}
\end{equation}

As we optimize over the \smplx pose parameters,  
we need to encourage the predicted poses, $\hat{\theta}$, to remain on the manifold of our \creach, so that we produce realistic human bodies. 
To this end, 
we employ the regularizer:
\begin{equation}\label{eq:supmat_pose_loss}
\mathcal{L}_{\theta} =\lVert\theta - \hat{\theta}\rVert^2   \text{.}
\end{equation}

Even after incorporating 
$\mathcal{L}_{\theta}$ 
into our optimization, we still observe some unrealistic results. In the left part of \cref{fig:supmat_problematic_cases}, we illustrate such an example, where the body ``tilts'' unrealistically toward the object.  
To account for this, we introduce an additional regularization loss term: 
\begin{equation}\label{eq:supmat_reg_loss}
\mathcal{L}_{reg} = cos^{-1}(\overrightarrow{\mathcal{F}\mathcal{P}}\hat{\overrightarrow{\mathcal{F}\mathcal{P}}})
\text{,}
\end{equation}
where $\mathcal{F}$ is the center of the feet, defined as the midpoint between the right and left ankle joints, and $\mathcal{P}$ the pelvis joint. 
By minimizing \cref{eq:supmat_reg_loss}, we prevent deviations from the initial posture of our \creach, eliminating ``tilting'' effects. 

To align 
the 
hand of the whole body (produced by \creach) with the one of the ``guiding'' hand grasp (produced by \cgrasp)
we use:
\begin{equation}
\mathcal{L}_{hm} = \lVert\mathcal{V}_{hm} - \hat{\mathcal{V}}_{b}^{hm}\rVert^2 + \lambda_{wrist}\lVert\mathcal{V}_{wrist} - \hat{\mathcal{V}}_{b}^{wrist}\rVert^2
\text{,}
\label{eq:supmat_hand_loss} 
\end{equation}
where
$\mathcal{V}_{hm}$ are the hand vertices and 
$\mathcal{V}_{wrist}$ are the wrist vertices of the output hand mesh of \cgrasp, 
and 
$\hat{\mathcal{V}}_{b}^{hm}$ and 
$\hat{\mathcal{V}}_{b}^{wrist}$ 
are the corresponding hand and wrist vertices of the optimized body's hand. 

We use the Adam optimizer with a learning rate of $0.01$. 
Our optimization has two stages. 
First, 
we optimize over all above losses for $800$ iterations. 
Then,
we exclude the penetration loss,
and use the rest of the losses to optimize over the pose parameters of the 
shoulder, elbow, and wrist that correspond to the matching hand. 
The main goal of the second optimization step is to 
refine the hand grasp.

\section{\cWGrasp~-- Perceptual Study}
To evaluate \cWGrasp we conduct a perceptual study. 
\Cref{fig:supmat_perceptual_description} provides the task description that is shown to participants. 
\Cref{fig:supmat_perceptual_supp_mat_results} shows some samples used in our study. 

\begin{figure}
    \centering
    \includegraphics[width=0.49 \linewidth]{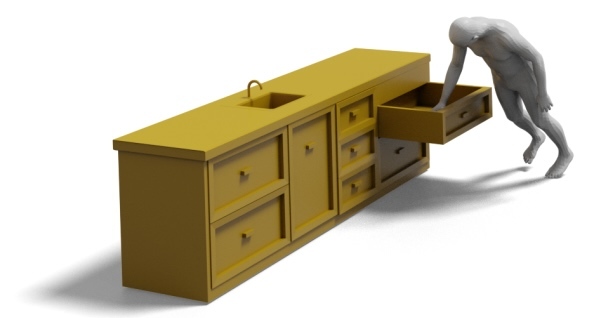}
    \includegraphics[width=0.49 \linewidth]{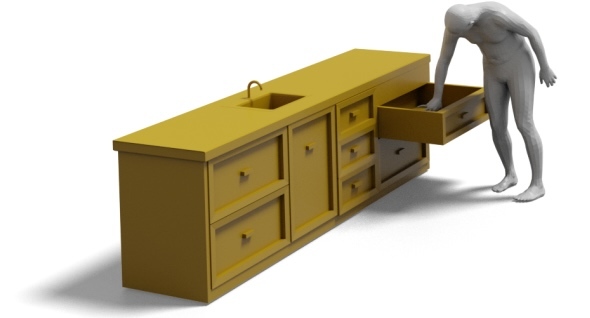}
    \vspace{-0.5 em}
    \caption{
                \textbf{Left:} 
                Failure case when we do not use our regularizer term $\mathcal{L}_{reg}$ in our loss function. 
                In some cases, when $\mathcal{L}_{reg}$ is not used, our model prefers to generate bodies that tilt unrealistically toward the object to prevent penetrations with the receptacle $\mathcal{M}$, while reaching the object with the hand. 
                \textbf{Right:} 
                The corresponding output of \cWGrasp when we use our $\mathcal{L}_{reg}$ loss term. 
    }
    \label{fig:supmat_problematic_cases}
    \vspace{-0.5 em}
\end{figure}

\begin{figure}
    \centering
    \vspace{-0.5 em}
    \includegraphics[width=0.99\linewidth]{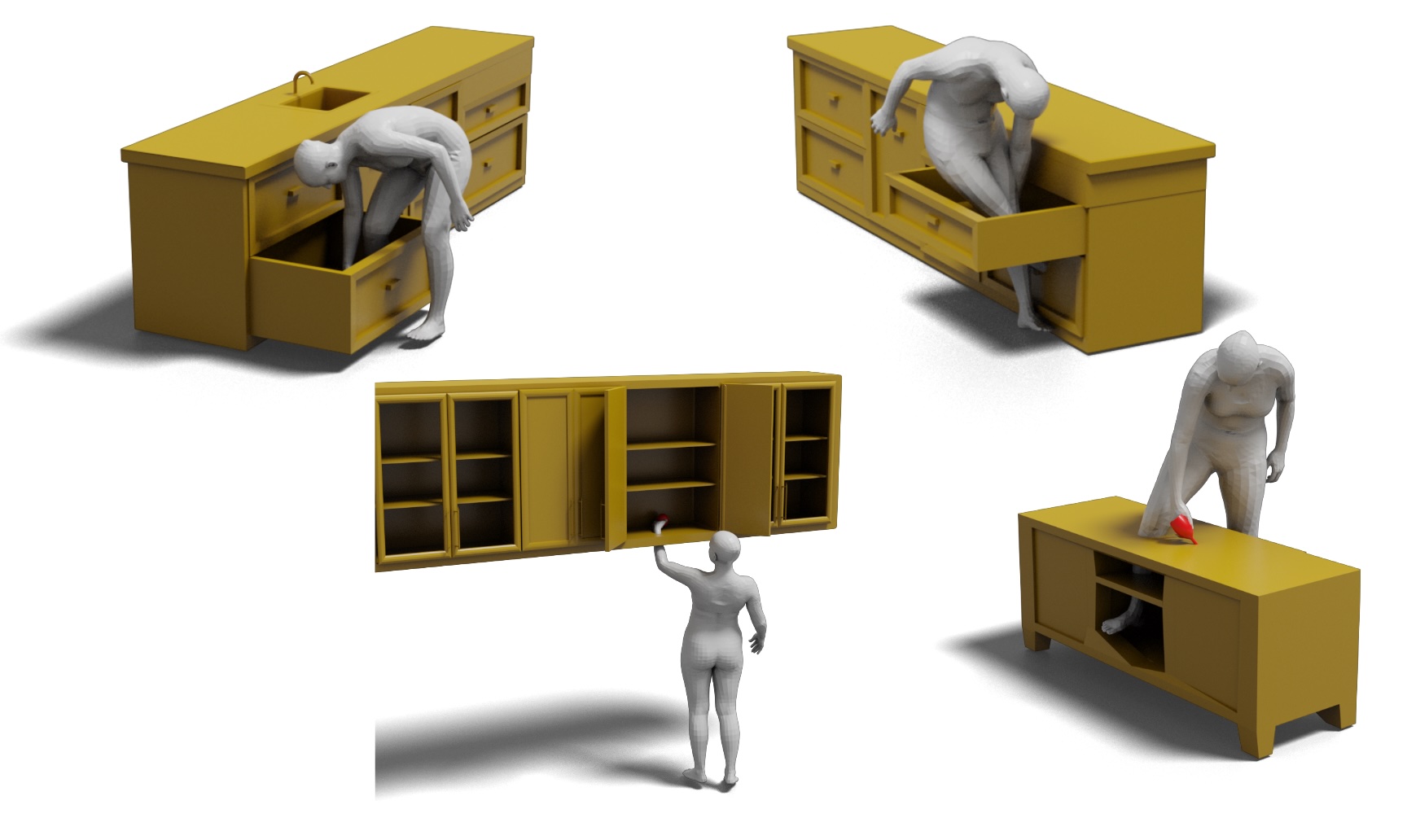} 
    \vspace{-1.5 em}
    \caption{
        \textbf{Failure cases} of our \cWGrasp method for various receptacles. 
        Our framework can fail for two reasons, namely due to penetrations between the body and the receptacle, or due to penetrations between the grasping hand and the object.
    }
    \label{fig:supmat_cwgrasp_failures}
\end{figure}

\begin{figure*}
    \centering  
    \includegraphics[width=1.45 \columnwidth]{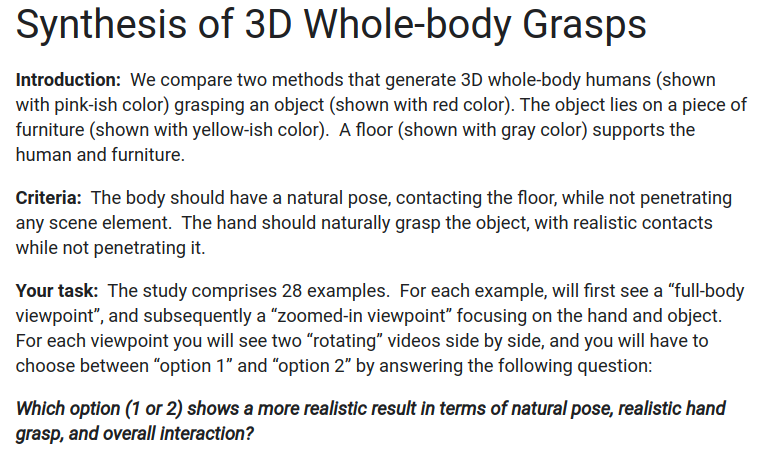}
    \vspace{-0.5 em}
         \caption{
            \textbf{Perceptual study protocol.}
            We ask $35$ participants to observe grasps generated by two methods for $28$ object-and-receptacle configurations, and to specify which grasp is more realistic in terms of natural pose, realistic hand grasp, and overall interaction realism. 
            This shows the task instructions shown to participants. 
            See also the samples of \cref{fig:supmat_perceptual_supp_mat_results}.
        }
         \label{fig:supmat_perceptual_description}
\end{figure*}

\begin{figure*}
    \centering  
    \includegraphics[width=\textwidth]{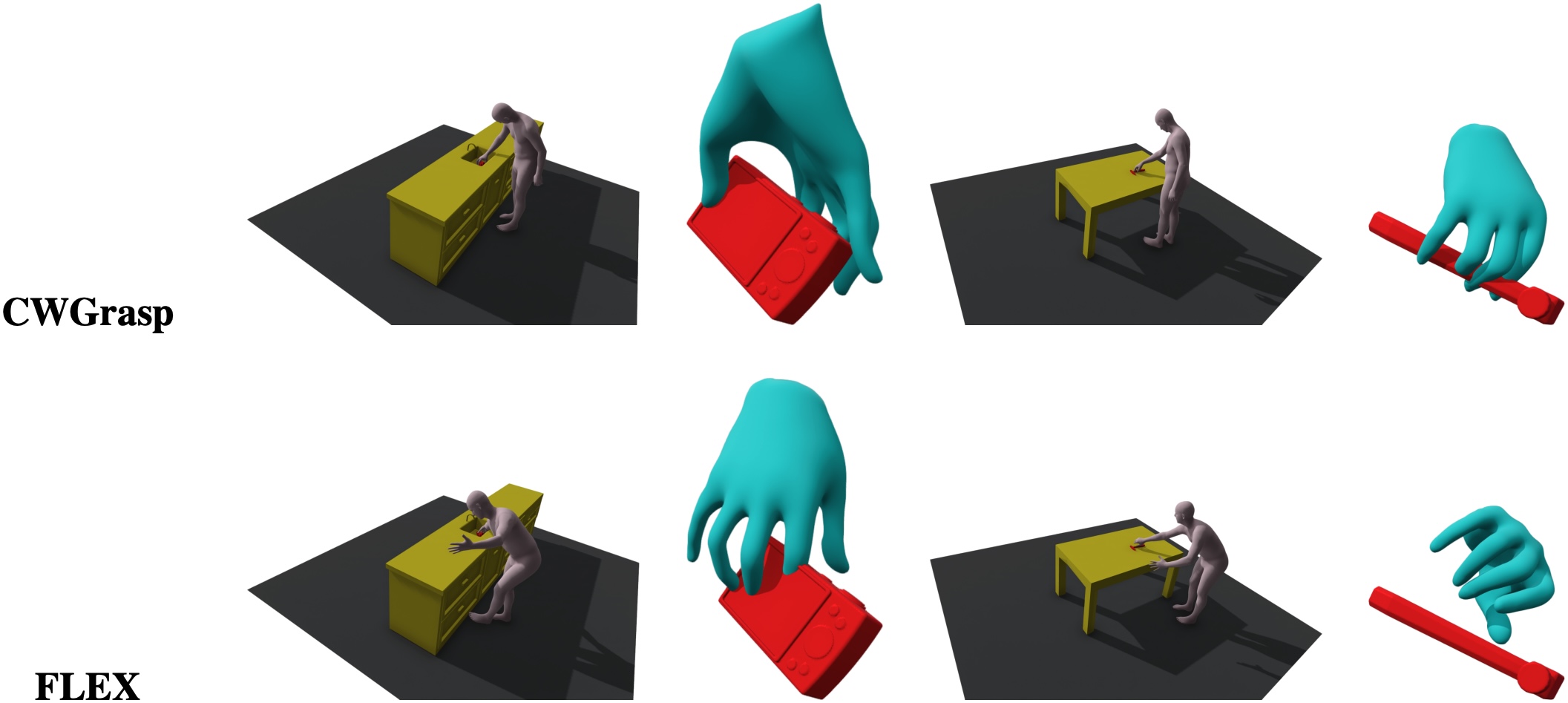}
    \vspace{-0.5em}
    \caption{
        \textbf{Perceptual study samples}. 
        The first row shows grasps generated by our \cWGrasp method, while the second row ones generated by \flex~\cite{flex}. 
        For each sample, we show a full-body view and a close-up view onto the hand and object. 
    }
    \label{fig:supmat_perceptual_supp_mat_results}
\end{figure*}

\pagebreak

\section{\cWGrasp~-- Qualitative Evaluation}
\label{sec:supmat_cwgrasp_qual}

Here we provide additional qualitative comparisons between \flex~\cite{flex} and our \cWGrasp framework; 
see 
\cref{fig:supmat_cwgrasp_qual1}, \cref{fig:supmat_cwgrasp_qual2}, and \cref{fig:supmat_cwgrasp_qual3}. 

For each example, we depict both a full-body view and a close-up view onto the hand and object. 
Often \flex bodies miss contacting the object,
or 
approach the object from unrealistic
directions or have unrealistic
orientations (\eg, they do not ``look" at the object). 
Instead, \cWGrasp produces significantly more realistic bodies and grasps, a finding supported also by our perceptual study. 

Note that only \cWGrasp generates both right- and left-hand interactions; for examples of the latter see \cref{fig:supmat_cwgrasp_qual_left}.

\Cref{fig:before_after_opt} shows results before and after optimization; we observe that optimization enhances realism significantly. 

\zheading{Failure cases}
In \cref{fig:supmat_cwgrasp_failures} we provide failure cases of our \cWGrasp method. 
The reaching arm and hand look plausible, except for the bottom-left case where the arm penetrates the receptacle. 
The latter shows that occasionally sampling our \reachingField might fail, so subsequent optimization can be trapped in a local minimum, however, empirically, this doesn't happen often. 
In all other cases, the body or legs get trapped in a local minimum. 
This could be tackled with an involved modeling of the full scene, but this is out of our scope here, so we leave it for future work. 

\newcommand{\captionCWGraspVsFLEX}{Qualitative comparison of our \cWGrasp method against \flex~\cite{flex}. 
The \textbf{first} and \textbf{third} rows show results generated by \cWGrasp, while the \textbf{second} and \textbf{fourth} ones show results generated by \flex, for the same object and receptacle configurations. 
For each grasp we depict both the full-body (with \textcolor{gray}{gray} meshes) and ``scene,'' as well as a close-up onto the hand and object (with \textcolor{blue}{blue} and \textcolor{BrickRed}{red} meshes, respectively, from a side view). 
For \flex, out of all its generated samples, we visualize the ``best'' one, \ie, the one with the smallest total loss. 
In contrast, our \cWGrasp generates only one sample.}

\begin{figure*}
    \centering
    \includegraphics[width=0.99 \textwidth]{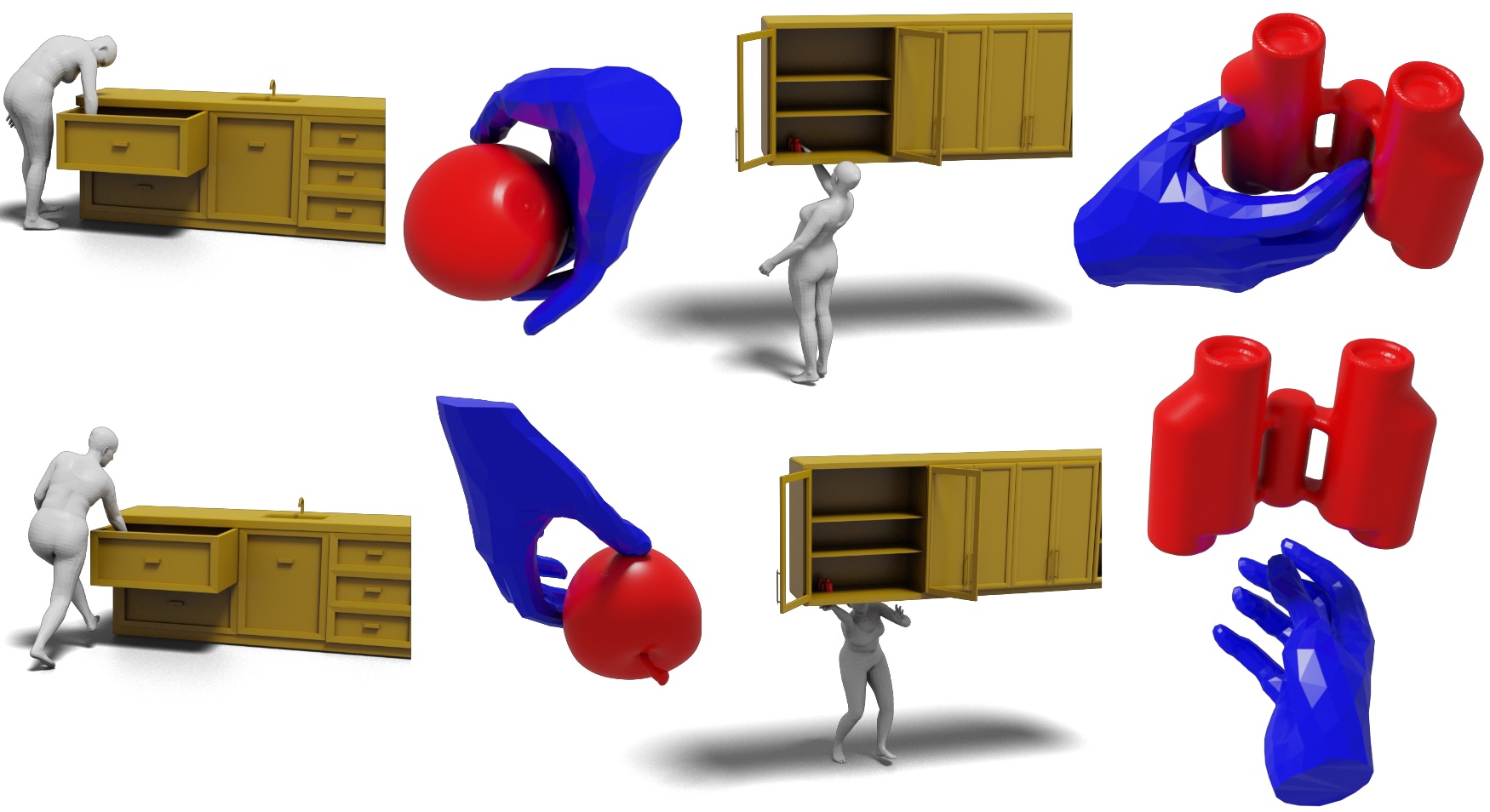}
    \\
    \includegraphics[width=0.99 \textwidth]{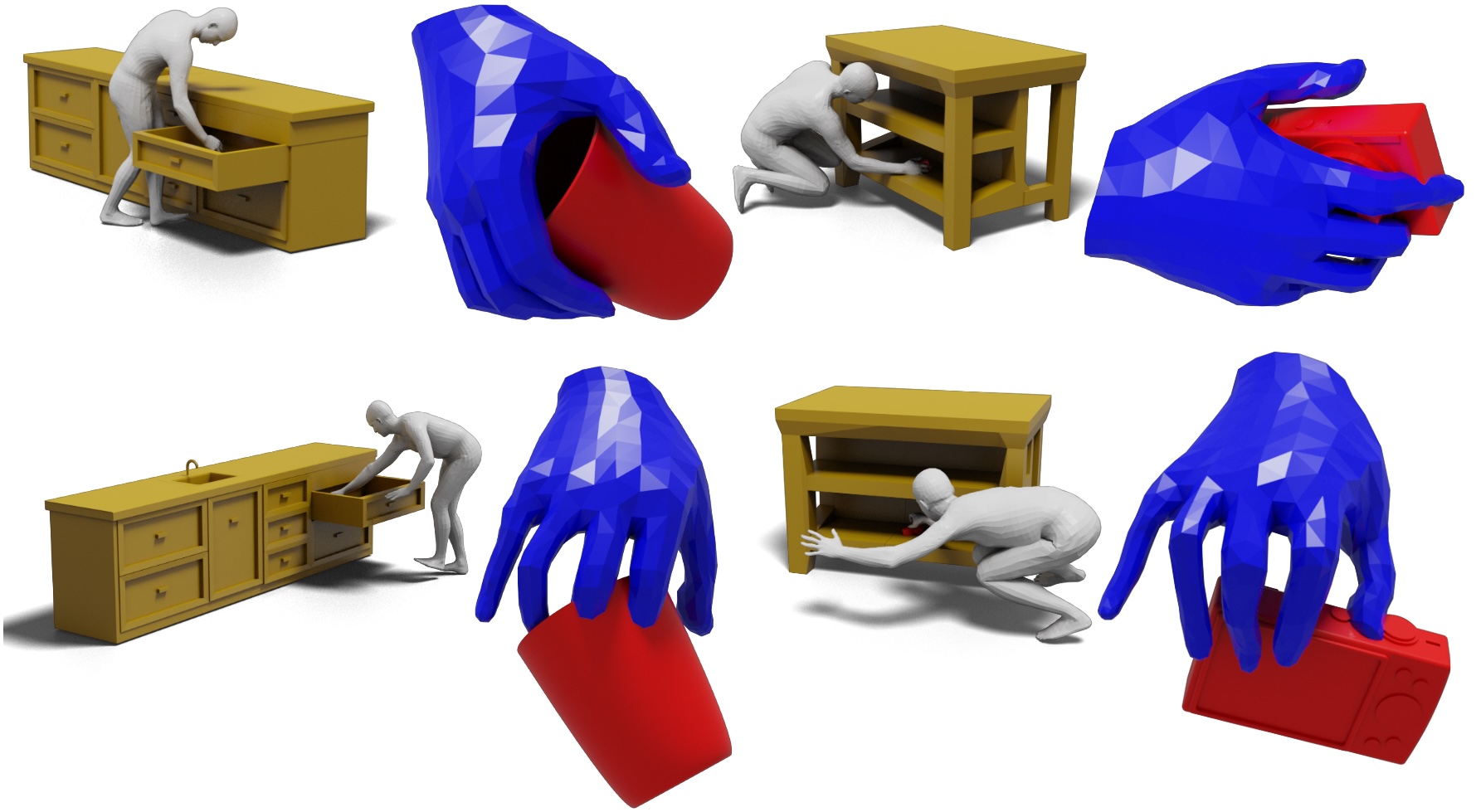} 
    \caption{\captionCWGraspVsFLEX}
    \label{fig:supmat_cwgrasp_qual1}
\end{figure*}
\begin{figure*}
    \centering
    \includegraphics[width=0.99 \textwidth]{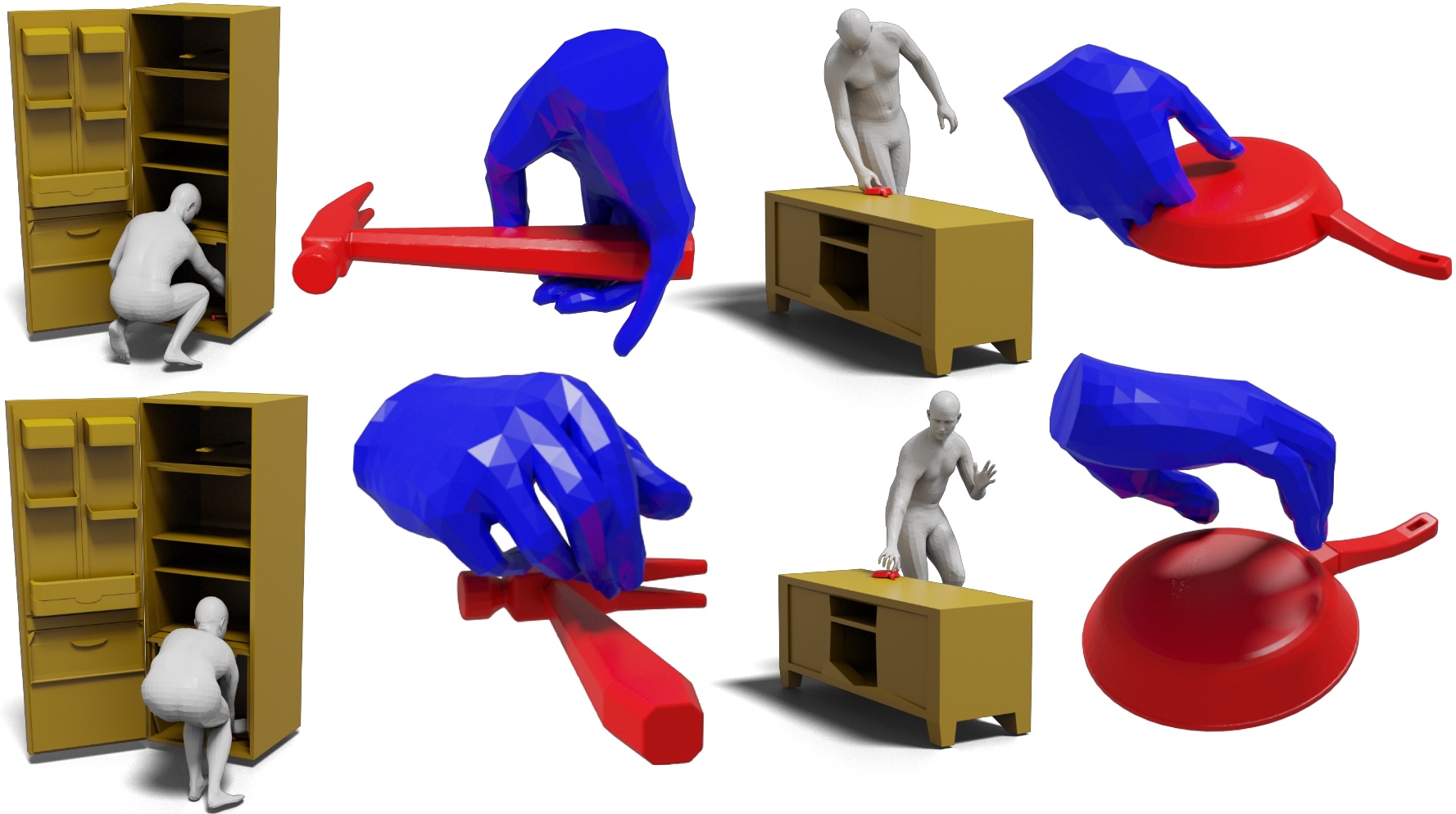}
    \\
    \includegraphics[width=0.99 \textwidth]{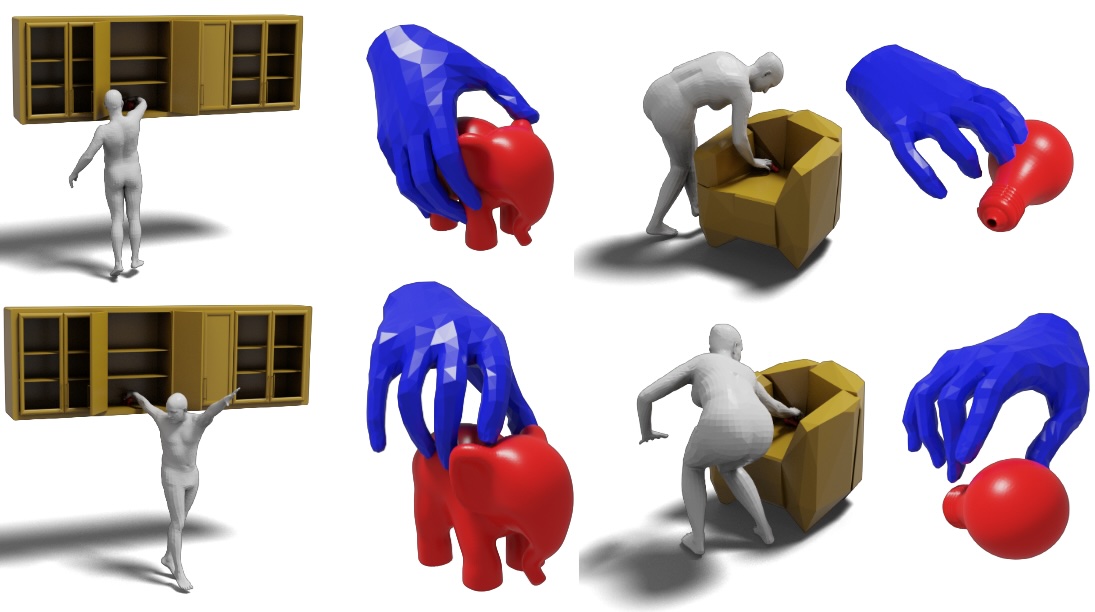} 
    \caption{\captionCWGraspVsFLEX}
    \label{fig:supmat_cwgrasp_qual2}
\end{figure*}
\begin{figure*}
    \centering
    \includegraphics[width=0.99 \textwidth]{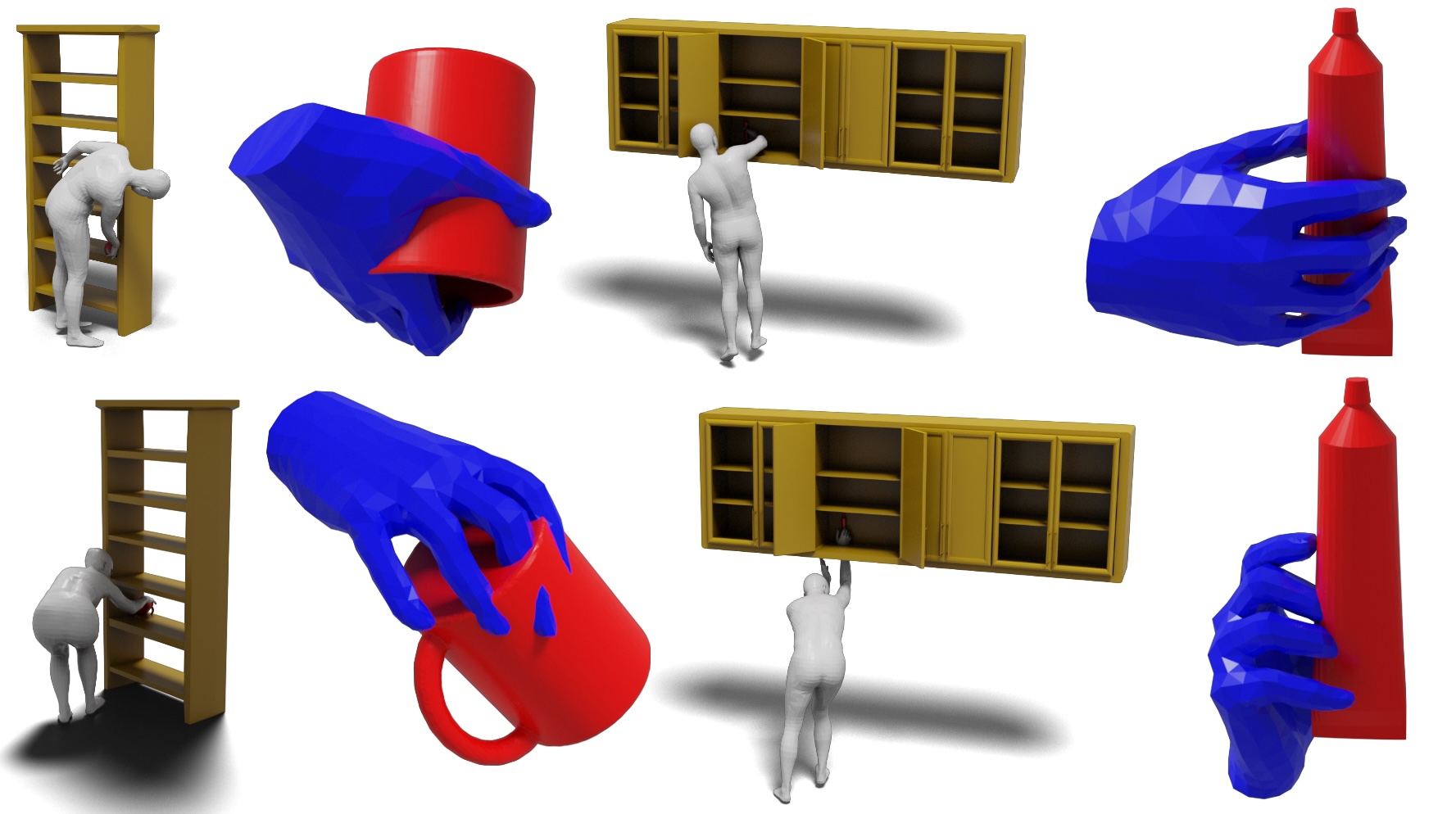} 
    \\
    \includegraphics[width=0.99 \textwidth]{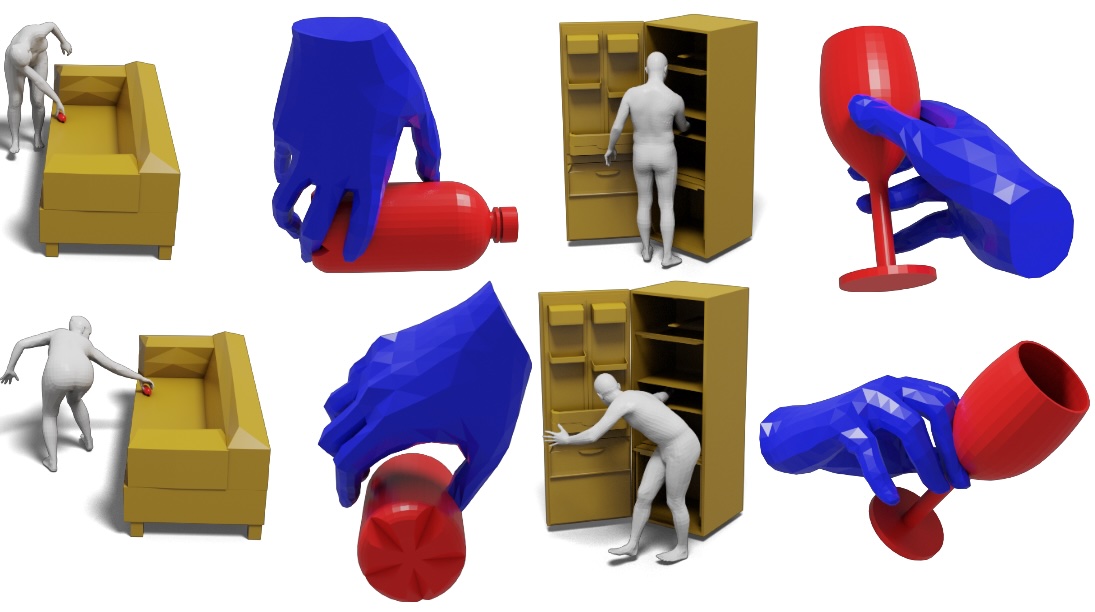} 
    \caption{\captionCWGraspVsFLEX}
    \label{fig:supmat_cwgrasp_qual3}
\end{figure*}

\begin{figure*}
    \centering
    \includegraphics[width=\textwidth]{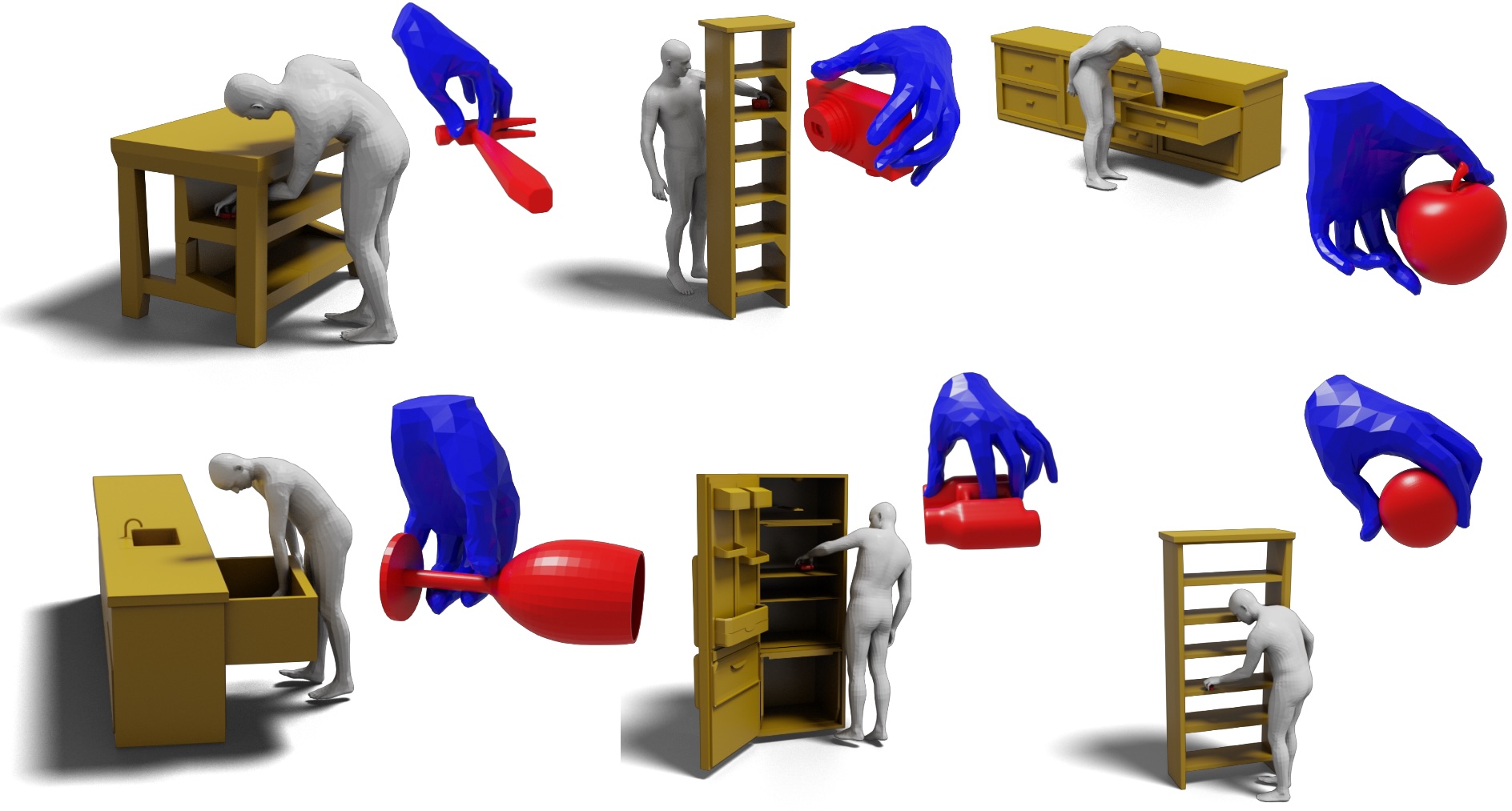} 
    \includegraphics[width=\textwidth]{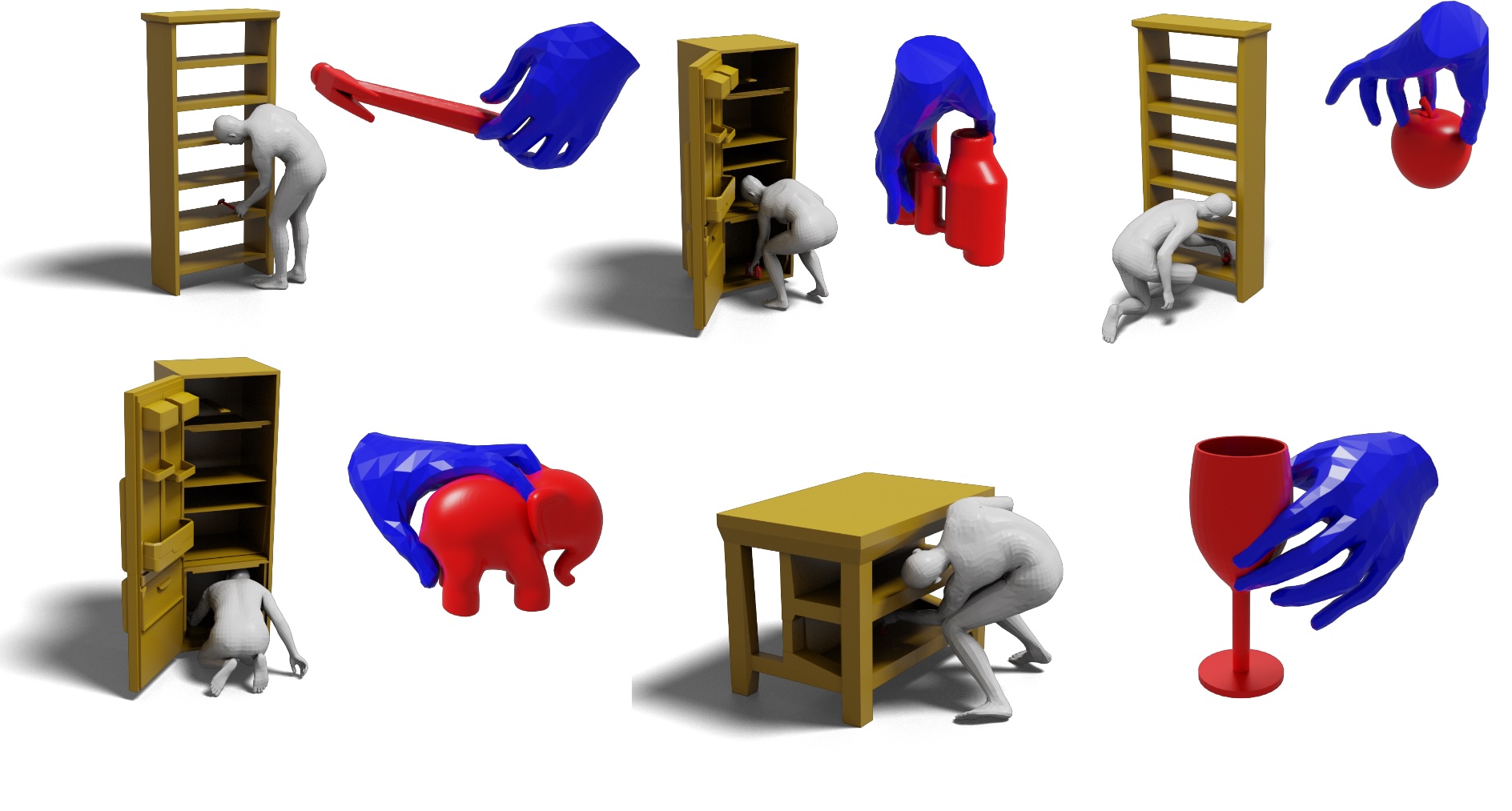}
    \caption{Qualitative results of \cWGrasp when using the left hand. Note that \cWGrasp is unique in generating both \rhand and \lhand whole-body grasps, while performance is similar for both cases. 
    }
    \label{fig:supmat_cwgrasp_qual_left}
\end{figure*}
\begin{figure*}
    \centering
    \includegraphics[width=0.9 \textwidth]{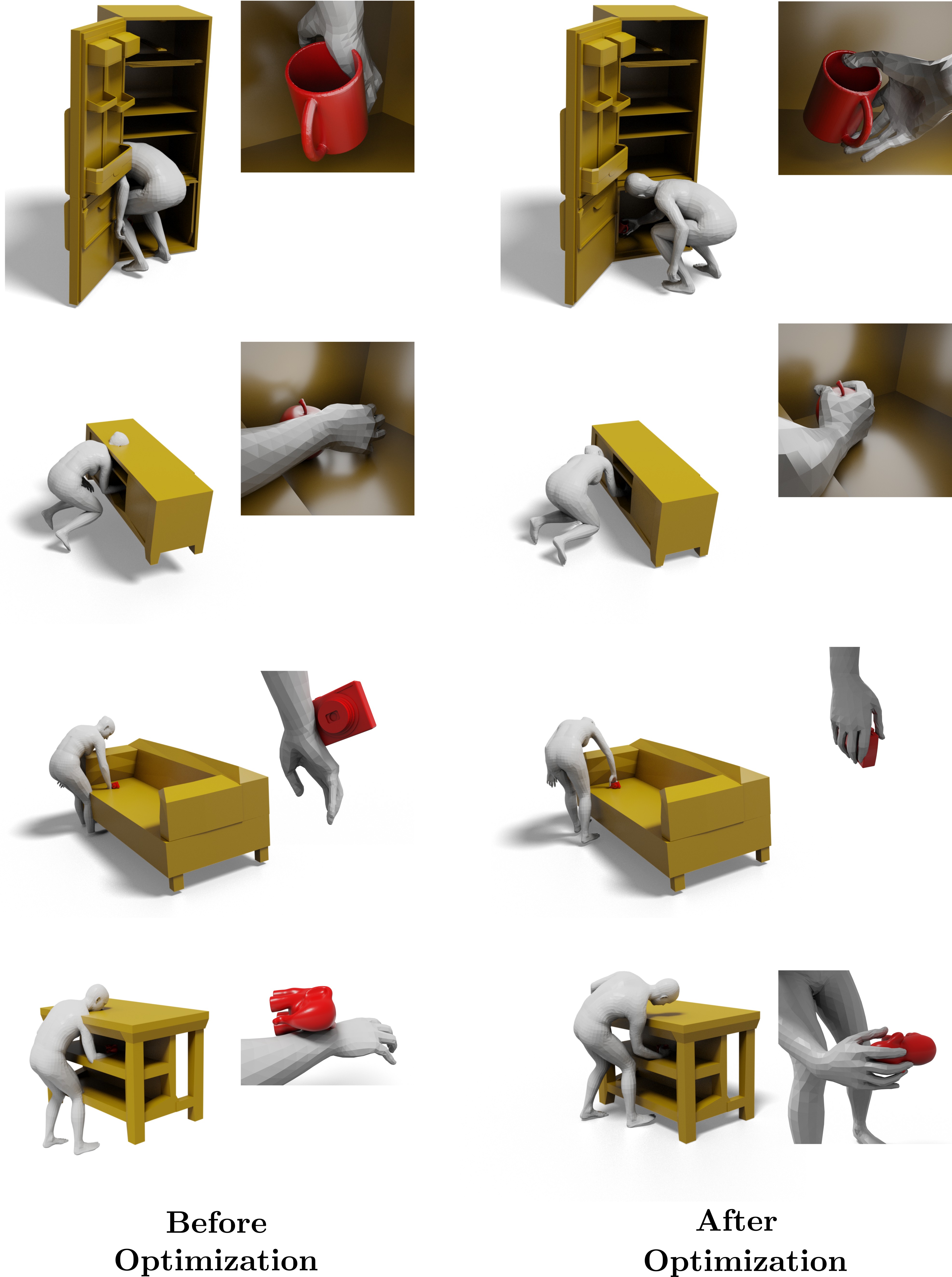} 
    \caption{
        Qualitative evaluation of \cWGrasp optimization performance. 
        We show the generated interacting bodies before (left) and after (right) optimization with \cWGrasp. 
        For each example we show both a full-body view and a close-up view on the hand and object.} 
    
    \label{fig:before_after_opt}
\end{figure*}

\end{document}